\documentclass[10pt,twocolumn,letterpaper]{article}

\usepackage{wacv}              

\definecolor{wacvblue}{rgb}{0.21,0.49,0.74}
\usepackage[pagebackref,breaklinks,colorlinks,allcolors=wacvblue]{hyperref}
\usepackage{cuted}
\usepackage{capt-of}
\usepackage{multirow}

\def\wacvPaperID{1190} 
\def\confName{WACV}
\def\confYear{2027}

\title{EGGHead: Efficient Generalizable Gaussian Heads from a Single Image}

\author{
Pilseo Park$^{1}$\thanks{Equal contributions.}, 
Andrew Hou$^{1}$\footnotemark[1], 
Michel Sarkis$^{2}$\thanks{Formerly of Qualcomm Technologies.}, 
Ning Bi$^{2}$,
Yiying Tong$^{1}$,
Xiaoming Liu$^{1,3}$\\[0.5ex]
$^{1}$Michigan State University \qquad
$^{2}$Qualcomm Technologies, Inc.\\
$^{3}$University of North Carolina at Chapel Hill\\[0.5ex]
{\tt\small \{parkpils, houandr1, ytong\}@msu.edu},
{\tt\small \{msarkis, nbi\}@qti.qualcomm.com},\\
{\tt\small liuxm@cs.unc.edu}
}

\DeclareMathOperator{\sigmoid}{sigmoid}

\begin{document}

\maketitle
\begin{strip}
\vspace{-24mm}
\centering
\includegraphics[width=\textwidth]{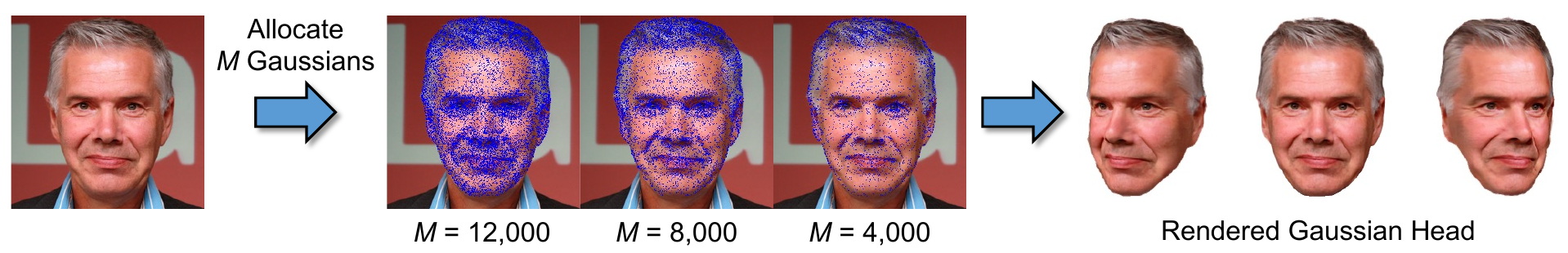}
{\captionsetup{hypcap=false}
\captionof{figure}{
    We allocate a fixed Gaussian budget \textit{M} for Gaussian head  reconstruction. Even with a compact Gaussian budget, our method preserves representational power, highlighting the efficiency of our allocation strategy. The rendered head shown here uses $M{=}12$\textit{k}.
}
\label{fig1}
}
\end{strip}

\begin{abstract}
Creating photorealistic human heads typically involves trade-offs between representation power, efficiency, and generalizability. Traditional mesh-based solutions relying on $3$DMMs generally sacrifice representation power but are efficient to render. On the other hand, NeRF-based methods improve representation power and fidelity while sacrificing efficiency. Gaussian splatting methods attempt to resolve these trade-offs by offering both high representation power and efficiency with real-time rendering. However, their efficiency nonetheless depends on the number of rasterized Gaussians, and the time needed to optimize and prepare the head (the enrollment time) is often high. In light of this, we propose EGGHead, which introduces a novel Gaussian allocation algorithm that efficiently assigns Gaussians to the head’s semantic regions given a fixed Gaussian budget. Our algorithm reduces the number of Gaussians needed while minimizing the impact on representation power. EGGHead operates in a single forward pass and outperforms previous single-image face modeling methods in reconstruction quality, novel view synthesis, and rendering speeds.
\end{abstract}
    
\section{Introduction}
\label{sec:intro}
Modeling photorealistic human heads has received widespread attention from the computer vision and graphics communities. It has a wide array of applications, including movie production~\cite{therelightables}, digital humans in AR/VR~\cite{park2021nerfies, gaussianheadavatar, relightablegaussianavatars, NeRFace}, and downstream tasks in the face domain~\cite{joint-face-alignment, disentangling-features}. There are several important axes to consider when designing a head reconstruction method, including representation power, the efficiency of the $3$D representation, and data requirements. An ideal method produces a high-quality head that captures fine identity details, enables real-time generation and novel view synthesis, and requires minimal data.

Traditional head representation methods adopt mesh-based approaches that rely on $3$DMMs~\cite{3DMM} to model the face. Although generally efficient to rasterize and render, and despite a plethora of single-image based methods~\cite{wang2022faceverse, towards-high-fidelity-nonlinear-3d-face-morphable-model, UncertaintyFaceReconstruction, LuanCVPR18, MonocularNeuralReflectance}, these $3$D representations generally lack high representation power due to reliance on PCA bases, carrying over artifacts from intrinsic decomposition into rendering. The follow-up methods~\cite{park2021nerfies, park2021hypernerf, infamousnerf, NeRFace, zhuang2022mofanerf, hong2021headnerf} relied instead on the implicit NeRF~\cite{NeRF} representation, which allowed for higher representation power thanks to per-pixel raytracing, the ability to model view-specific texture variations, and the volumetric rendering design. However, NeRF-based methods sacrifice efficiency since the raytracing inference stage is often costly. More recently, the face modeling community has favored mesh-driven 3D Gaussian Splatting~\cite{gaussiansplatting} (3D-GS): a hybrid explicit-implicit representation that allows for both high fidelity and real-time rendering.

Gaussian Splatting, however, is not without limitations. While considered a representation that enables real-time rendering, the efficiency is still affected by the number of Gaussians used to represent the head. In practice, high-quality Gaussian heads require a large number of Gaussians to capture fine geometric and appearance details. Reducing the number of Gaussians is thus an important priority, but if done naively will simply reduce the head's fidelity. To address this trade-off between fidelity and efficiency, we propose a novel Gaussian Allocation Module (GAM), which efficiently allocates a fixed budget of $M$ Gaussians while minimizing the impact on the head's fidelity.

Moreover, 3D-GS traditionally relies on monocular or multi-view image/video inputs, is trained per subject instead of being generalizable, and requires an expensive optimization for each subject ranging from a few minutes to one hour~\cite{qiangaussianavatars, chen2024monogaussianavatar, gaussianheadavatar}. GPS-Gaussian~\cite{Zheng2023GPSGaussianGP} resolves some of these limitations by producing a generalizable method that takes a stereo pair of images as input and predicts Gaussian attributes. However, it requires a monocular video as input, since the method depends on the nearest-neighbor stereo pair for each queried viewpoint. Moreover, it makes the naive assumption of assigning one Gaussian to each pixel, which will naturally reduce the rendering speed. We instead reduce this requirement from a monocular video to a single image while also proposing a generalizable method given one image as input. Our novel GAM improves rendering speed by allocating Gaussians over semantic regions rather than on a per-pixel basis, achieving high fidelity with fewer Gaussians. In addition, we employ Voronoi tessellation~\cite{Budninskiy2016OptimalVT} to guide Gaussian attribute computation in a single-view setting with sparse Gaussians. Finally, we introduce a lightweight, network-free Face Symmetry Module to render occluded face regions, avoiding expensive post-processing networks and diffusion-based multi-view generation during inference.

Our contributions are thus as follows: 

$\diamond$ We propose EGGHead, a single-image and generalizable Gaussian head model that outperforms existing approaches in rendering quality and speed while achieving competitive enrollment time.

$\diamond$ We introduce a Gaussian Allocation Module (GAM) that allocates Gaussians under a fixed budget with minimal impact, employ Voronoi tessellation to guide Gaussian shape computation in a single-view setting with sparse Gaussians, and incorporate a lightweight Face Symmetry Module to improve head completeness without relying on additional networks.

$\diamond$ We demonstrate superior head representation capabilities and fidelity among single-image head reconstruction methods, both quantitatively and qualitatively.
\section{Related Work}
\label{sec:rw}
\paragraph{Face Modeling.} Existing $3$D face representations include explicit mesh-based methods~\cite{LuanCVPR18, 3d-face-modeling-from-diverse-raw-scan-data, towards-high-fidelity-nonlinear-3d-face-morphable-model, TowardsHighFidelityFaceReconstruction, face-relighting-with-geometrically-consistent-shadows, disentangling-features, wang2022faceverse, UncertaintyFaceReconstruction, MonocularNeuralReflectance, shang2020self, joint-face-alignment, riggable-3d-face-reconstruction-via-in-network-optimization, DFNRMVS, on-learning-3d-face-morphable-model-from-in-the-wild-images}, implicit NeRF-based~\cite{NeRF} methods~\cite{NeRFace, LOLNeRF, park2021nerfies, 3DMM-RF, RigNeRF, NHA, park2021hypernerf, Gao-portraitnerf, CompositionalRF, zhuang2022mofanerf, hong2021headnerf, chanmonteiro2020pi-GAN, zhang2022fdnerf, infamousnerf}, diffusion models conditioned with $3$D priors~\cite{diffusionrig, Kant2025Pippo}, $3$D-aware GANs that incorporate triplane representations~\cite{PanoHead, lan2022e3dge, EG3D, stylenerf, QuantManip, NeRFInvertor, shi2023pof3d, GRAF, GeoD, deng2022gram, yu25gaia, he2025head}, and $3$D Gaussian Splatting ($3$D-GS)~\cite{gaussiansplatting, relightablegaussianavatars, gagavatar, qiangaussianavatars, ma2024gaussianblendshapes, chen2024monogaussianavatar, shao2024splattingavatar, gaussianheadavatar, TaoAvatar, tang2024gaf, WangARXIV2025, kirschstein2024gghead}.
Mesh-based 3DMMs~\cite{3DMM} enable fast rendering but are limited in fidelity by their PCA basis. NeRF-based methods improve fidelity through view-dependent neural rendering, but require costly ray sampling. 3D-GS offers high-quality real-time rendering, yet its efficiency depends on the number of Gaussians and still requires seconds to minutes of per-subject enrollment. To address this, we introduce EGGHead, a generalizable model that effectively reduces the Gaussian budget.

\vspace{-4mm}

\paragraph{Generalizable Gaussian Splatting.} Although 3D-GS is traditionally optimized per scene or subject, generalizable 3D-GS methods have recently gained attention. GPS-Gaussian~\cite{Zheng2023GPSGaussianGP} synthesizes unseen identities but assigns one Gaussian per pixel, limiting efficiency. GAGAvatar~\cite{gagavatar} predicts Gaussians from a single image using FLAME~\cite{FLAME} and dual lifting, but its additional neural renderer may compromise identity consistency. Diffusion-based methods~\cite{cap4d, facelift} achieve high quality but rely on costly multi-view generation. In contrast, EGGHead uses GAM to merge Gaussians in fidelity-insensitive regions, avoiding dense per-pixel allocation, and multi-view generation.

\vspace{-4mm}

\paragraph{Efficient Gaussian Splatting.} Several methods improve the efficiency of $3$D-GS by compressing Gaussian parameters or pruning low-impact Gaussians. Compression methods~\cite{Lee_2024_CVPR, Niedermayr_2024_CVPR, navaneet2023compact3d} reduce storage and rendering costs by removing parameter redundancy, while pruning-based methods~\cite{fan2023lightgaussian, eagles, reducingmemfootprintGS} reduce the Gaussian count. However, they still rely on costly multi-view optimization. In contrast, our GAM operates from a single image and directly allocates a compact set of Gaussians.

\vspace{-4mm}

\paragraph{Symmetry in Face Modeling.} Face symmetry is a strong prior that preserves the naturally symmetrical structure of human faces, especially in the ill-posed single-image setting where occluded regions rely on face priors to be properly reconstructed. While previous methods~\cite{3DFaceFill,towards-high-fidelity-nonlinear-3d-face-morphable-model,FFHQUV} impose symmetry as a loss function to guide networks toward symmetric $3$D face reconstruction, EGGHead introduces a lightweight Face Symmetry Module that directly mirrors visible Gaussians across the facial symmetry plane to complement occluded regions without additional networks.

\begin{figure*}[t!]
\centering
\includegraphics[width=1.0\textwidth]{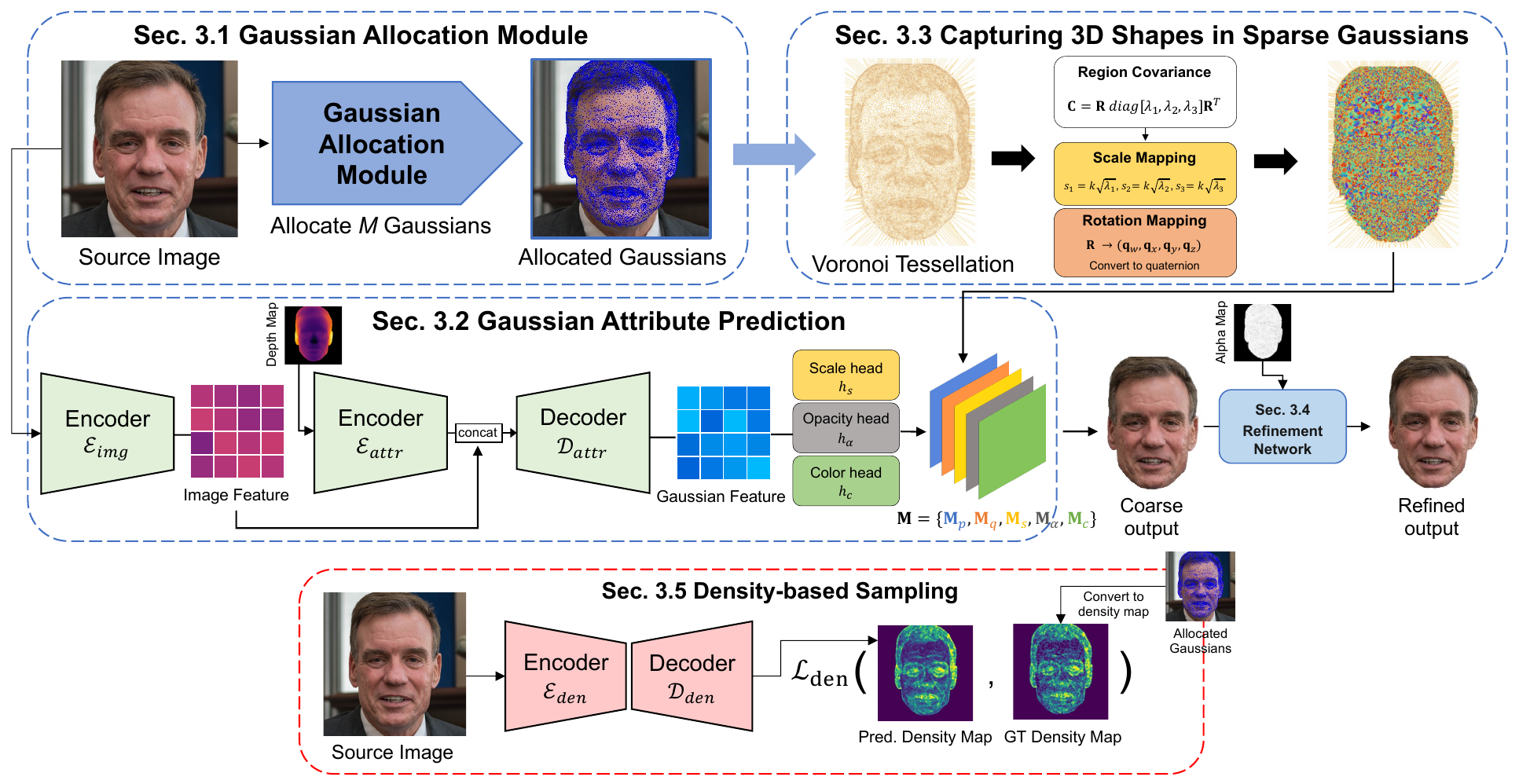}
\vspace{-5mm}
\caption{
\small \textbf{Training Pipeline.}
We propose a single-image, generalizable, and lightweight Gaussian head model that adaptively allocates \textit{M} Gaussians with minimal impact on rendering quality. Secs. 3.1–3.4 constitute the main reconstruction pipeline: GAM first allocates \textit{M} Gaussians from the source image, Voronoi tessellation provides sparse geometric priors, and image/depth features are used to predict Gaussian attributes for rendering and refinement. Sec. 3.5 is trained in parallel to predict a density map supervised by the GAM-derived density map. It is independent of the main reconstruction pipeline during training and is used at inference time for density-based Gaussian allocation.
\vspace{-2mm}}
\label{fig2}
\end{figure*}

\section{Method}
\label{sec:method}
\paragraph{Problem Formulation.} Given a single source image $\mathbf{I}_s$, our goal is to reconstruct a photorealistic head of an unseen subject. To tackle this non-trivial task, we employ the $3$D Gaussian Splatting ($3$D-GS) pipeline~\cite{gaussiansplatting}. $3$D-GS models a scene with a set of $3$D Gaussians, each defined by a $3$D covariance matrix $\Sigma$ and a center (mean) position $p$:
\begin{equation}
\mathbf{G(x)} = e^{-\frac{1}{2}(\mathbf{x} - \mathbf{p})^{T} \Sigma^{-1} (\mathbf{x} - \mathbf{p})}.
\label{eq:eq1}
\end{equation}
Each Gaussian $\mathbf{G}$ is parameterized by $\mathcal{G} = \{ \mathbf{p}, \mathbf{q}, \mathbf{s}, \alpha, \mathbf{c} \}$ where $\mathbf{p} \in \mathbb{R}^3$ is the center position, $\mathbf{q} \in \mathbb{R}^4$ is the rotation quaternion, $\mathbf{s} \in \mathbb{R}^3$ is the scaling factor, $\alpha \in \mathbb{R}^1$ is the opacity, and $\mathbf{c} \in \mathbb{R}^3$ is the color.

Applying $3$D-GS to our task is non-trivial, as we aim to render novel views under (1) a single-view setting, (2) a generalizable model without per-subject fine-tuning, and (3) a reduced number of Gaussians rather than a per-pixel allocation. We explain our solution in the following sections.

\vspace{-5mm}

\paragraph{Architecture.}
Fig.~\ref{fig2} illustrates our pipeline. Given a single source image, our approach generates a photorealistic head representation using a fixed Gaussian budget without per-subject optimization. During training, we introduce the Gaussian Allocation Module (GAM) to achieve an effective trade-off between the number of Gaussians and rendering quality (Sec.~\ref{sec:gaussianallocation}). We employ a Gaussian Attribute Regressor to predict Gaussian attribute maps (Sec.~\ref{sec:gspred}), guided by Voronoi tessellation~\cite{Budninskiy2016OptimalVT} for more accurate shape computation (Sec.~\ref{sec:voronoi}). The framework is trained to minimize the difference between the rendered and input images (Sec.~\ref{sec:training}). In inference, GAM is replaced with density-based sampling for efficiency (Sec.~\ref{sec:sampling}), and a Face Symmetry Module completes the occluded regions (Sec.~\ref{sec:symm}).

\subsection{Gaussian Allocation Module}
\label{sec:gaussianallocation}
Our objective is to efficiently allocate a fixed Gaussian budget while preserving rendering quality.
We begin by assigning one Gaussian per pixel, compute the error between adjacent Gaussian pairs based on their mean texture difference, and iteratively merge the pair with the lowest error until reaching the target number of Gaussians $M$. After each merge, all impacted neighboring Gaussians have their mean texture error updated with the newly merged Gaussian. To prevent oversized Gaussians and sparsely covered regions, we set a maximum threshold such that no Gaussian can occupy more than $\frac{5*B}{M}$ pixels, where $B$ is the number of foreground pixels in the binary face mask.

The resulting positions of these per-pixel Gaussians form a pseudo ground-truth position map in pixel coordinates. In practice, we convert these pseudo ground-truth position maps into continuous density maps and use them to train a network to directly predict a Gaussian position density map given a source image. We then sample our $M$ Gaussians from the predicted density map (See Sec.~\ref{sec:sampling}). Using the predicted depth map $\mathbf{D}$ together with a scaling factor $s_\text{img}$, the sampled Gaussian centers $(i_m, j_m)$ can be lifted from the image plane to $3$D as:
\begin{equation}
\mathbf{M}_p = \{ \big(s_\text{img}(i_m-c_i,\,j_m-c_j),\, \mathbf{D}_{i_m,j_m}\big) \}_{m=1}^{M},
\label{eq:eq2}
\end{equation}
where $(c_i,c_j)$ is the center of the image.

\subsection{Gaussian Attribute Prediction}
\label{sec:gspred}
We aim to model a generalizable function $F(\mathbf{I}_s, \mathbf{D}) = \{\mathcal{G}_m\}_{m=1}^{M}$ that estimates the attributes of $M$ Gaussians conditioned on each subject. 

Following the allocated Gaussians introduced in Sec.~\ref{sec:gaussianallocation}, we formulate their attributes as pixel-wise maps, allowing a $2$D CNN to aggregate neighboring information for local coherence and computational efficiency. Specifically, we compute pixel-wise attribute maps $\mathbf{M} = { \mathbf{M}_{p}, \mathbf{M}_{q}, \mathbf{M}_{s}, \mathbf{M}_{\alpha}, \mathbf{M}_{c} }$ for position, rotation, scale, opacity, and color, respectively. 

The position map $\mathbf{M}_p$ is constructed directly as described in Sec.~\ref{sec:gaussianallocation}, without regression. The rotation map $\mathbf{M}_q$ is obtained from covariance matrices computed by the Voronoi tessellation of sampled Gaussians (see Sec.~\ref{sec:voronoi}). The scale map $\mathbf{M}_s$ is initialized by Voronoi tessellation and refined with the scale adjustment map $\mathbf{M}_s^\gamma$. Similarly, RGB values are directly used as the color map for efficiency and further refined by the color adjustment map $\mathbf{M}_c^\Delta$ to improve quality. Therefore, only three attribute maps are regressed:
\begin{equation}
\mathbf{M} = \{ \mathbf{M}_{s}^\gamma, \mathbf{M}_{\alpha},  \mathbf{M}_{c}^\Delta\}.
\label{eq:eq3}
\end{equation}

To predict the attribute maps, a source image $\mathbf{I}_s \in [0,1]^{H \times W \times 3}$ is fed into an image encoder ${\mathcal{E}}_\text{img}$. The image feature $\mathcal{E}_\text{img}(\mathbf{I}_s)$ and the depth feature $\mathcal{E}_\text{attr}(\mathbf{D})$ are concatenated and then passed to the decoder ${\mathcal{D}}_\text{attr}$ that generates a pixel-wise Gaussian feature:
\begin{equation}
\mathbf{F} = \mathcal{D}_\text{attr}(\mathcal{E}_\text{img}(\mathbf{I}_s) || \mathcal{E}_\text{attr}(\mathbf{D})),
\label{eq:eq4}
\end{equation}
where $\mathbf{F} \!\in\! \mathbb{R}^{H \times W \times C}$ and $\cdot || \cdot$ denotes concatenations across all feature levels. 
Note that we only regress the scale adjusting map $\mathbf{M}_{s}^\gamma$, the opacity map $\mathbf{M}_{\alpha}$, and the color adjusting map $\mathbf{M}_{c}^\Delta$, with corresponding prediction heads ${h}_s$, ${h}_\alpha$ and $h_c$ composed of two convolutional layers:
\begin{equation}
\begin{array}{c}
\mathbf{M}_{s}^{\gamma} = \sigmoid(h_{s}(\mathbf{F})), \quad
\mathbf{M}_{\alpha} = \sigmoid(h_{\alpha}(\mathbf{F})), \\[4pt]
\mathbf{M}_{c} = \mathbf{I}_s + \mathbf{M}_{c}^\Delta
= \mathbf{I}_s + \tanh(h_{c}(\mathbf{F})).
\end{array}
\label{eq:eq5}
\end{equation}
Details of the network architecture are provided in the supplementary materials.

\subsection{Capturing 3D Shapes in Sparse Gaussians}
\label{sec:voronoi}
In single-view and sparse Gaussian settings, accurately capturing $3$D Gaussian shapes via regression is non-trivial due to limited depth information~\cite{Hu2023gaussianAvatarTR}. To address this, we leverage Voronoi tessellation~\cite{Budninskiy2016OptimalVT} to provide structural guidance. Given Gaussian centers in $3$D coordinates $P := \{ p_1, \dots, p_M \}$ on a surface, a Voronoi tessellation divides the surface into $M$ regions, each containing points closer to one center. Then, the covariance matrix of the resulting $3$D polygon captures its spatial variance and orientation.

However, such intrinsic tessellation has $O(M^2)$ time complexity, which is prohibitively high for efficient processing. We therefore perform tessellation in 2D with $O(M\log M)$ time complexity, and lift each region $R$ to $3$D using the depth. To compute its covariance matrix $\mathbf{C}$, we partition it into $N$ triangles $T_i$:
\begin{equation}
R = T_1 \cup T_2 \cup \cdots \cup T_{N-1} \cup T_N,
\label{eq:eq6}
\end{equation}
where each triangle vertex is lifted to $3$D using its corresponding depth value obtained via nearest-neighbor search.

With the barycenter of the region $R$ computed through an area-weighted average of the triangle centers for the constituent triangles, $R$'s covariance matrix can be assembled by the covariance matrices of the triangles. The resulting covariance matrix reliably represents the geometric structure of the Voronoi region, capturing its spatial distribution and orientation through its eigen decomposition $\mathbf{C}=\mathbf{R}\;\mathrm{diag}[\lambda_1, \lambda_2,\lambda_3]\; \mathbf{R}^T,$ where $\lambda_1$, $\lambda_2$, and $\lambda_3$ represent the eigenvalues in descending order and $\mathbf{R}$ is the rotation matrix assembled from the eigenvectors.

We then assign the scale attributes of Gaussians as:
\begin{equation}
s_1 = k \sqrt{\lambda_1}, \quad s_2 = k \sqrt{\lambda_2}, \quad s_3 = k \sqrt{\lambda_3},
\label{eq:eq7}
\end{equation}
where $k$ denotes the confidence interval, which in our approach is the scale adjusting map $\mathbf{M}_s^\gamma$. Thus, the final scale map $\mathbf{M}_{s}$ is represented as:
\begin{equation}
\mathbf{M}_s = \mathbf{M}_s^\gamma \odot \{ \sqrt{\lambda} \}_{m=1}^{M}.
\label{eq:eq8}
\end{equation}

Next, the rotation matrix $\mathbf{R}$ is converted to its quaternion representation $\mathbf{q} \in \mathbb{R}^4$. The rotation 
map $\mathbf{M}_{q}$ is thus:
\begin{equation}
\mathbf{M}_{q} = \{ (\mathbf{q}_w, \mathbf{q}_x, \mathbf{q}_y, \mathbf{q}_z) \}_{m=1}^{M}.
\label{eq:eq9}
\end{equation}
With all parameters of $\mathcal{G} = \{ \mathbf{p}, \mathbf{q}, \mathbf{s}, \alpha, \mathbf{c} \}$ defined, we render the reconstructed image $\mathbf{I}_{r}$.

\subsection{Refinement Network}
\label{sec:refnet}
High-fidelity facial rendering typically requires a large number of points to capture appearance details~\cite{gagavatar}. However, our approach reduces the number of Gaussians for efficiency, and to ensure robust generalization across a vast range of identities, we adopt a source-view reconstruction setting instead of video-based supervision. This inherently limits the ability to recover facial details during novel view synthesis. To address this, we introduce a refinement network. To maintain overall efficiency, we adopt a lightweight U-Net style module. The network takes the coarse output $\mathbf{I}_{r}$ and the alpha map $\mathbf{I}_{\alpha}$ obtained via alpha blending of the Gaussians as input and generates the final output $\mathbf{I}_{r}'$. Please refer to the supplementary material for more details on the network architecture.

\subsection{Density-based Sampling}
\label{sec:sampling}
\paragraph{Density Map Prediction.} During inference, we replace GAM with density-based sampling to improve computational efficiency. Since directly predicting the discrete coordinates of $M$ Gaussians is sensitive to subtle variations, we learn a continuous density map instead. We generate the ground-truth density map $\mathbf{M}^{GT}_{d}$ by placing Gaussian centers obtained from GAM on a binary map and smoothing it with a $5 \times 5$ Gaussian kernel.

Using the resulting density map $\mathbf{M}^{GT}_{d}$ as ground-truth, we construct a density prediction network that takes an image $\mathbf{I}_s$ as input and predicts the corresponding density map:
\begin{equation}
\mathbf{M}_{d} = \mathcal{D}_\text{den}(\mathcal{E}_\text{den}(\mathbf{I}_s)),
\label{eq:eq10}
\end{equation}
where $\mathbf{M}_{d}$ is the predicted density map, $\mathcal{E}_\text{den}$ the density encoder, and $\mathcal{D}_\text{den}$ is the density decoder.

\paragraph{Adaptive Sampling.} From the predicted density map $\mathbf{M}_{d}$, we normalize its values to create a probability distribution $\mathcal{P}(i, j)$:
\begin{equation}
\mathcal{P}(i, j) = \frac{\mathbf{M}_{d}(i, j)}{\sum_{i=1}^{H} \sum_{j=1}^{W} \mathbf{M}_{d}(i, j)},
\label{eq:eq11}
\end{equation}
where $H$ and $W$ denote the height and width of the density map. This distribution turns the density map into a likelihood surface, where larger values yield higher sampling probabilities. We then perform Monte Carlo sampling~\cite{Hastings1970MonteCS} $\mathcal{S}$ to obtain a final set of $M$ Gaussian samples:
\begin{equation}
\{(i_m, j_m)\}_{m=1}^{M} \sim \mathcal{S}(\mathcal{P}(i, j)),
\label{eq:eq12}
\end{equation}
where $(i_m, j_m)$ represents the coordinates of sampled Gaussian centers. This ensures that areas of higher density yield more samples, capturing regions of interest effectively while maintaining variability and robustness against noise.

\begin{figure}
\centering
\includegraphics[width=\columnwidth]{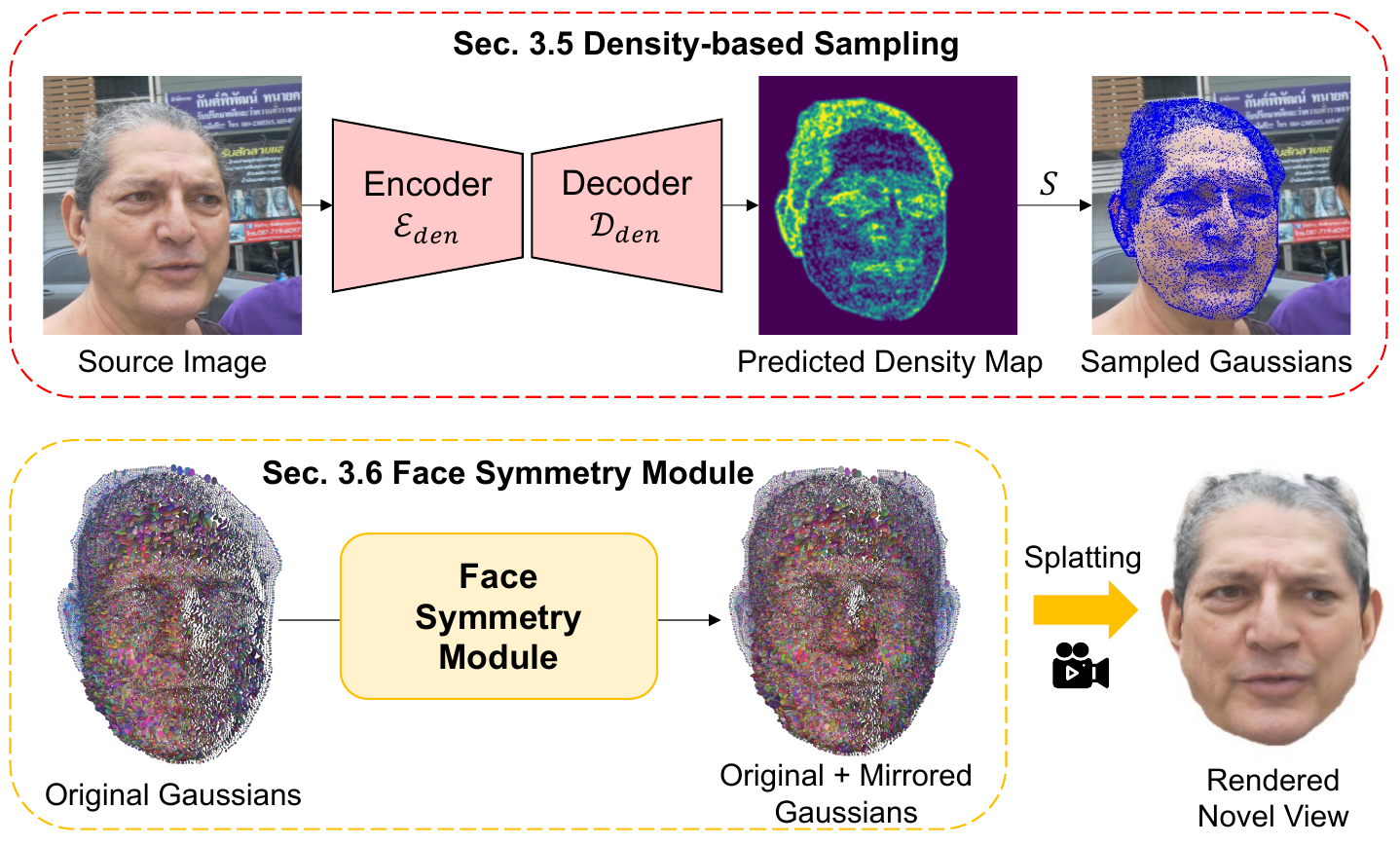}
\caption{\textbf{Inference-time Pipeline.} During inference, we replace GAM with density-based sampling to reduce computation, and incorporate a Face Symmetry Module to complete occluded regions. \vspace{-3mm}}
\label{fig:fig3}
\vspace{-5mm}
\end{figure}

\subsection{Face Symmetry Module}
\label{sec:symm}
While our method covers visible regions of the head from the input image well, occluded regions also need to be handled to improve the completeness of the reconstructed head. Human faces typically have sufficient bilateral symmetry to serve as a practical prior. We leverage this symmetry to generate the missing side of the head from a single input image. This design avoids the need for diffusion-based multi-view generation~\cite{cap4d, facelift} or an additional neural network~\cite{kwon2024ghg, gagavatar}, making the method computationally lightweight.

We introduce a Face Symmetry Module to leverage this prior. We first transform the Gaussians into a canonical view using the camera parameters estimated by DECA~\cite{DECA}, where the symmetry plane is fixed as $x=0$. We then mirror the Gaussians across this plane to create a complementary set for the partially occluded side. Directly combining the original and mirrored Gaussians leads to conflicts in overlapping regions, so we assign low opacity to mirrored Gaussians in overlapped regions and increase their opacity with distance. This yields a smooth transition between the two sets and avoids visible seams. Fig.~\ref{fig:fig3} shows the processes described in Sec.~\ref{sec:sampling} and Sec.~\ref{sec:symm}.

\subsection{Training Objective}
\label{sec:training}
Our training pipeline follows two parallel branches starting from the same input image $\mathbf{I}_s$, each with its own objective.
\subsubsection{Density Prediction Network.}
Given $\mathbf{I}_s$, the network predicts a density map $\mathbf{M}_{d}$, and is trained with L2 loss between $\mathbf{M}_{d}$ and $\mathbf{M}^{GT}_{d}$:
\begin{equation}
\mathcal{L}_{\text{den}} = \|\mathbf{M}_{d} - \mathbf{M}_{d}^{GT}\|_2^2.
\label{eq:eq13}
\end{equation}

\subsubsection{Gaussian Attribute Regressor.}
We minimize the regressor loss:
\begin{equation}
\mathcal{L}_{\text{reg}} = \lambda_{ref} \mathcal{L}_{ref} + \lambda_\text{coarse} \mathcal{L}_{\text{coarse}} + \lambda_\alpha \mathcal{L}_\alpha,
\label{eq:eq14}
\end{equation}
which consists of a refinement loss $\mathcal{L}_{ref}$, a coarse loss $\mathcal{L}_{coarse}$ and an alpha loss $\mathcal{L}_{\alpha}$. Further details on loss functions can be found in the supplementary material.

\begin{figure*}[t]
\centering
\setlength{\tabcolsep}{1pt}
\renewcommand{\arraystretch}{0.75}
\resizebox{\textwidth}{!}{%
\begin{tabular}{c c@{}c c@{}c c@{}c c@{}c}
\includegraphics[width=0.1\textwidth]{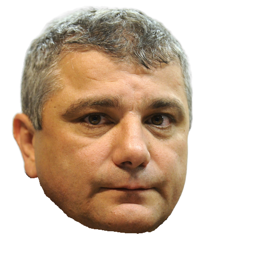} &
\includegraphics[width=0.1\textwidth]{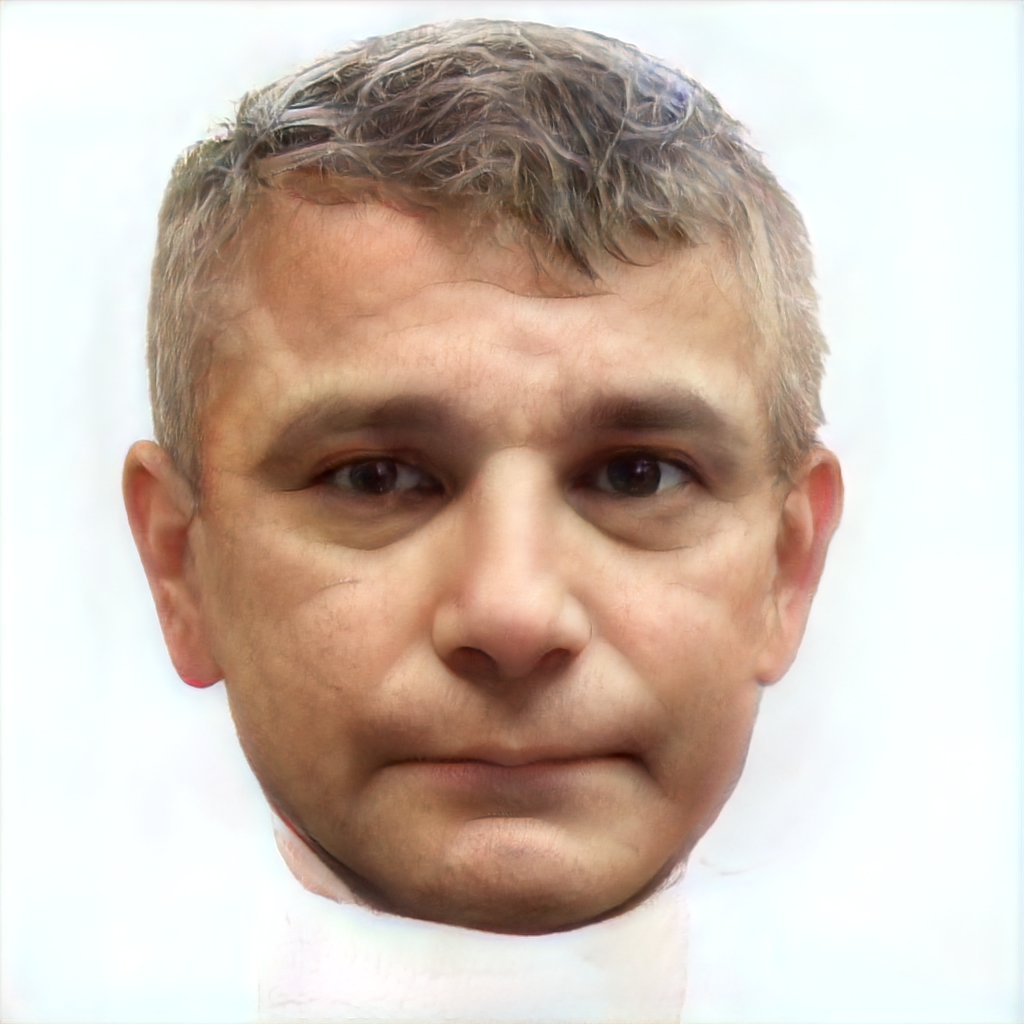} &
\includegraphics[width=0.1\textwidth]{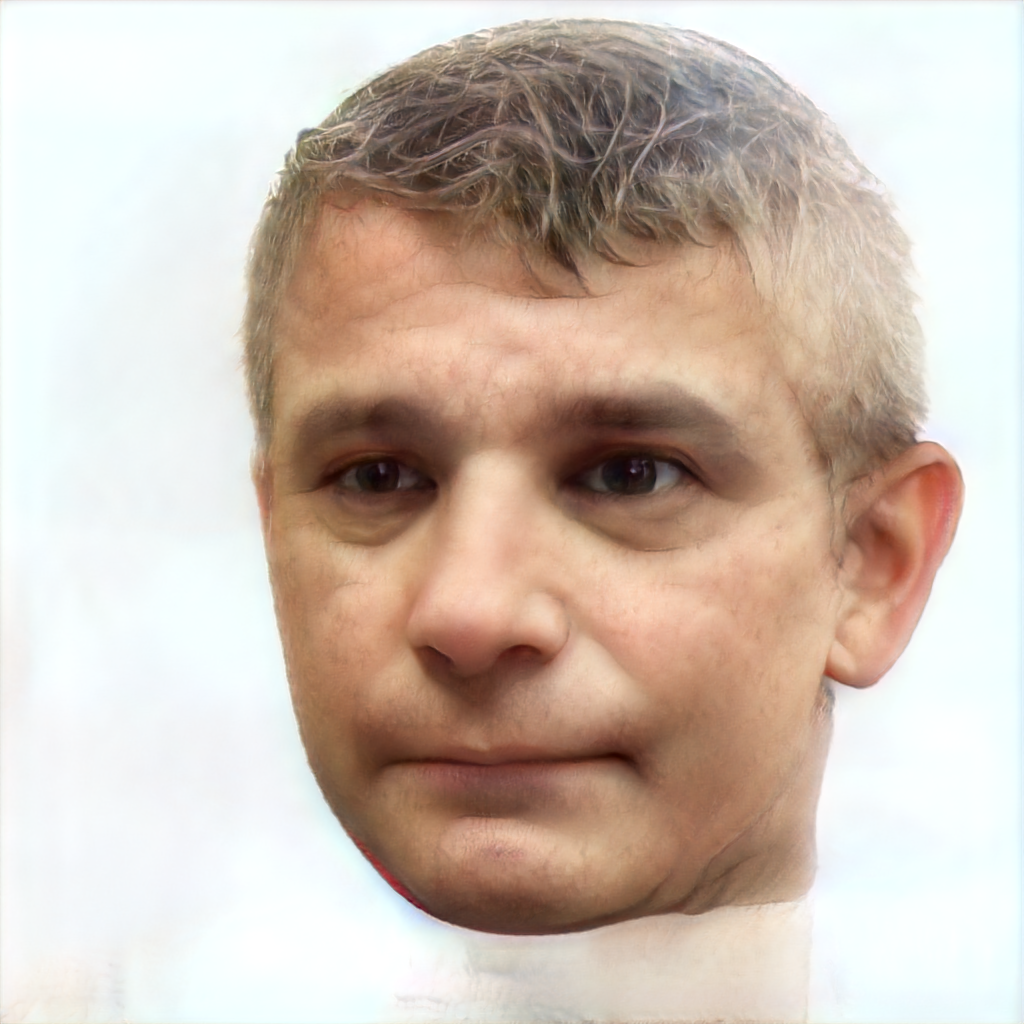} &
\includegraphics[width=0.1\textwidth]{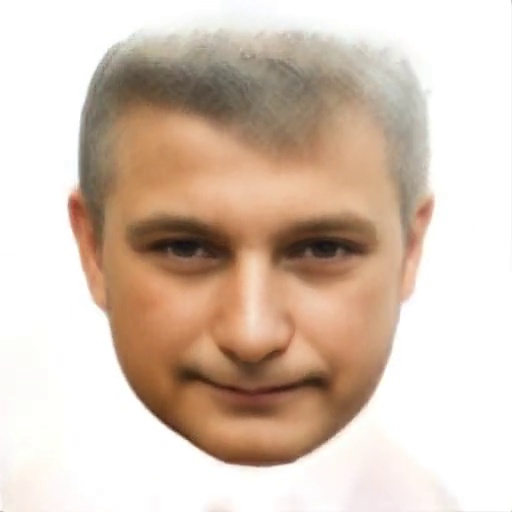} &
\includegraphics[width=0.1\textwidth]{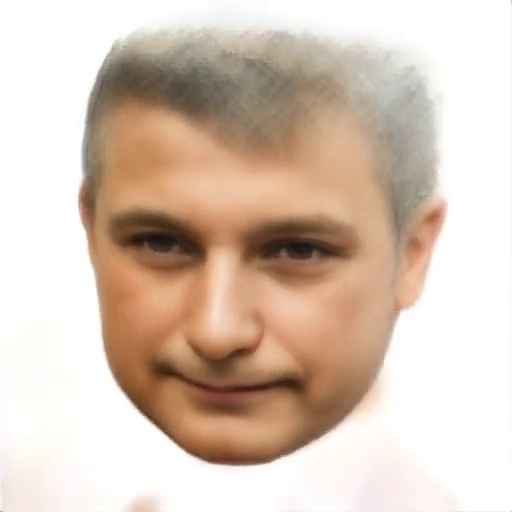} &
\includegraphics[width=0.1\textwidth]{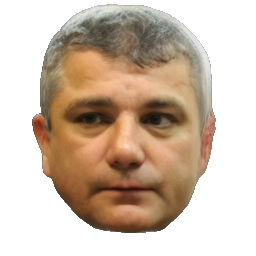} &
\includegraphics[width=0.1\textwidth]{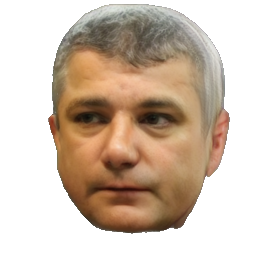} &
\includegraphics[width=0.1\textwidth]{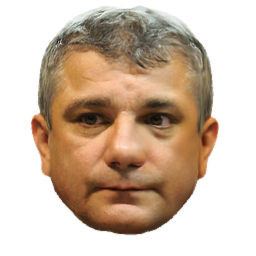} &
\includegraphics[width=0.1\textwidth]{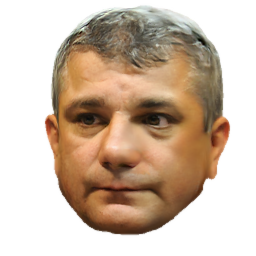} \\ \noalign{\vspace{-2pt}}
\includegraphics[width=0.1\textwidth]{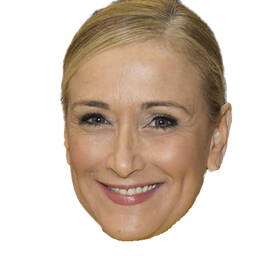} &
\includegraphics[width=0.1\textwidth]{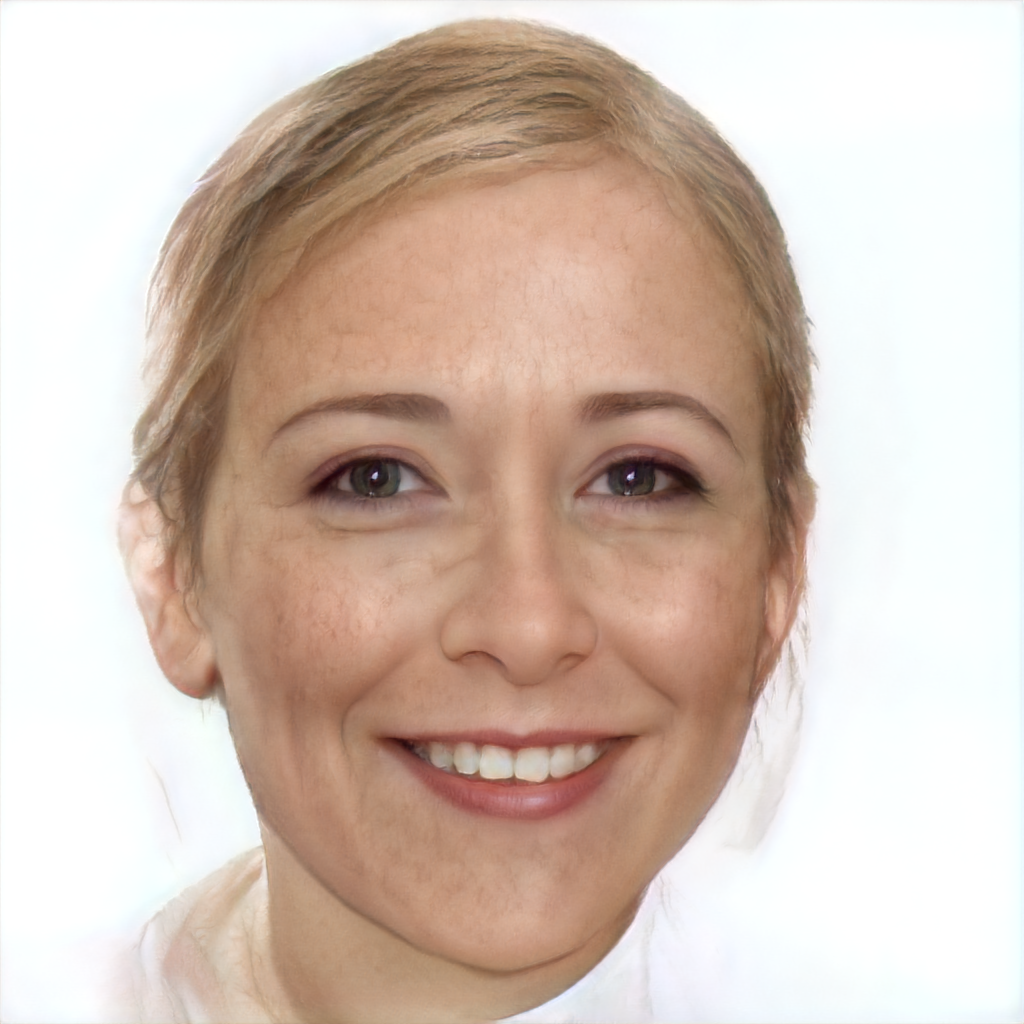} &
\includegraphics[width=0.1\textwidth]{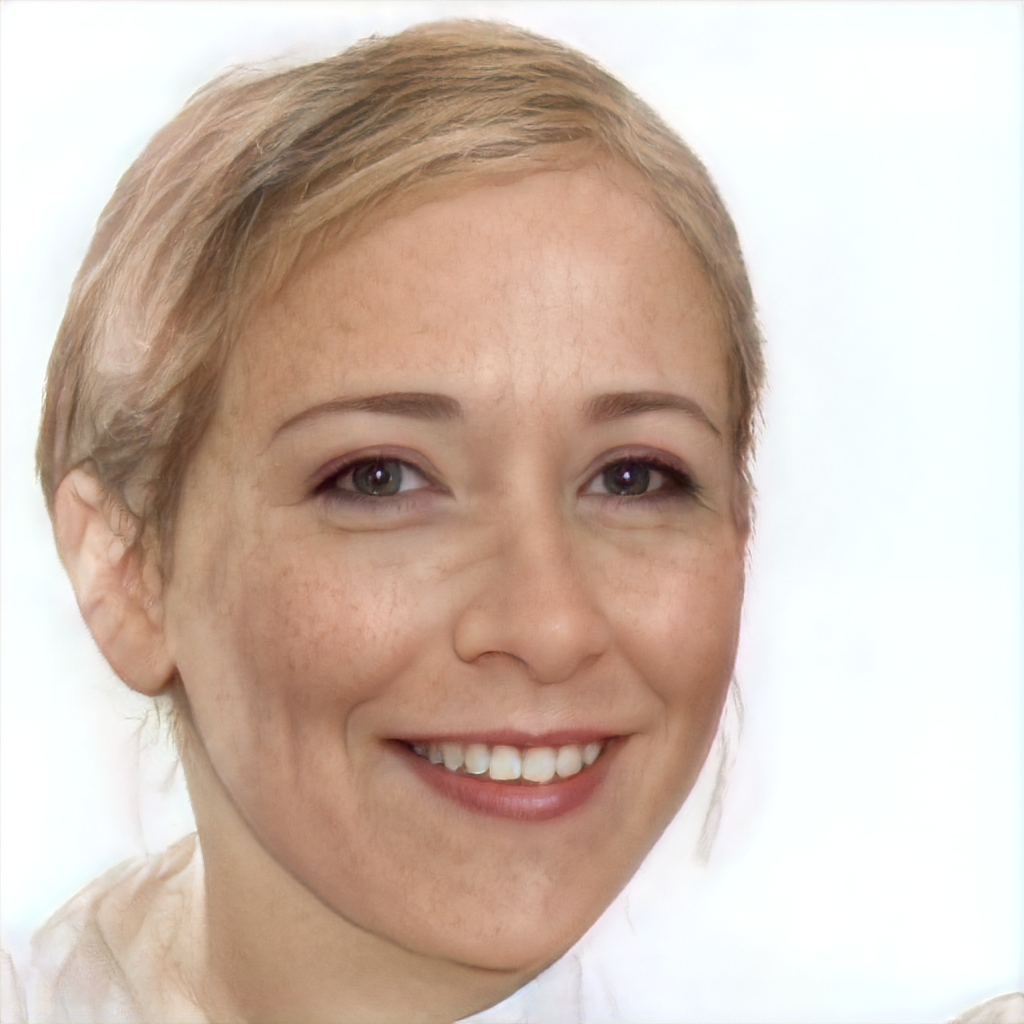} &
\includegraphics[width=0.1\textwidth]{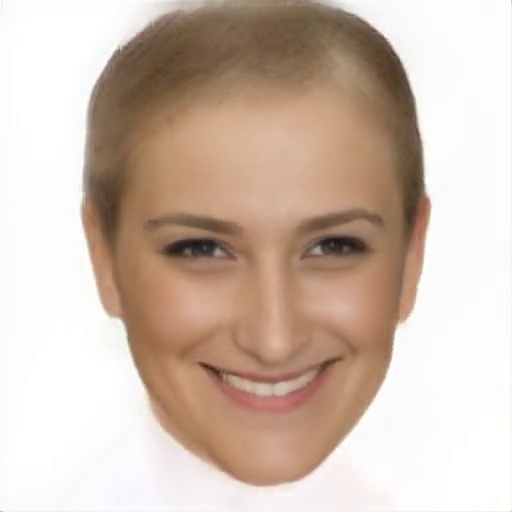} &
\includegraphics[width=0.1\textwidth]{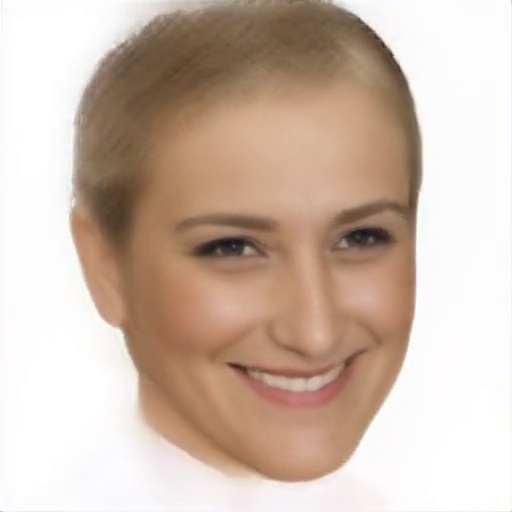} &
\includegraphics[width=0.1\textwidth]{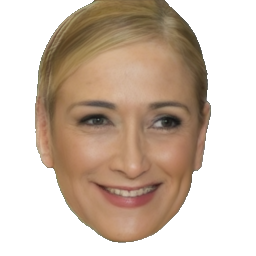} &
\includegraphics[width=0.1\textwidth]{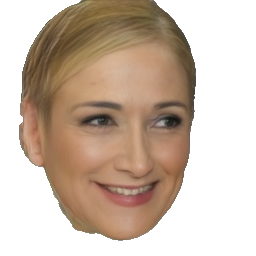} &
\includegraphics[width=0.1\textwidth]{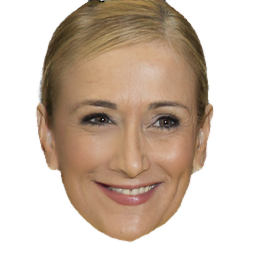} &
\includegraphics[width=0.1\textwidth]{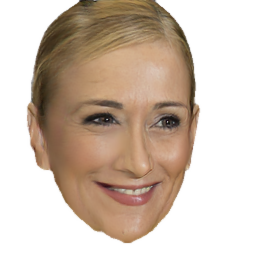} \\ \noalign{\vspace{-2pt}}
\includegraphics[width=0.1\textwidth]{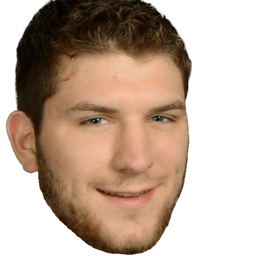} &
\includegraphics[width=0.1\textwidth]{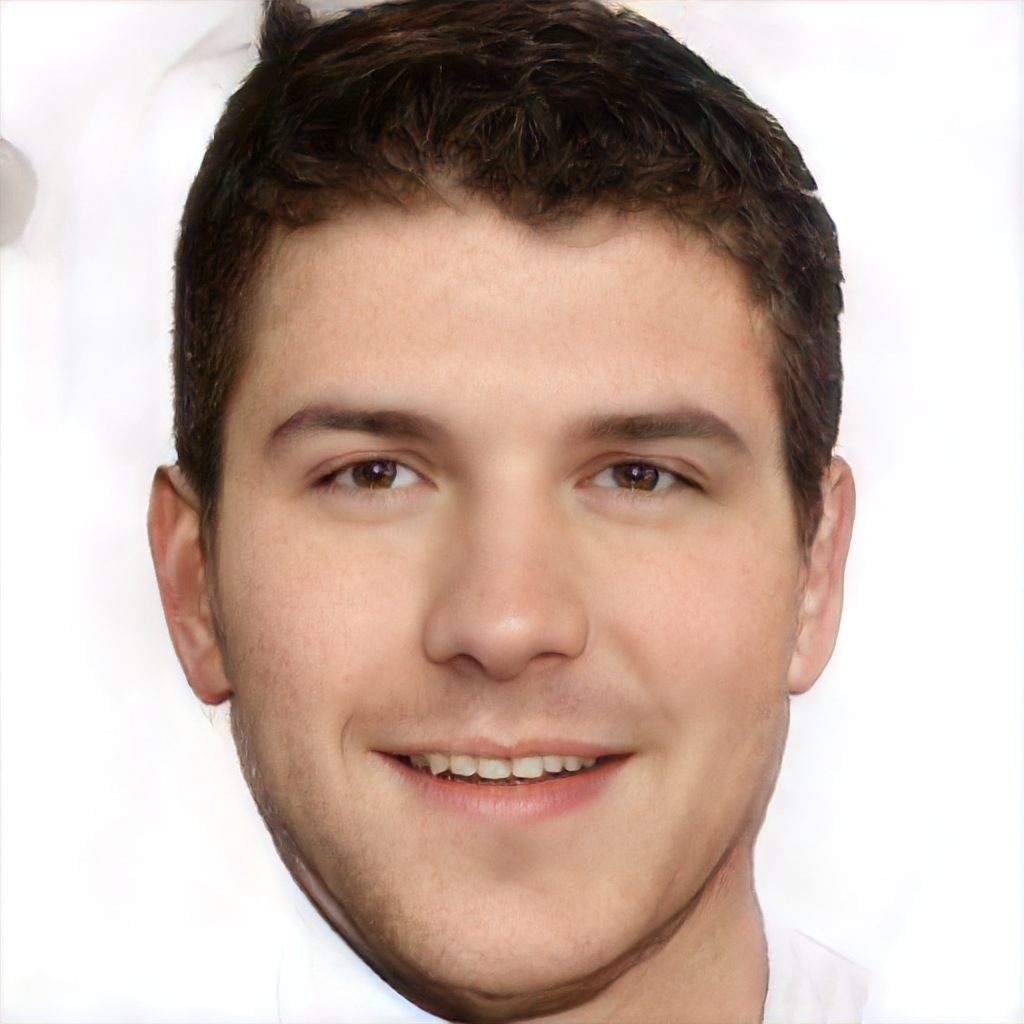} &
\includegraphics[width=0.1\textwidth]{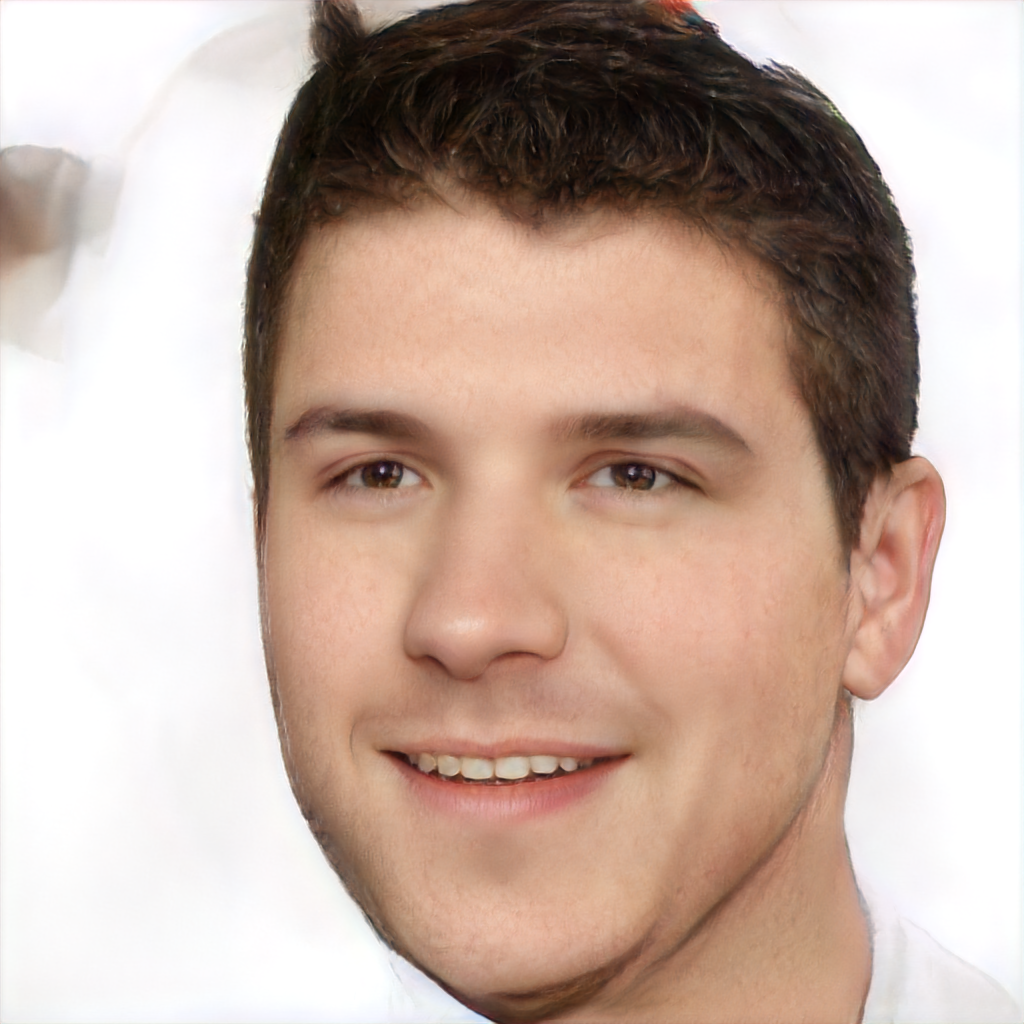} &
\includegraphics[width=0.1\textwidth]{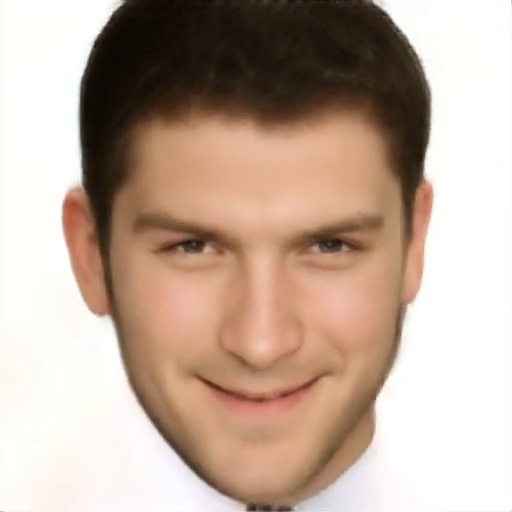} &
\includegraphics[width=0.1\textwidth]{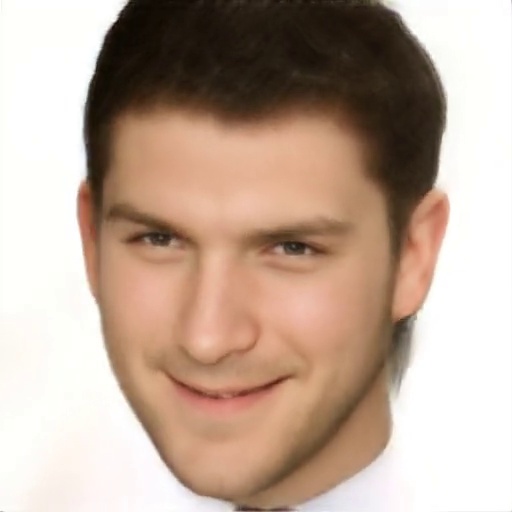} &
\includegraphics[width=0.1\textwidth]{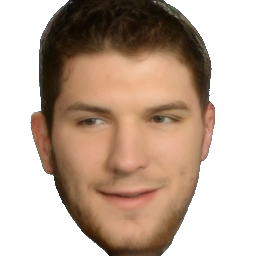} &
\includegraphics[width=0.1\textwidth]{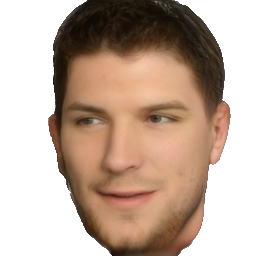} &
\includegraphics[width=0.1\textwidth]{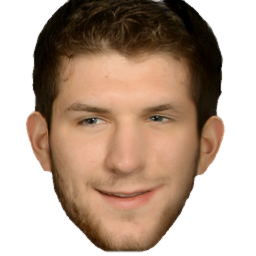} &
\includegraphics[width=0.1\textwidth]{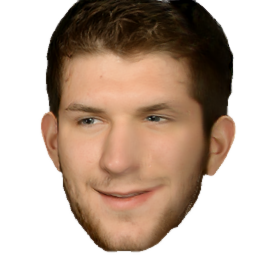} \\ \noalign{\vspace{-2pt}}
\includegraphics[width=0.1\textwidth]{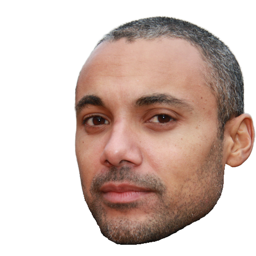} &
\includegraphics[width=0.1\textwidth]{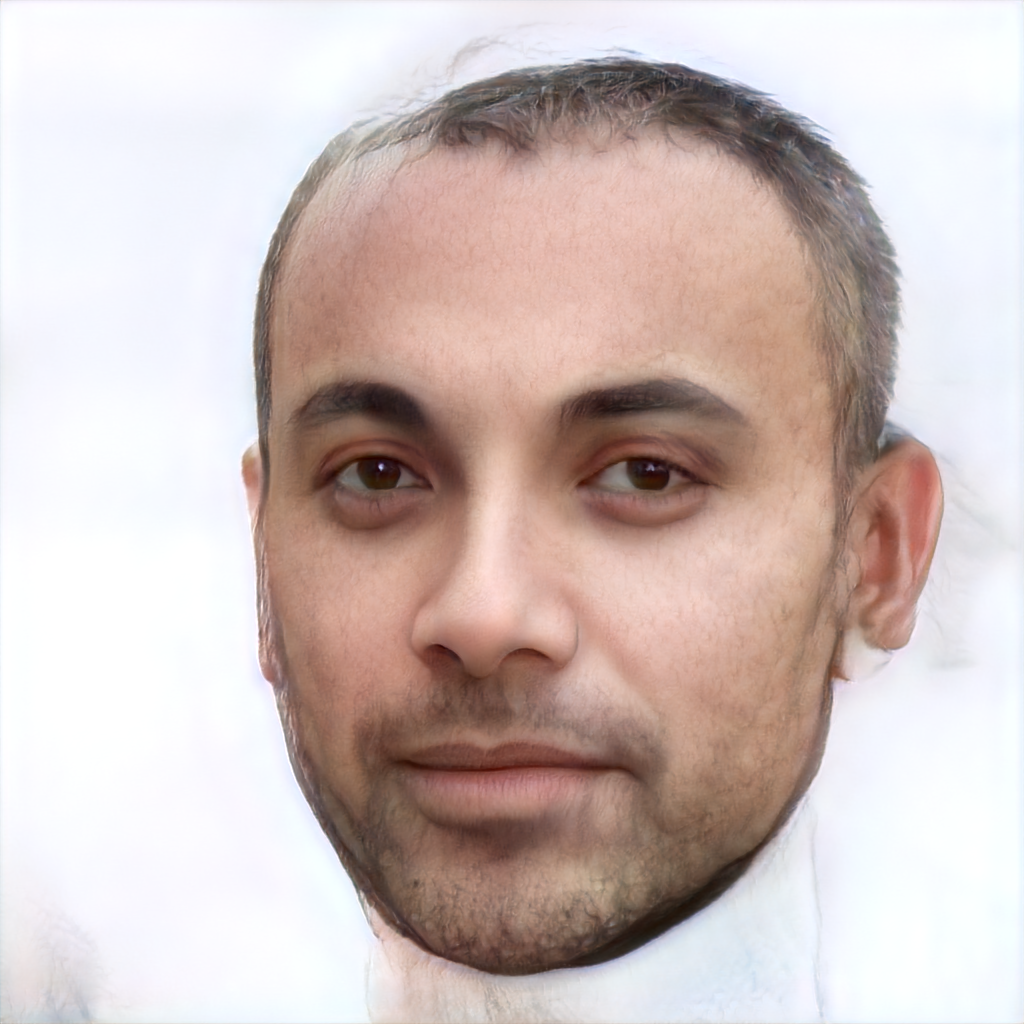} &
\includegraphics[width=0.1\textwidth]{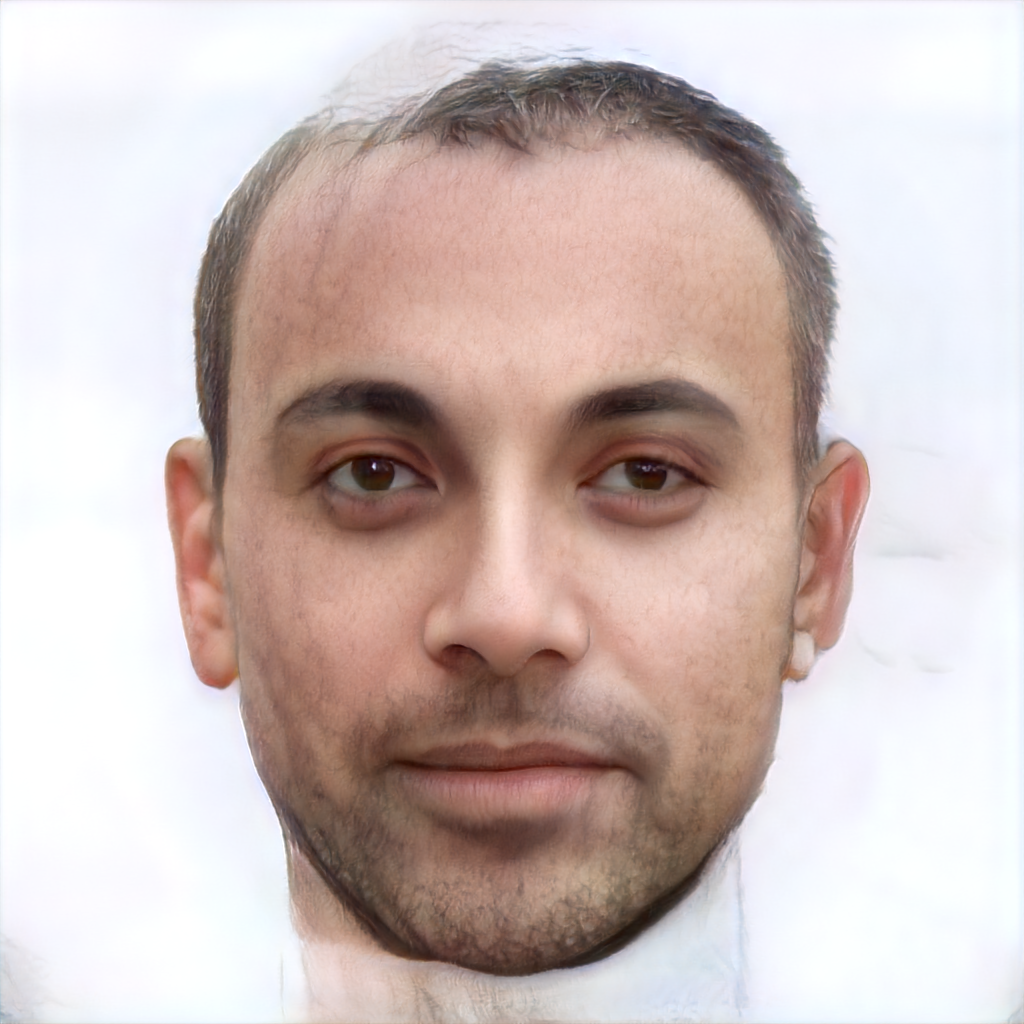} &
\includegraphics[width=0.1\textwidth]{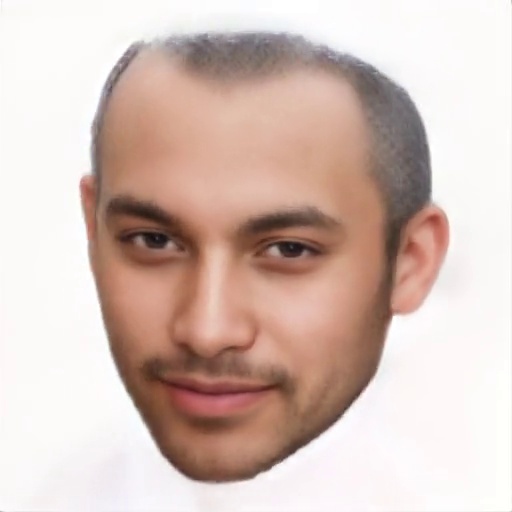} &
\includegraphics[width=0.1\textwidth]{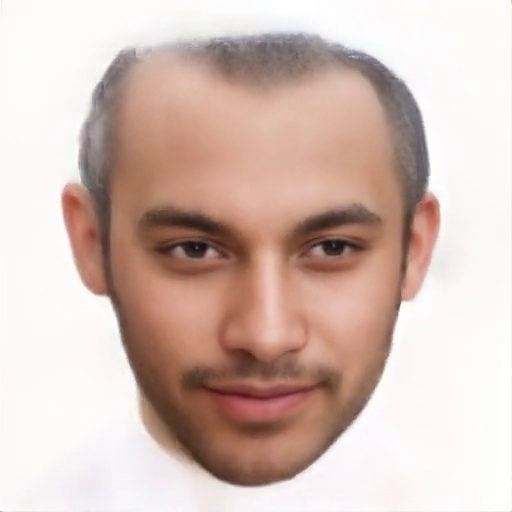} &
\includegraphics[width=0.1\textwidth]{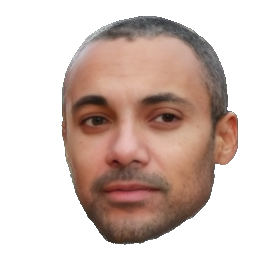} &
\includegraphics[width=0.1\textwidth]{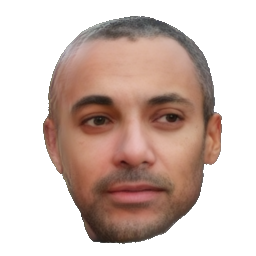} &
\includegraphics[width=0.1\textwidth]{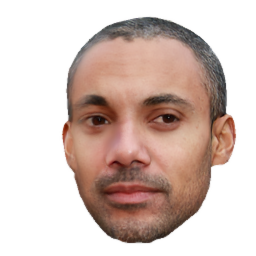} &
\includegraphics[width=0.1\textwidth]{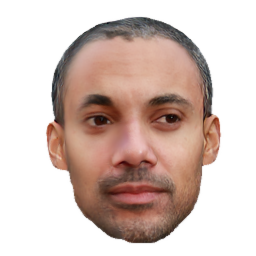} \\
Input & \multicolumn{2}{c}{E$3$DGE} & \multicolumn{2}{c}{G-NeRF} & \multicolumn{2}{c}{GAGAvatar} & \multicolumn{2}{c}{Ours} \\
\end{tabular}}
\caption{\small \textbf{Novel View Synthesis.} We qualitatively compare our novel view synthesis performance.}
\label{fig:nvs}
\end{figure*}

\begin{figure}[t]
\centering
\setlength{\tabcolsep}{0pt}
\renewcommand{\arraystretch}{0}
\resizebox{\linewidth}{!}{%
\begin{tabular}{c@{}c@{}c@{}c@{}c}
\includegraphics[width=0.2\linewidth]{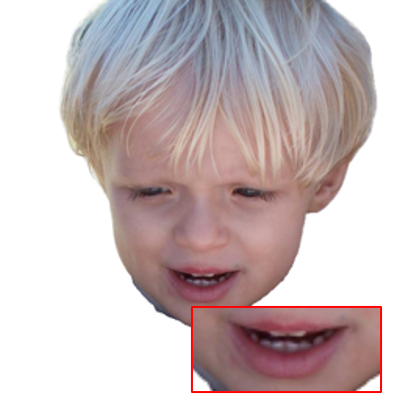}%
&\includegraphics[width=0.2\linewidth]{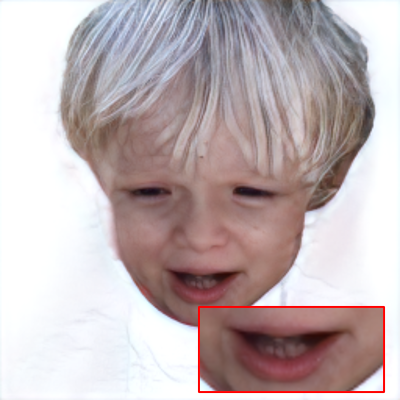}%
&\includegraphics[width=0.2\linewidth]{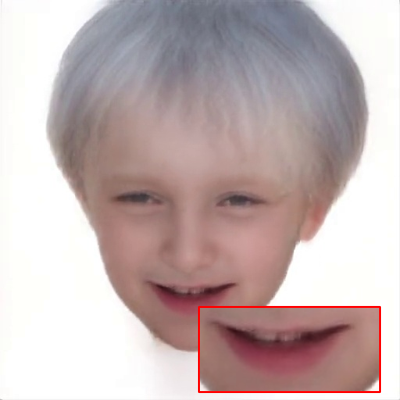}%
&\includegraphics[width=0.2\linewidth]{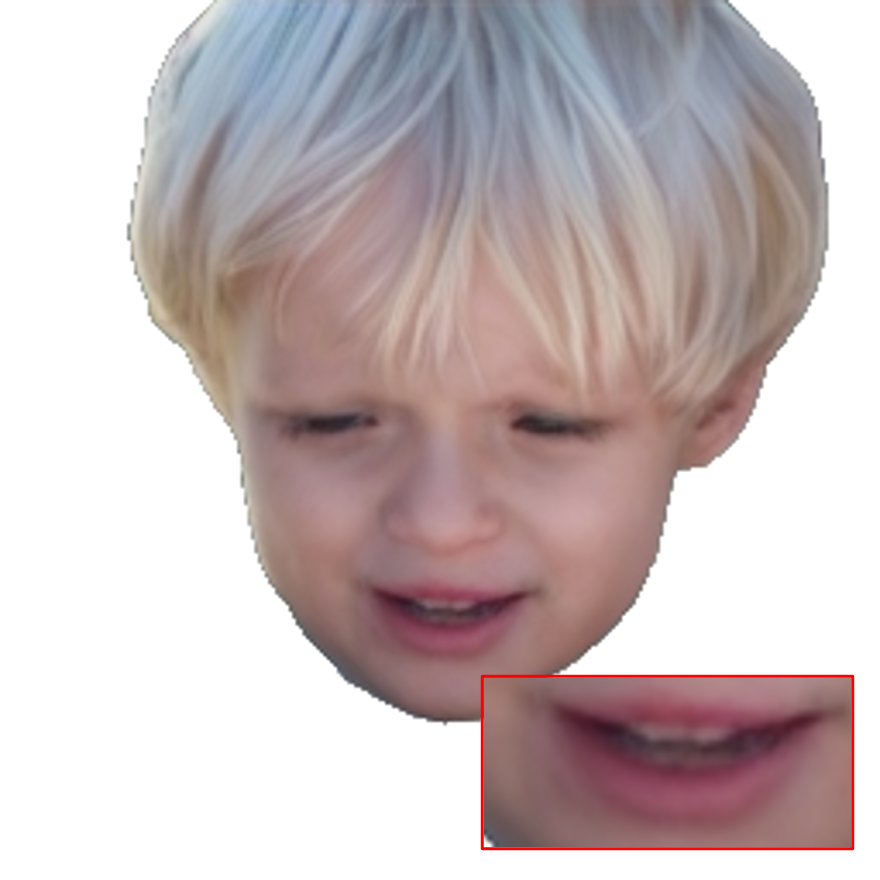}%
&\includegraphics[width=0.2\linewidth]{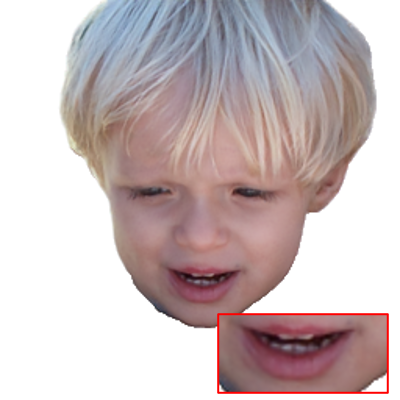}%
\\[2pt]
\includegraphics[width=0.2\linewidth]{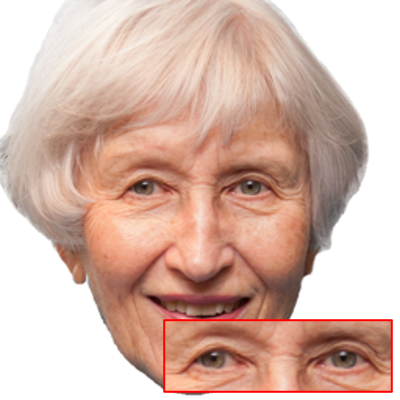}%
&\includegraphics[width=0.2\linewidth]{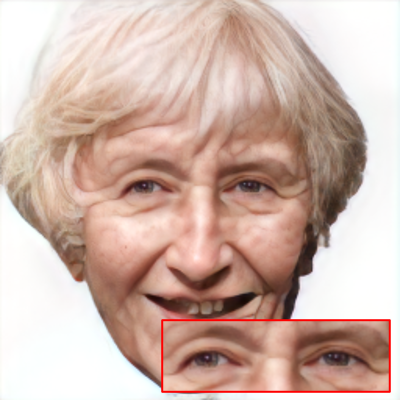}%
&\includegraphics[width=0.2\linewidth]{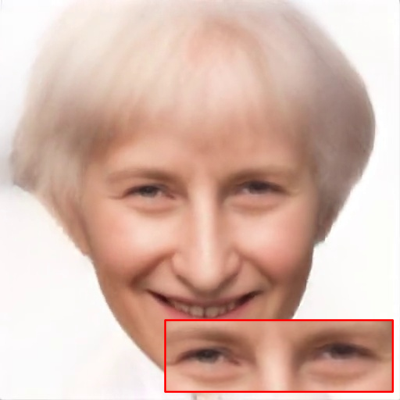}%
&\includegraphics[width=0.2\linewidth]{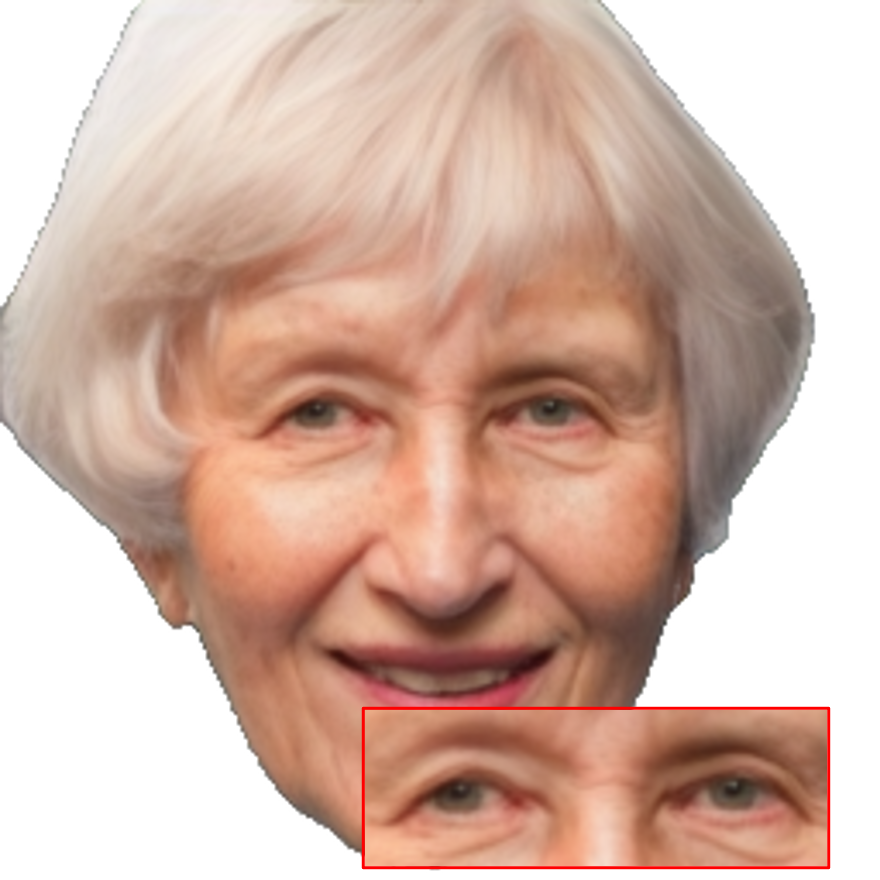}%
&\includegraphics[width=0.2\linewidth]{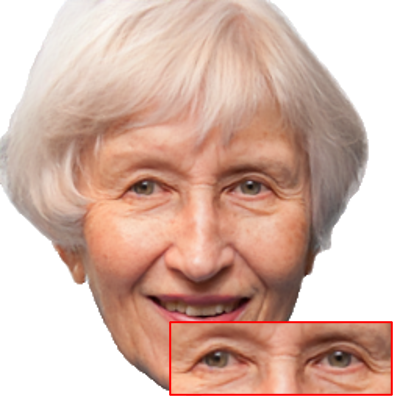}%
\\[3pt]
\small Input & \small (a) & \small (b) & \small (c) & \small (d)
\end{tabular}}
\caption{\small \textbf{Reconstruction Quality.} We compare the reconstruction quality on the FFHQ and CelebA-HQ datasets. (a) E$3$DGE; (b) G-NeRF; (c) GAGAvatar; (d) Ours.}
\label{fig:recons}
\end{figure}

\section{Experiments}
\label{sec:experiment}
\subsection{Experimental Setup}
\paragraph{Datasets and Baselines.}We train on $30,000$ FFHQ~\cite{FFHQ} images. For evaluation, we use a disjoint set of $1,000$ FFHQ images and $1,000$ CelebA-HQ~\cite{CelebA-HQ} images. To train our density prediction network, we generate pseudo ground-truth Gaussian center maps for all training images using our proposed GAM (Sec.~\ref{sec:gaussianallocation}).

For the baselines, we compare EGGHead’s Gaussian representation quality against single-image head reconstruction methods, including the 3D-aware GANs E3DGE~\cite{lan2022e3dge} and G-NeRF~\cite{GNeRF}, as well as the 3D-GS based method GAGAvatar~\cite{gagavatar}, using the same $M$=$12$\textit{k} Gaussians setting where applicable.

\vspace{-7mm}

\paragraph{Implementation Details.}
Our networks are trained on a single RTX A6000 GPU using PyTorch~\cite{Pytorch} and the AdamW~\cite{Loshchilov2017DecoupledWD} optimizer with an input image size of $256 \times 256$. We train the density prediction network for $100$ epochs and the Gaussian attribute regressor for $100$\textit{k} iterations. For depth estimation, we use Sapiens~\cite{Khirodkar2024SapiensFF}.

\begin{table}[t]
\centering
\resizebox{0.9\columnwidth}{!}{%
\setlength\tabcolsep{2pt}
\begin{tabular}{c c c c c c c}
\toprule
\multirow{2}{*}{Method} & \multicolumn{3}{c}{FFHQ} & \multicolumn{3}{c}{CelebA-HQ} \\
\cline{2-7}
 & PSNR$\uparrow$ & SSIM$\uparrow$ & LPIPS$\downarrow$ & PSNR$\uparrow$ & SSIM$\uparrow$ & LPIPS$\downarrow$ \\
\midrule
E$3$DGE~\cite{lan2022e3dge} & $23.43$ & $0.598$ & $0.133$ & $23.64$ & $0.589$ & $0.128$ \\
G-NeRF~\cite{GNeRF}         & $19.68$ & $0.673$ & $0.256$ & $19.81$ & $0.566$ & $0.296$ \\
GAGAvatar~\cite{gagavatar}  & $25.03$ & $0.888$ & $0.110$ & $25.06$ & $0.874$ & $0.119$ \\
Ours                        & $\mathbf{29.75}$ & $\mathbf{0.951}$ & $\mathbf{0.017}$ & $\mathbf{29.25}$ & $\mathbf{0.939}$ & $\mathbf{0.018}$ \\
\bottomrule
\end{tabular}}
\caption{\small
\textbf{Reconstruction Quality.} We quantitatively compare the reconstruction quality.
}
\label{tab:recons}
\vspace{-2mm}
\end{table}

\subsection{Reconstruction Quality}
We first evaluate the reconstruction quality of all methods to assess how well each model reproduces the source view. This provides a direct measure of representation power and fidelity before novel-view synthesis.
Specifically, we measure the reconstruction error using PSNR, SSIM~\cite{SSIM}, and LPIPS~\cite{LPIPS}. In this experiment, we set the number of Gaussians at $M$=$12$\textit{k} for all test images. As shown in Tab.~\ref{tab:recons}, our model achieves the best performance across all metrics.  We also qualitatively compare reconstruction quality in Fig.~\ref{fig:recons}. EGGHead captures the subject's skin tone more accurately and improves the detail of the hair, mouth, and teeth, thanks to our GAM assigning more Gaussians to those regions.

\begin{table}
\centering
\resizebox{0.9\columnwidth}{!}{%
\setlength\tabcolsep{8pt}
\begin{tabular}{c c c c c}
\toprule
\multirow{2}{*}{Method} & \multicolumn{2}{c}{FFHQ} & \multicolumn{2}{c}{CelebA-HQ} \\
\cline{2-5}
 & ID$\uparrow$ & FID$\downarrow$ & ID$\uparrow$ & FID$\downarrow$ \\
\midrule
E$3$DGE~\cite{lan2022e3dge} & $0.7744$ & $41.60$ & $0.7547$ & $40.87$ \\
G-NeRF~\cite{GNeRF}         & $0.6621$ & $54.67$ & $0.6375$ & $50.45$ \\
GAGAvatar~\cite{gagavatar}  & $0.8377$ & $32.55$ & $0.8480$ & $\mathbf{30.01}$ \\
Ours                        & $\mathbf{0.8517}$ & $\mathbf{29.01}$ & $\mathbf{0.8532}$ & $32.23$ \\
\bottomrule
\end{tabular}}
\caption{\small
\textbf{Novel View Synthesis.} We compare novel view synthesis quality using pairwise multi-view identity consistency (ID) and Frechet Inception Distance (FID).
}
\label{tab:nvs}
\vspace{-2mm}
\end{table}

\begin{figure}[t]
\centering
\setlength{\tabcolsep}{0pt}
\renewcommand{\arraystretch}{0}
\resizebox{\linewidth}{!}{%
\begin{tabular}{c@{}c@{}c@{}c@{}c@{}c}
\includegraphics[width=0.16\linewidth]{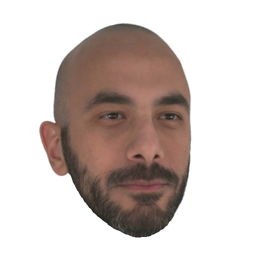}%
&\includegraphics[width=0.16\linewidth]{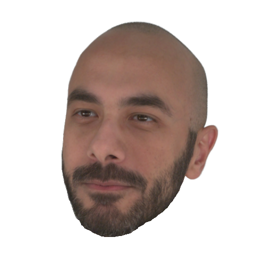}%
&\includegraphics[width=0.16\linewidth]{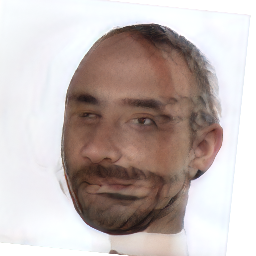}%
&\includegraphics[width=0.16\linewidth]{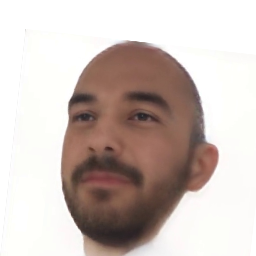}%
&\includegraphics[width=0.16\linewidth]{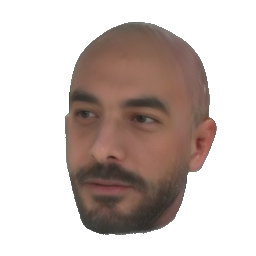}%
&\includegraphics[width=0.16\linewidth]{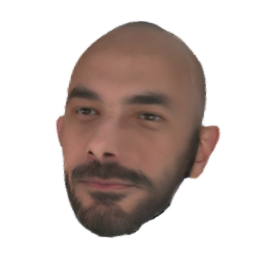}%
\\[2pt]
\includegraphics[width=0.16\linewidth]{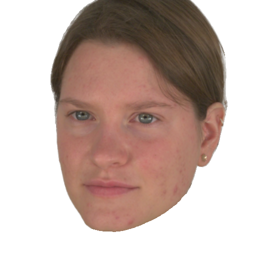}%
&\includegraphics[width=0.16\linewidth]{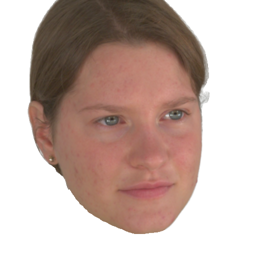}%
&\includegraphics[width=0.16\linewidth]{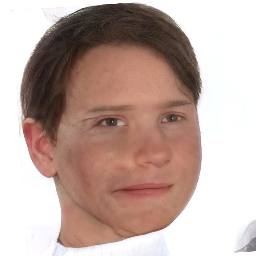}%
&\includegraphics[width=0.16\linewidth]{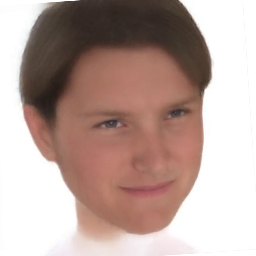}%
&\includegraphics[width=0.16\linewidth]{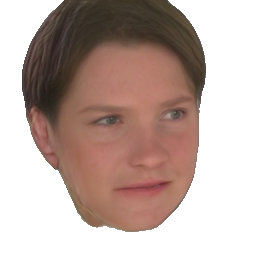}%
&\includegraphics[width=0.16\linewidth]{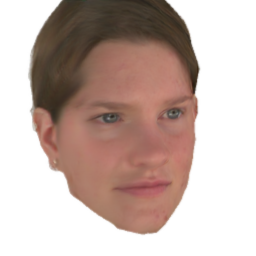}%
\\[3pt]
\small Input & \small GT & \small (a) & \small (b) & \small (c) & \small (d)
\end{tabular}}
\caption{\small \textbf{NeRSemble NVS.} (a) E$3$DGE; (b) G-NeRF; (c) GAGAvatar; (d) Ours.}
\label{fig:nersemble}
\vspace{-3mm}
\end{figure}

\subsection{Novel View Synthesis}
We evaluate EGGHead’s novel view synthesis by rendering subjects under unseen viewpoints. Fig.~\ref{fig:nvs} qualitatively compares our results with baselines. All methods produce visually plausible results. However, 3D-aware GANs~\cite{lan2022e3dge,GNeRF} struggle to maintain color fidelity, particularly in skin tones, hair, ears, and mouth regions, while GAGAvatar~\cite{gagavatar} suffers from identity inconsistencies due to weak constraints in occluded regions. In contrast, EGGHead preserves fine facial details through effective and efficient Gaussian allocation and better maintains subject identity, synthesizing novel views directly from subject-specific facial information without relying on additional networks.

We also quantitatively evaluate novel view synthesis in Tab.~\ref{tab:nvs}. Similar to~\cite{EG3D}, we compute pairwise identity consistency (ID) across novel views using AdaFace~\cite{kim2022adaface} and FID~\cite{FID} between the generated novel views and real FFHQ/CelebA-HQ images. EGGHead performs best on both metrics for FFHQ and achieves the best ID score for CelebA-HQ, demonstrating superior identity preservation, multi-view consistency, and overall face quality.

We further evaluate on the multi-view NeRSemble dataset~\cite{NeRSembleMR} to show the generalization capability of our approach. As shown in Fig.~\ref{fig:nersemble}, our method best preserves identity while synthesizing novel views, whereas the baselines lose fine details (e.g., facial hair, eyes, earrings). We also achieve the best performance across all metrics quantitatively for NeRSemble novel view synthesis (See Tab.~\ref{tab:nersemble_nvs}).

\begin{table}[t]
\centering
\resizebox{0.9\columnwidth}{!}{%
\setlength\tabcolsep{12pt}
\begin{tabular}{c c c c}
\toprule
Method & PSNR$\uparrow$ & SSIM$\uparrow$ & LPIPS$\downarrow$ \\
\midrule
E$3$DGE~\cite{lan2022e3dge} & $20.60$ & $0.777$ & $0.136$ \\
G-NeRF~\cite{GNeRF}         & $19.26$ & $0.768$ & $0.135$ \\
GAGAvatar~\cite{gagavatar}  & $21.43$ & $0.832$ & $0.101$ \\
Ours                        & $\mathbf{22.91}$ & $\mathbf{0.835}$ & $\mathbf{0.096}$ \\
\bottomrule
\end{tabular}}
\caption{\small
\textbf{NeRSemble NVS.} Quantitative comparison on the NeRSemble dataset.
}
\label{tab:nersemble_nvs}
\vspace{-5mm}
\end{table}

\begin{table}[t]
\centering
\resizebox{0.9\columnwidth}{!}{%
\setlength\tabcolsep{4pt}
\begin{tabular}{c c c c}
\toprule
Method & Enrollment time (sec) & Render speed (FPS)  \\
\midrule
E$3$DGE~\cite{lan2022e3dge} & $8.63$ & $5.26$ \\
G-NeRF~\cite{GNeRF} & $5.96$ & $33.61$ \\
GAGAvatar~\cite{gagavatar} & $\mathbf{0.06}$ & $43.43$ \\
Ours & $1.54$ & $\mathbf{121.19}$ \\
\bottomrule
\end{tabular}}
\caption{\textbf{Efficiency.} We compare efficiency in enrollment time and rendering speed.}
\label{tab:Efficiency}
\vspace{-4mm}
\end{table}

\subsection{Efficiency} We compare model efficiency on two fronts: enrollment time and rendering speed, measured on a single RTX A6000 GPU with AMD EPYC 7713 CPU. One key component in the efficiency of head representations is the latency required to prepare them before rendering. As shown in Tab.~\ref{tab:Efficiency}, 3D-aware GANs~\cite{lan2022e3dge,GNeRF} require several seconds, while ours reduces this cost to around one second. GAGAvatar~\cite{gagavatar} directly predicts Gaussian attributes during preparations and relies on a neural renderer in each frame, resulting in a short enrollment time. In contrast, our method uses slightly more computation during preparation to predict accurate Gaussian attributes and renders $3\times$ faster, achieving the fastest rendering speed.

\subsection{Ablation Studies}
We evaluate the effectiveness of our designs in more detail through ablations. All ablations are trained and evaluated on the aforementioned FFHQ dataset.

\vspace{-4mm}

\begin{figure}[t]
\centering
\newcommand{\figscale}{1.0}
\resizebox{\figscale\columnwidth}{!}{
\begin{tabular}{@{}ccccc@{}}
\includegraphics[width=0.19\linewidth]{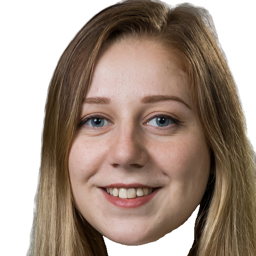} &
\includegraphics[width=0.19\linewidth]{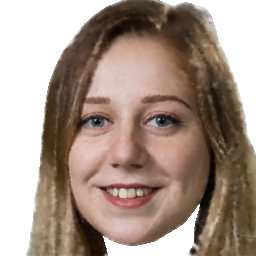} &
\includegraphics[width=0.19\linewidth]{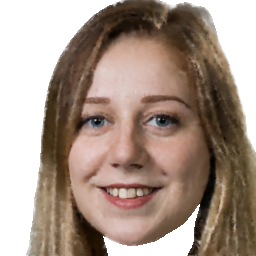} &
\includegraphics[width=0.19\linewidth]{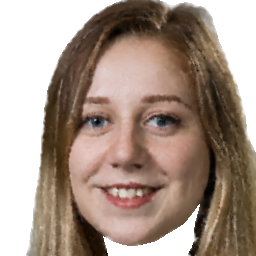} &
\includegraphics[width=0.19\linewidth]{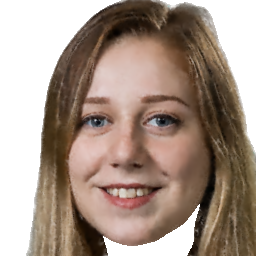} \\[3pt]
\footnotesize GT & \footnotesize (a) & \footnotesize (b) & \footnotesize (c) & \footnotesize (d)
\end{tabular}}
\vspace{-2.0mm}
\caption{\textbf{Ablation on Sampling Strategy.}
$M{=}12$\textit{k}. (a) Landmark; (b) Parsing; (c) Uniform; (d) Ours.}
\label{fig:ablation_sampling}
\vspace{-2.0mm}
\end{figure}

\paragraph{Sampling Strategy.}
We compare our allocation strategy (adaptive sampling) with three alternative sampling strategies (uniform, landmark, and parsing-guided) in reconstruction quality across Gaussian budgets $M$. Uniform sampling uniformly samples the head region, and landmark sampling allocates more samples around facial landmarks~\cite{dlib09}. Parsing-guided sampling leverages face parsing~\cite{yu2018bisenet} and allocates more Gaussians to targeted regions such as eyes and hair using fixed budgets for each semantic region. We evaluate under three settings of $M$: $12$\textit{k}, $8$\textit{k}, and $4$\textit{k}. As shown in Tab.~\ref{tab:ablation_sampling}, our adaptive sampling generally outperforms other strategies, with lower SSIM than uniform sampling. Fig.~\ref{fig:ablation_sampling} visualizes improvements in reconstruction quality. Our allocation algorithm captures fine details such as hair strands, eyes, and teeth more faithfully than uniform sampling. Although landmark sampling yields higher facial detail by allocating more samples around facial landmarks, it results in blurred regions elsewhere, such as hair. We also find that a fixed, deterministic budget for each facial semantic region cannot outperform our adaptive sampling that adjusts for each individual. Our method thus utilizes a limited Gaussian budget more effectively.

\begin{table}[t]
\centering
\resizebox{0.85\columnwidth}{!}{%
\setlength\tabcolsep{8pt}
\begin{tabular}{c c c c c}
\toprule
$M$ & Method & PSNR$\uparrow$ & SSIM$\uparrow$ & LPIPS$\downarrow$ \\
\midrule
\multirow{3}{*}{$12$\textit{k}}
& Landmark & 28.64 & 0.936 & 0.024 \\
& Parsing & 29.02 & 0.943 & 0.022 \\
& Uniform & 29.13 & \textbf{0.953} & 0.019 \\
& Adaptive (Ours) & \textbf{29.75} & 0.951 & \textbf{0.017} \\
\midrule
\multirow{3}{*}{$8$\textit{k}}
& Landmark & 28.06 & 0.918 & 0.032 \\
& Parsing & 28.35 & 0.925 & 0.030 \\
& Uniform & 28.43 & \textbf{0.938} & 0.026 \\
& Adaptive (Ours) & \textbf{29.27} & 0.937 & \textbf{0.021} \\
\midrule
\multirow{3}{*}{$4$\textit{k}}
& Landmark & 27.03 & 0.887 & 0.045 \\
& Parsing & 26.97 & 0.891 & 0.047 \\
& Uniform & 27.64 & \textbf{0.914} & 0.036 \\
& Adaptive (Ours) & \textbf{28.25} & 0.910 & \textbf{0.032} \\
\bottomrule
\end{tabular}}
\caption{\textbf{Ablation on Sampling Strategy.} 
We evaluate whether our allocation strategy (adaptive) more effectively distributes Gaussians under a fixed budget $M$ compared to uniform and landmark sampling. Our method achieves better reconstruction quality across $M$ settings, with lower SSIM than uniform sampling.}
\label{tab:ablation_sampling}
\vspace{-2mm}
\end{table}

\vspace{-4mm}

\begin{figure}
\centering
\setlength{\tabcolsep}{1pt}
\renewcommand{\arraystretch}{1.0}
\begin{tabular}{c c@{}c c@{}c}
\includegraphics[width=0.19\linewidth]{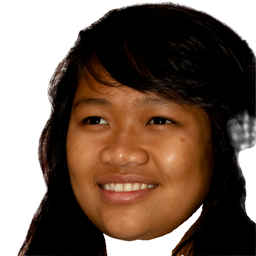} &
\includegraphics[width=0.19\linewidth]{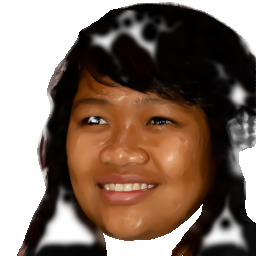} &
\includegraphics[width=0.19\linewidth]{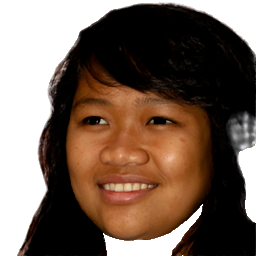} &
\includegraphics[width=0.19\linewidth]{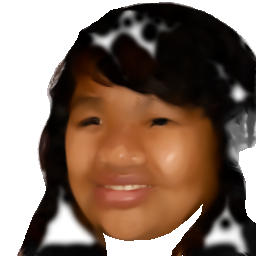} &
\includegraphics[width=0.19\linewidth]{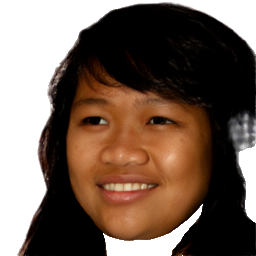} \\
\includegraphics[width=0.19\linewidth]{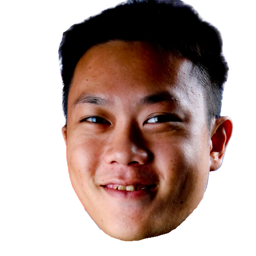} &
\includegraphics[width=0.19\linewidth]{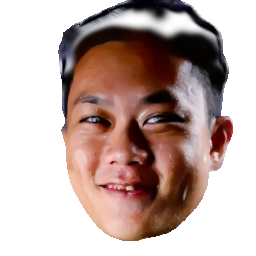} &
\includegraphics[width=0.19\linewidth]{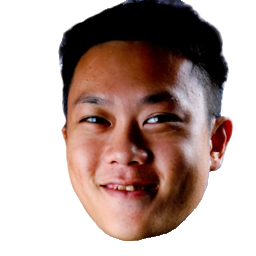} &
\includegraphics[width=0.19\linewidth]{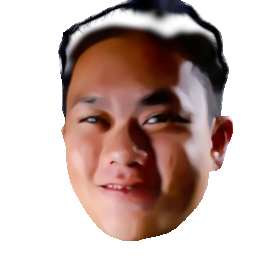} &
\includegraphics[width=0.19\linewidth]{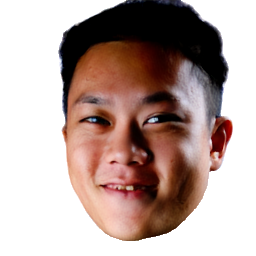} \\
Input & w/o VT & w/ VT & w/o VT & w/ VT \\
& \multicolumn{2}{c}{($M{=}12$\textit{k})} & \multicolumn{2}{c}{($M{=}4$\textit{k})} \\
\end{tabular}
\caption{\textbf{Ablation on Voronoi Tessellation.} We show the impact of Voronoi tessellation (VT) on capturing 3D Gaussian shapes under single-view and sparse-Gaussian settings.}
\label{fig:ablation_voronoi}
\vspace{-3mm}
\end{figure}

\paragraph{Voronoi Tessellation.}
We found accurately capturing 3D Gaussian shapes in single-view and sparse-Gaussian settings inherently non-trivial. Thus, we leverage Voronoi tessellation to guide the prediction of $\mathbf{M}_{q}$ and $\mathbf{M}_{s}$. To evaluate its influence, we remove this guidance and let the Gaussian attribute regressor directly predict $\mathbf{M}_{q}$ and $\mathbf{M}_{s}$ as well. As shown in Fig.~\ref{fig:ablation_voronoi}, with the same training sets, directly regressing Gaussian attributes without tessellation leads to noticeable hole artifacts in the rendered results. Moreover, under sparser Gaussians ($M{=}4$\textit{k}), scale attributes are overly enlarged, causing facial details to become blurred. Without tessellation, the model achieves a PSNR of $18.53$ and $18.14$ at $M{=}12$\textit{k} and $M{=}4$\textit{k}, respectively. See Adaptive (Ours) in Tab.~\ref{tab:ablation_sampling} for results with tessellation. This shows that tessellation greatly improves the robustness of the regressor in predicting Gaussian attributes.

\begin{figure}
\centering
\setlength{\tabcolsep}{1pt}
\renewcommand{\arraystretch}{1.0}
\begin{tabular}{c c@{}c c@{}c}
\includegraphics[width=0.19\linewidth]{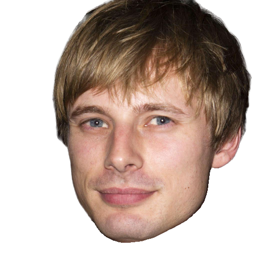} &
\includegraphics[width=0.19\linewidth]{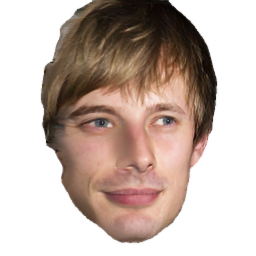} &
\includegraphics[width=0.19\linewidth]{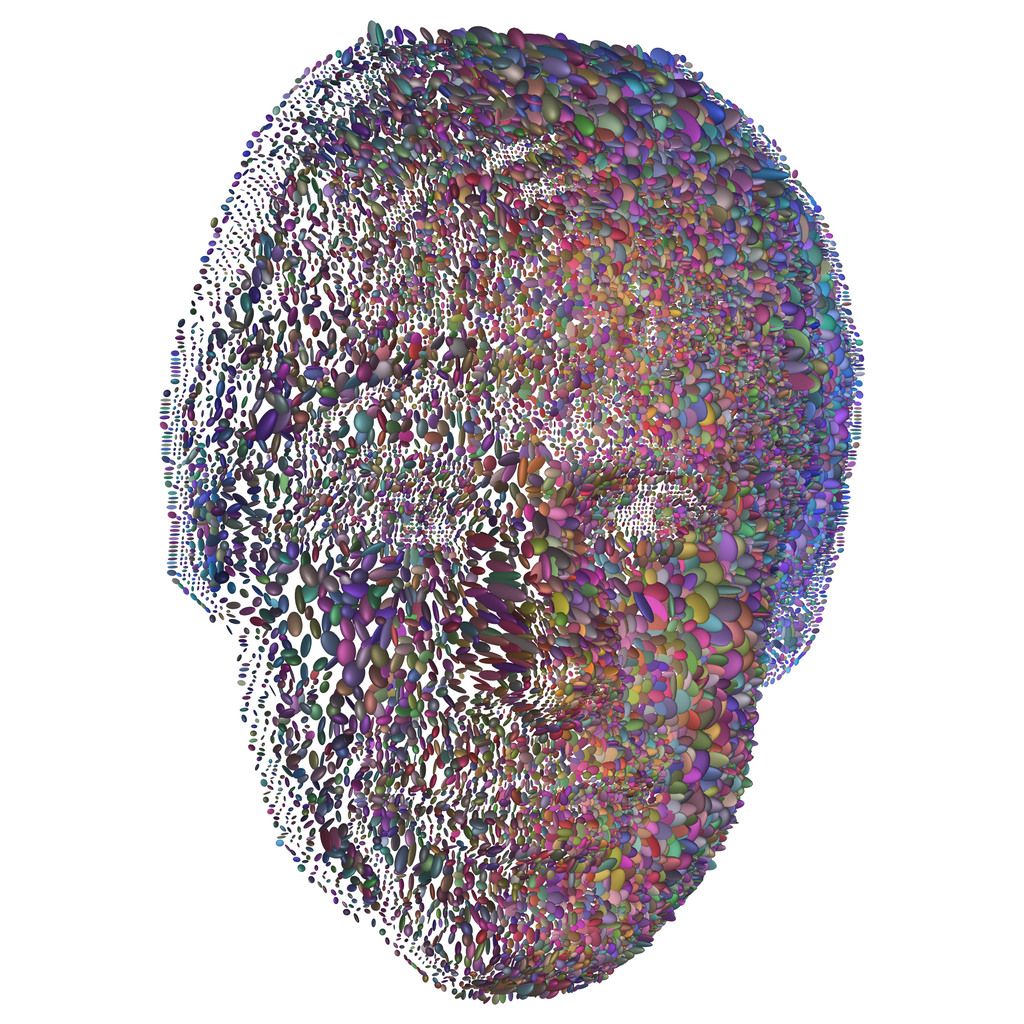} &
\includegraphics[width=0.19\linewidth]{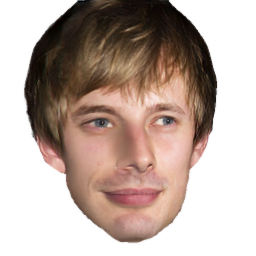} &
\includegraphics[width=0.19\linewidth]{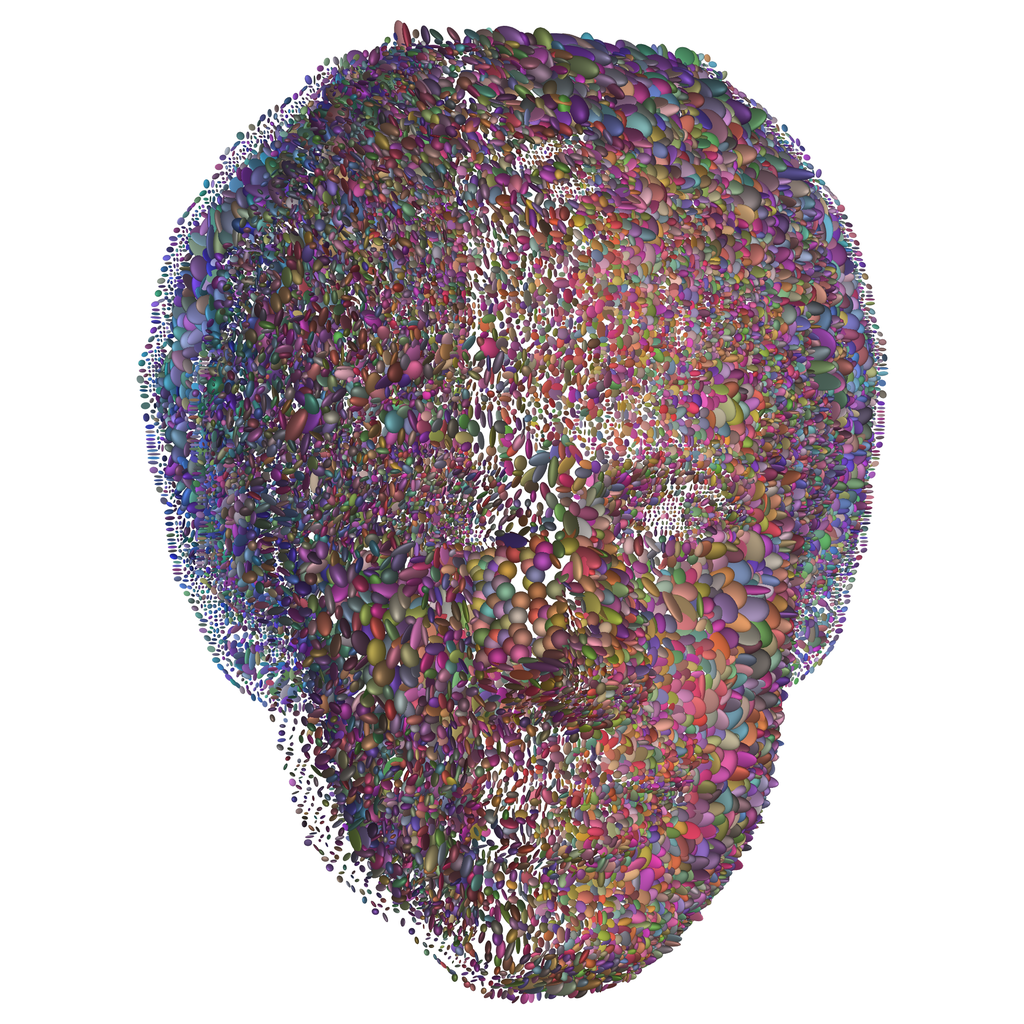} \\
\includegraphics[width=0.19\linewidth]{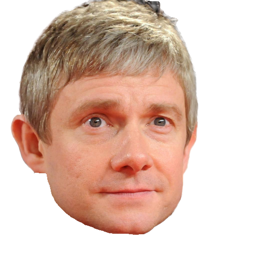} &
\includegraphics[width=0.19\linewidth]{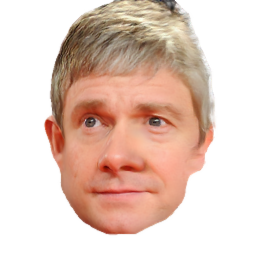} &
\includegraphics[width=0.19\linewidth]{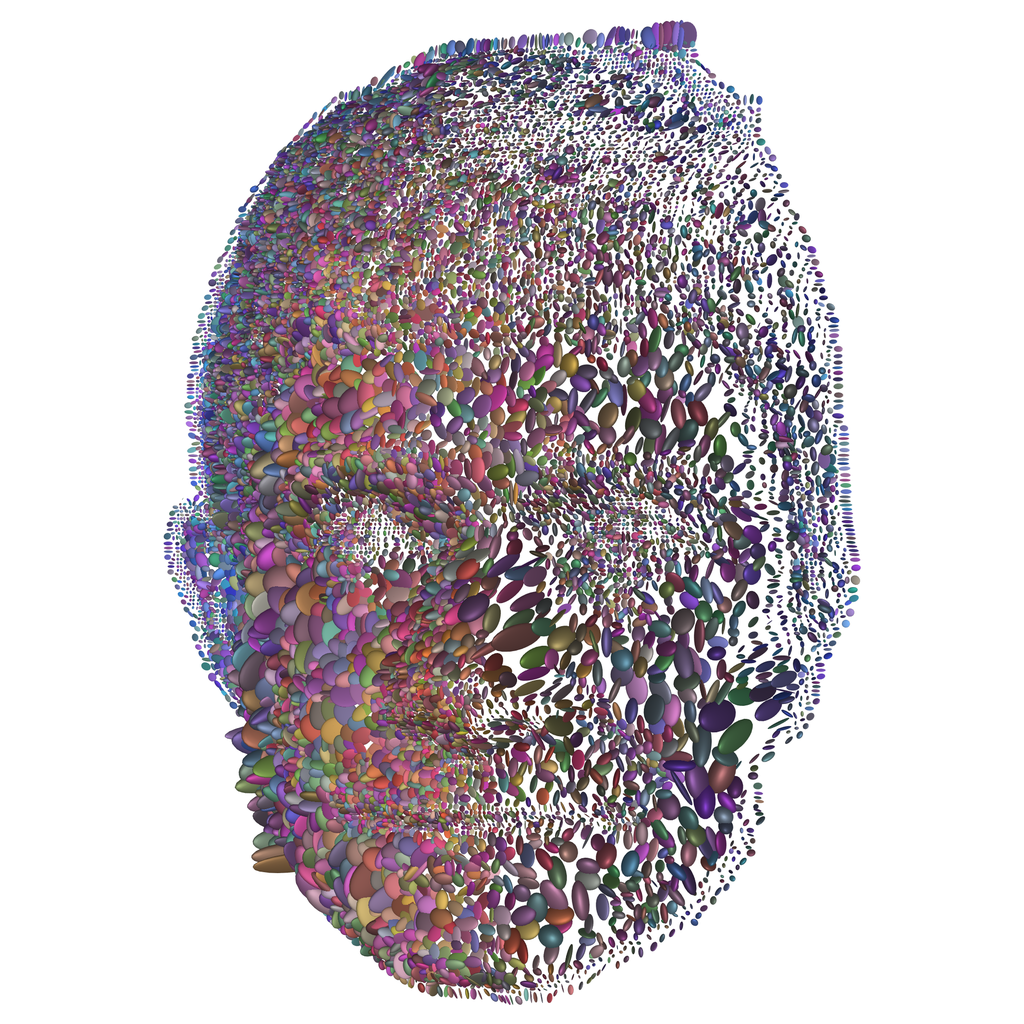} &
\includegraphics[width=0.19\linewidth]{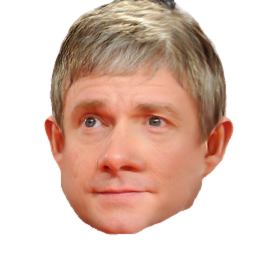} &
\includegraphics[width=0.19\linewidth]{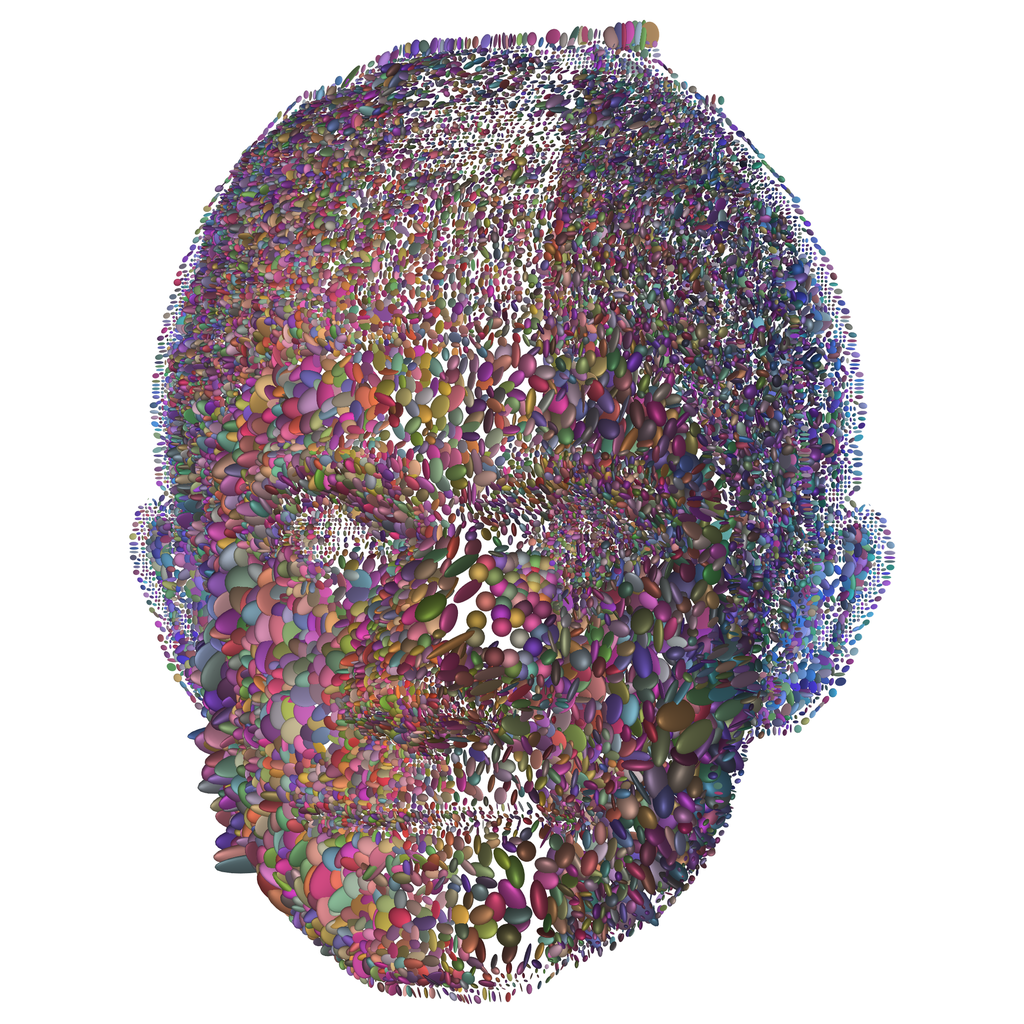} \\
Input &
\multicolumn{2}{c}{w/o mirroring} &
\multicolumn{2}{c}{w/ mirroring} \\
\end{tabular}
\caption{\textbf{Ablation on Mirroring Gaussians.} We show the effect of mirroring Gaussians for handling occluded regions. Our method naturally fills in the occluded areas.}
\label{fig:ablation_mirroring}
\vspace{-5mm}
\end{figure}

\vspace{-4mm}

\paragraph{Mirroring Gaussians.}
We fill occluded regions by mirroring Gaussians without relying on diffusion-based multi-view generation or an additional network to keep the model lightweight and efficient (see Fig.~\ref{fig:ablation_mirroring}). As the mirrored Gaussians are exclusively added to the occluded regions, our face symmetry module generates novel views without modifying the visible region. Despite its simplicity, it effectively improves completeness under pose variations while preserving the fidelity of visible regions. This design choice aligns with our objective of achieving reliable head reconstruction while minimizing computational overhead.
\vspace{-1mm}
\section{Conclusion}
We've proposed EGGHead, a single-image generalizable head modeling method that offers fast enrollment and rendering while maintaining high reconstruction quality. Our Gaussian allocation algorithm assigns appropriate numbers of Gaussians to different semantic head regions, preserving visual fidelity even with lower Gaussian budgets. Our feed-forward design combines a Gaussian allocation and density prediction network with a Gaussian attribute regressor, eliminating expensive per-subject optimization typically required in 3D-GS. In addition, Voronoi tessellation-guided attribute prediction improves structural accuracy under sparse single-view settings, while the Face Symmetry Module enhances completeness without additional heavy networks. We hope our method encourages future work on efficient, high-quality, and generalizable Gaussian heads from single images.
\vspace{-4mm}
\paragraph{Limitations.} Our model does not decompose texture into albedo and illumination and thus cannot support relighting. Moreover, pseudo-ground truth for Gaussian allocation is derived from single in-the-wild FFHQ images. Multi-view images of the same identity would provide more reliable supervision and higher-quality ground truth. Uneven lighting can also challenge the face symmetry assumption, causing incorrect reflection of shadows or overexposed regions under non-diffuse lighting. This can be mitigated by preprocessing the input image with an off-the-shelf delighting method~\cite{blind-removal-of-facial-foreign-shadows, he21unsupervised} during enrollment. Finally, fully frontal inputs may prevent complete reconstruction of the back of the head, as symmetry-based completion of occluded regions benefits from a partially visible face profile.

\clearpage
\setcounter{page}{1}
\maketitlesupplementary

\section{Network Architecture}
We describe the architectures of the Gaussian attribute regressor, the density prediction network, and the refinement network. Please refer to Tab.~\ref{tab:regressor}, Tab.~\ref{tab:density_net} and Tab.~\ref{tab:ref_net} respectively.

\paragraph{Gaussian Attribute Regressor.} The image encoder $\mathcal{E}_{img}$ and the attribute encoder $\mathcal{E}_{attr}$ share the same architecture, consisting of convolution, $\mathcal{E}_{1}$, $\mathcal{E}_{2}$, and $\mathcal{E}_{3}$. Image and depth features are concatenated and passed to the attribute decoder $\mathcal{D}_{attr}$ through skip connections. Each $\mathcal{E}$ and $\mathcal{D}$ stage consists of two residual blocks. ${h}_{s}$, ${h}_{\alpha}$, and ${h}_{c}$ denote the scale, opacity, and color heads, respectively, which operate on the decoded feature to predict the corresponding Gaussian attributes.

\begin{table}[h]
\centering
\resizebox{0.9\columnwidth}{!}{
\setlength\tabcolsep{8pt}
\begin{tabular}{c c c c c}
\toprule
Layer & Filter Size & Input Channel & Output Channel & Activation \\
\midrule
$\text{Conv}$     & $5 \times 5$ & $3$   & $32$  & ReLU \\
$\mathcal{E}_{1}$ & $3 \times 3$ & $32$  & $32$  & ReLU \\
$\mathcal{E}_{2}$ & $3 \times 3$ & $32$  & $48$  & ReLU \\
$\mathcal{E}_{3}$ & $3 \times 3$ & $48$  & $96$  & ReLU \\
$\mathcal{D}_{3}$ & $3 \times 3$ & $192$ & $96$  & ReLU \\
$\mathcal{D}_{2}$ & $3 \times 3$ & $192$ & $64$  & ReLU \\
$\mathcal{D}_{1}$ & $3 \times 3$ & $128$ & $48$  & ReLU \\
$\text{Conv}$     & $3 \times 3$ & $52$  & $32$  & ReLU \\
${h}_s$           & $3 \times 3$ & $32$  & $1$   & Sigmoid \\
${h}_\alpha$      & $3 \times 3$ & $32$  & $1$   & Sigmoid \\
${h}_c$           & $3 \times 3$ & $32$  & $3$   & Tanh \\
\bottomrule
\end{tabular}}
\caption{\textbf{Structure of Gaussian Attribute Regressor.}}
\label{tab:regressor}
\end{table}

\paragraph{Density Prediction Network.} The network follows a U-net-based encoder decoder design. The encoder $\mathcal{E}_{den}$ consists of four layers, $\mathcal{E}_{1}$--$\mathcal{E}_{4}$, which increase the number of channels from 3 to 256 while downsampling via max pooling. The decoder $\mathcal{D}_{den}$ comprises three layers, $\mathcal{D}_{1}$--$\mathcal{D}_{3}$, each upsampling with transposed convolutions and incorporating skip connections from the corresponding encoder layers. Each stage contains two residual blocks, and a final $1 \times 1$ convolution followed by Sigmoid generates the density map.

\begin{table}[h]
\centering
\resizebox{0.9\columnwidth}{!}{
\setlength\tabcolsep{8pt}
\begin{tabular}{c c c c c}
\toprule
Layer & Filter Size & Input channel & Output channel & Activation \\
\midrule
$\mathcal{E}_{1}$ & $3 \times 3$ & $3$   & $32$  & ReLU \\
$\mathcal{E}_{2}$ & $3 \times 3$ & $32$  & $64$  & ReLU \\
$\mathcal{E}_{3}$ & $3 \times 3$ & $64$  & $128$ & ReLU \\
$\mathcal{E}_{4}$ & $3 \times 3$ & $128$ & $256$ & ReLU \\
$\mathcal{D}_{3}$ & $2 \times 2$ & $256$ & $128$ & ReLU \\
$\mathcal{D}_{2}$ & $2 \times 2$ & $128$ & $64$  & ReLU \\
$\mathcal{D}_{1}$ & $2 \times 2$ & $64$  & $32$  & ReLU \\
Conv              & $1 \times 1$ & $32$  & $1$   & Sigmoid \\
\bottomrule
\end{tabular}}
\caption{\textbf{Structure of Density Prediction Network.}}
\label{tab:density_net}
\end{table}

\paragraph{Refinement Network.} The input is constructed by concatenating the coarse image $\mathbf{I}_{r}$ and the alpha map $\mathbf{I}_{\alpha}$, producing a 4-channel input. The encoder consists of four stages, $\mathcal{E}_{1}$--$\mathcal{E}_{4}$, which increase the channel dimension from 4 to 384 while reducing the spatial resolution.  Each stage contains two $3 \times 3$ convolutions followed by group normalization and SiLU activation. A bottleneck layer $\mathcal{M}$ further processes the latent feature. The decoder comprises three stages, $\mathcal{D}_{1}$--$\mathcal{D}_{3}$, each restoring the spatial resolution with skip connections from the encoder. A final $1 \times 1$ convolution predicts a 3-channel residual image, which is multiplied by $\mathbf{I}_{\alpha}$ and added to $\mathbf{I}_{r}$ to produce the refined output $\mathbf{I}_{r}'$.

\begin{table}[h]
\vspace{-1.0mm}
\centering
\resizebox{0.9\columnwidth}{!}{
\setlength\tabcolsep{8pt}
\begin{tabular}{c c c c c}
\toprule
Layer & Filter Size & Input channel & Output channel & Activation \\
\midrule
$\mathcal{E}_{1}$ & $3 \times 3$ & $4$   & $48$  & SiLU \\
$\mathcal{E}_{2}$ & $3 \times 3$ & $48$  & $96$  & SiLU \\
$\mathcal{E}_{3}$ & $3 \times 3$ & $96$  & $192$ & SiLU \\
$\mathcal{E}_{4}$ & $3 \times 3$ & $192$ & $384$ & SiLU \\
$\mathcal{M}$     & $3 \times 3$ & $384$ & $384$ & SiLU \\
$\mathcal{D}_{3}$ & $3 \times 3$ & $576$ & $192$ & SiLU \\
$\mathcal{D}_{2}$ & $3 \times 3$ & $288$ & $96$  & SiLU \\
$\mathcal{D}_{1}$ & $3 \times 3$ & $144$ & $48$  & SiLU \\
Conv              & $1 \times 1$ & $48$  & $3$   & - \\
\bottomrule
\end{tabular}}
\vspace{-1.0mm}
\caption{\textbf{Structure of Refinement Network.}}
\label{tab:ref_net}
\vspace{-5.0mm}
\end{table}

\section{Regressor Loss}
The regressor loss is defined as:
\begin{equation}
\mathcal{L}_{\text{reg}} = \lambda_{ref} \mathcal{L}_{ref} + \lambda_\text{coarse} \mathcal{L}_{\text{coarse}} + \lambda_\alpha \mathcal{L}_\alpha,
\label{eq:eq_reg1}
\end{equation}
which consists of the refinement loss $\mathcal{L}_{ref}$, the coarse loss $\mathcal{L}_{coarse}$ and the alpha loss $\mathcal{L}_{\alpha}$:
\begin{equation}
\begin{aligned}
\mathcal{L}_{\text{ref}}
&= \lambda_{\mathcal{L}_{1}}
   \mathcal{L}_{1}(\mathbf{I}_{r}', \mathbf{I}_{s})
 + \lambda_{\text{ssim}}
   \mathcal{L}_{\text{ssim}}(\mathbf{I}_{r}', \mathbf{I}_{s}) \\
&\quad + \lambda_{\text{perc}}
   \mathcal{L}_{\text{perc}}(\mathbf{I}_{r}', \mathbf{I}_{s}), \\[2pt]
\mathcal{L}_{\text{coarse}}
&= \lambda_{\mathcal{L}_{1}}
   \mathcal{L}_{1}(\mathbf{I}_{r}, \mathbf{I}_{s})
 + \lambda_{\text{ssim}}
   \mathcal{L}_{\text{ssim}}(\mathbf{I}_{r}, \mathbf{I}_{s}), \\[2pt]
\mathcal{L}_{\alpha}
&= \mathcal{L}_{\text{bce}}
   (\mathbf{I}_{\alpha}, \mathbf{I}_{\text{mask}}),
\end{aligned}
\label{eq:eq_reg2}
\end{equation}
where $\mathbf{I}_{\text{mask}}$ denotes the foreground mask extracted from the input image $\mathbf{I}_{s}$. We set $\lambda_{ref} = 0.7$, $\lambda_{coarse} = 0.3$, $\lambda_{\alpha} = 0.1$, 
$\lambda_{\mathcal{L}_1} = 0.8$, $\lambda_{\text{ssim}} = 0.2$ and $\lambda_{\text{perc}} = 0.05$.

\section{Edge-aware Alignment}
In addition to Gaussian merging, we incorporate an edge-aware alignment to ensure smoother transitions near internal boundaries in GAM. Specifically, we extract internal edges from the source image using the Canny edge detector~\cite{Canny1986ACA}. For each Gaussian center, we then compute its distance to the nearest internal edge and determine its symmetric position $\tilde{P}$ across the nearest edge.
\begin{equation}
\tilde{P} = 2 \cdot \mathrm{proj}(P) - P,
\label{eq:edge}
\end{equation}
where $\mathrm{proj}(P)$ denotes the projection of $P$ onto the nearest edge. We pair $P$ and $\tilde{P}$ across the boundary while preserving the total budget $M$. This alignment produces smoother transitions along the edges and preserves overall rendering consistency while reducing discontinuities.

\begin{figure}[h]
  \centering
  \resizebox{0.9\columnwidth}{!}{
  \setlength{\tabcolsep}{0.5pt}
  \renewcommand{\arraystretch}{1.0}
  \begin{tabular}{c c c c}
  \includegraphics[width=0.24\columnwidth]{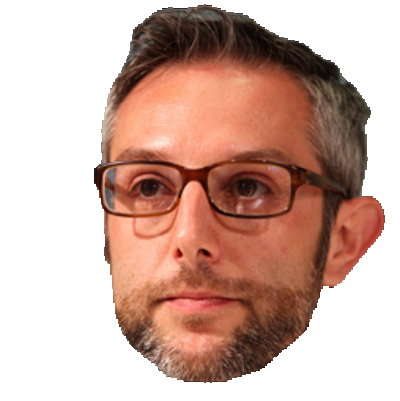} &
  \includegraphics[width=0.24\columnwidth]{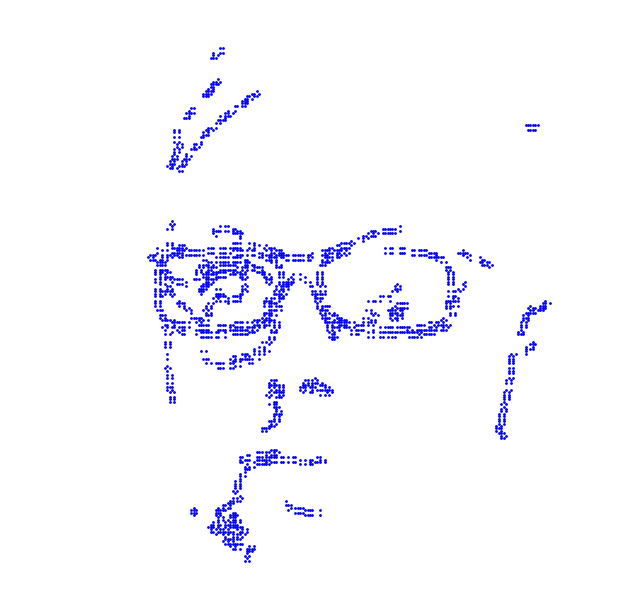} &
  \includegraphics[width=0.24\columnwidth]{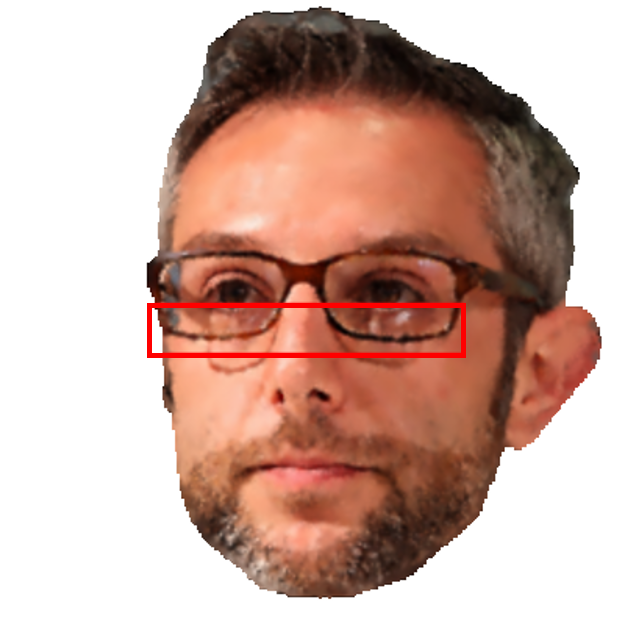} &
  \includegraphics[width=0.24\columnwidth]{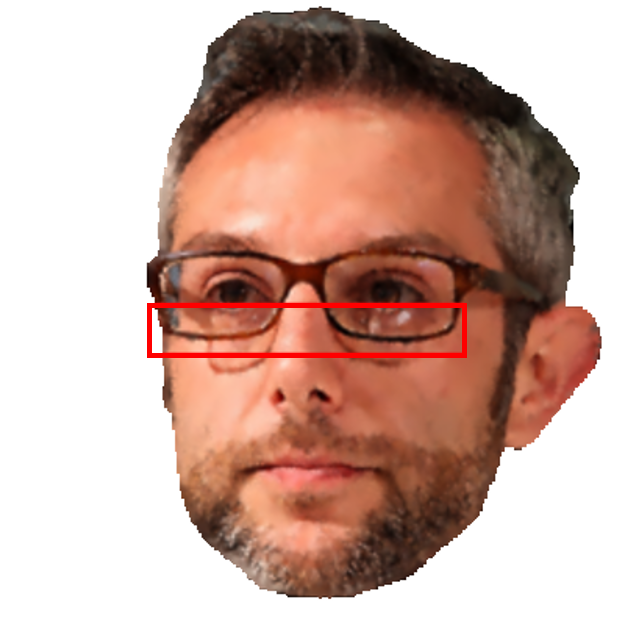} \\
  \footnotesize Input & \footnotesize Aligned positions & \footnotesize w/o Alignment & \footnotesize w/ Alignment \\
  \end{tabular}
  }
  \caption{
    \textbf{Edge-aware alignment.}
    Aligning Gaussian positions along the edges improves boundary consistency and reduces artifacts compared to the unaligned configuration.
  }
  \label{fig:edge}
\vspace{-5mm}
\end{figure}

\section{Density Prediction}
\begin{figure}[h]
\centering
\resizebox{0.9\columnwidth}{!}{
\setlength{\tabcolsep}{0.5pt}
\renewcommand{\arraystretch}{1.0}
\begin{tabular}{c c c c}
\includegraphics[width=0.24\columnwidth]{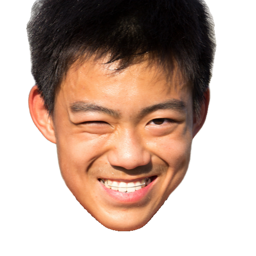} &
\includegraphics[width=0.24\columnwidth]{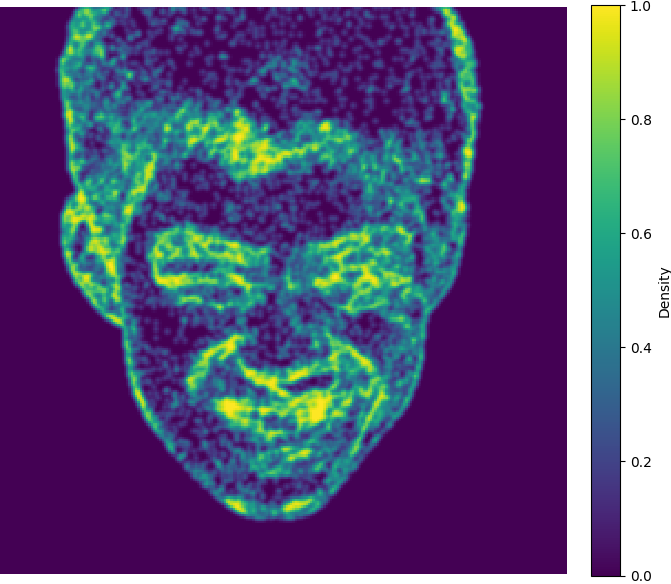} &
\includegraphics[width=0.24\columnwidth]{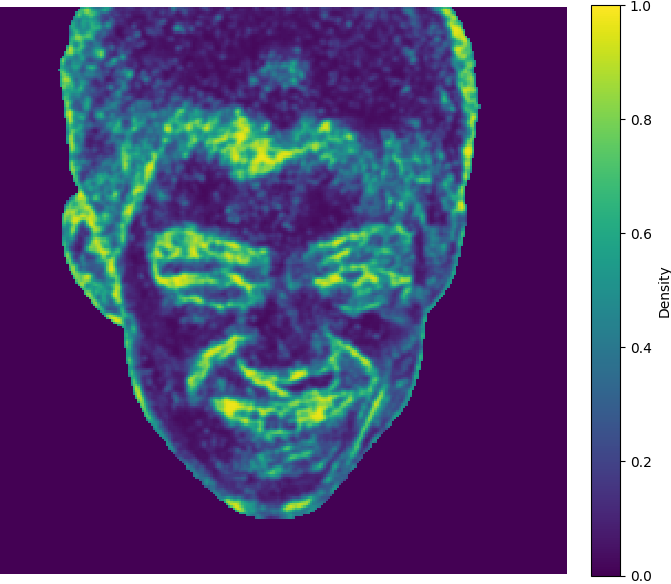} &
\includegraphics[width=0.24\columnwidth]{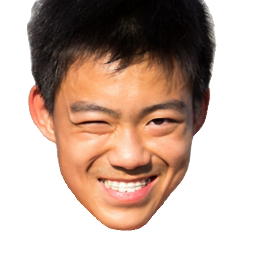} \\

\includegraphics[width=0.24\columnwidth]{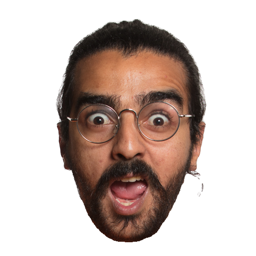} &
\includegraphics[width=0.24\columnwidth]{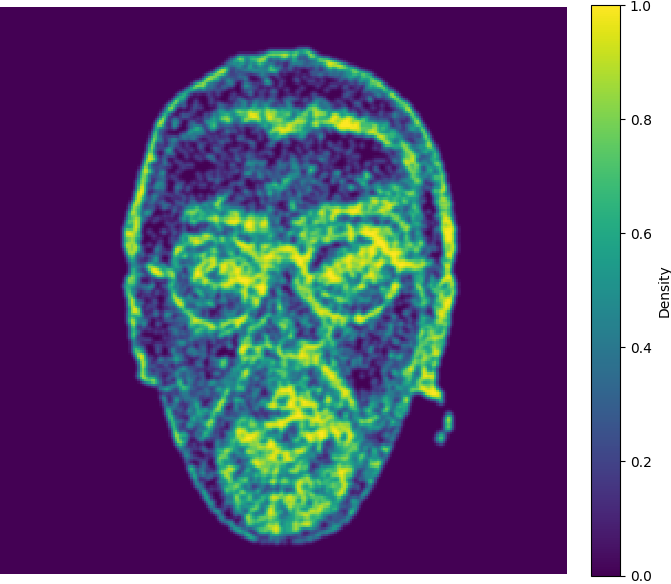} &
\includegraphics[width=0.24\columnwidth]{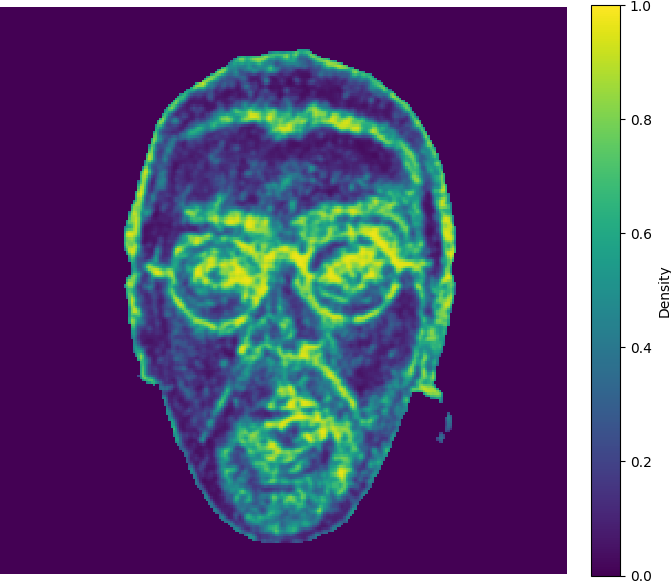} &
\includegraphics[width=0.24\columnwidth]{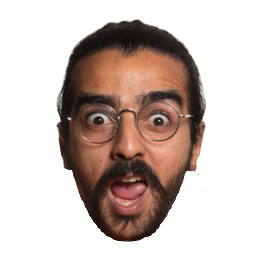} \\
Input & GT &  Predicted &  Rendered \\
\end{tabular}
}
\caption{
\textbf{Predicted Density Map.}
The predicted density maps align well with the GT density maps and assign higher density to fine detail regions which enables adaptive Gaussian allocation.
}
\label{fig:density}
\vspace{-5mm}
\end{figure}

We visualize the predicted density maps that are used to sample Gaussian positions. As shown in Fig.~\ref{fig:density}, the predicted density maps are generally consistent with the GT density maps and assign higher density to detail-sensitive regions, indicating that the density prediction network learns an effective density distribution for adaptive Gaussian allocation. Notably, EGGHead's density prediction network also handles more extreme expressions well such as ``smiling with teeth" and ``surprised with mouth open".

\section{Rendering Speed}
The 121.19 FPS reported in the main paper includes the refinement network and head mask generation~\cite{yu2018bisenet} after splatting, and these steps make the final FPS nearly constant regardless of the number of Gaussians. To show the actual rendering efficiency, we measure pure rendering speed without these steps and further evaluate larger Gaussian budgets of $120$\textit{k}, $80$\textit{k}, and $40$\textit{k}, as summarized in Tab.~\ref{tab:render_speed}. Even with $120$\textit{k} Gaussians, the pure rendering speed reaches 179.91 FPS, exceeding the 121.19 FPS of the full pipeline with $M$=$12$\textit{k}. This shows that EGGHead maintains real-time rendering performance even with larger Gaussian budgets.

\begin{table}[h]
\centering
\resizebox{0.9\columnwidth}{!}{
\setlength\tabcolsep{8pt}
\begin{tabular}{c c}
\toprule
Number of Gaussians (\textit{M}) & Rendering speed (FPS) \\
\midrule
$120$\textit{k} & 179.91 \\
$80$\textit{k}  & 214.01 \\
$40$\textit{k}  & 240.24 \\
$12$\textit{k}  & 309.60 \\
$8$\textit{k}   & 316.62 \\
$4$\textit{k}   & 331.11 \\
\bottomrule
\end{tabular}}
\caption{
\textbf{Rendering Speed.} We report rendering speed for different numbers of Gaussians without additional refinement steps.
}
\label{tab:render_speed}
\vspace{-2mm}
\end{table}

\section{Expression Editing}
Although our framework is trained for source-view reconstruction with static Gaussians, it enables expression editing at inference time by deforming Gaussians bound to the FLAME mesh~\cite{FLAME}. Specifically, each Gaussian center is bound to its closest point on the source FLAME mesh and deformed using the displacement between the source and driving meshes. Compared to GAGAvatar~\cite{gagavatar}, EGGHead is able to match the expression in the driving image more closely. For example, in the second driving image shown below, notice that GAGAvatar's result still has an open mouth whereas EGGHead is able to appropriately close the lips. GAGAvatar also modifies the gaze direction incorrectly (See the first driving image). In terms of efficiency, we find that editing expressions has very little impact on the rendering speed ($121.19$ FPS $\rightarrow$ $119.21$ FPS), and EGGHead remains real-time. The enrollment time increases from $1.54$ seconds to $2.14$ seconds due to binding and fitting Gaussians to the source mesh, but it remains shorter than those of E3DGE~\cite{lan2022e3dge} and G-NeRF~\cite{GNeRF}.

\begin{figure}[h]
\centering
\resizebox{0.9\columnwidth}{!}{
\setlength{\tabcolsep}{0.5pt}
\renewcommand{\arraystretch}{1.0}
\begin{tabular}{c c c c}
\includegraphics[width=0.24\columnwidth]{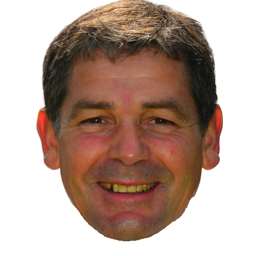} &
\includegraphics[width=0.24\columnwidth]{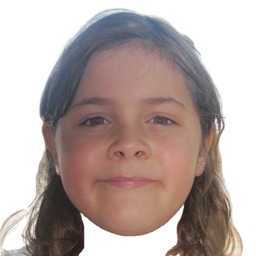} &
\includegraphics[width=0.24\columnwidth]{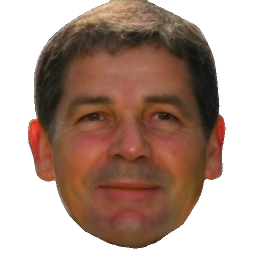} &
\includegraphics[width=0.24\columnwidth]{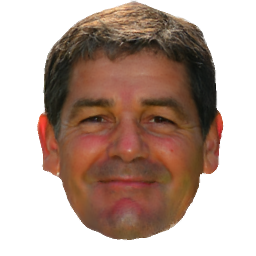} \\

\includegraphics[width=0.24\columnwidth]{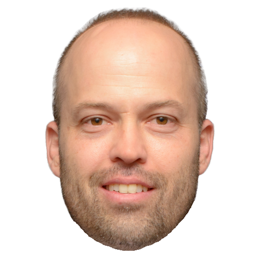} &
\includegraphics[width=0.24\columnwidth]{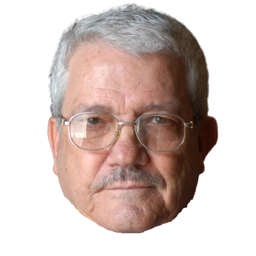} &
\includegraphics[width=0.24\columnwidth]{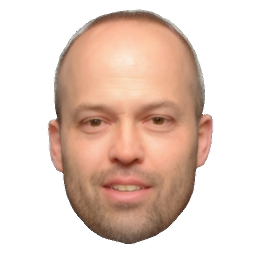} &
\includegraphics[width=0.24\columnwidth]{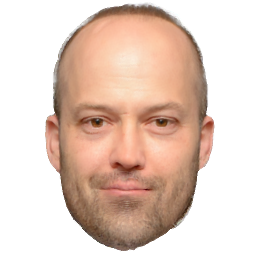} \\
Source & Driving & GAGAvatar & Ours \\
\end{tabular}
}
\caption{\textbf{Expression Editing.} We show the expression from the driving image is properly transferred to the source identity by deforming Gaussians bound to the mesh.}
\label{fig:exp_edit}
\end{figure}

\clearpage
\begin{figure*}[!p]
\centering
\setlength{\tabcolsep}{1pt}
\renewcommand{\arraystretch}{1.0}
\scalebox{0.9}{%
\begin{tabular}{c c@{}c c@{}c c@{}c c@{}c}
\includegraphics[width=0.105\textwidth]{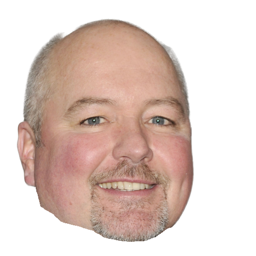} &
\includegraphics[width=0.105\textwidth]{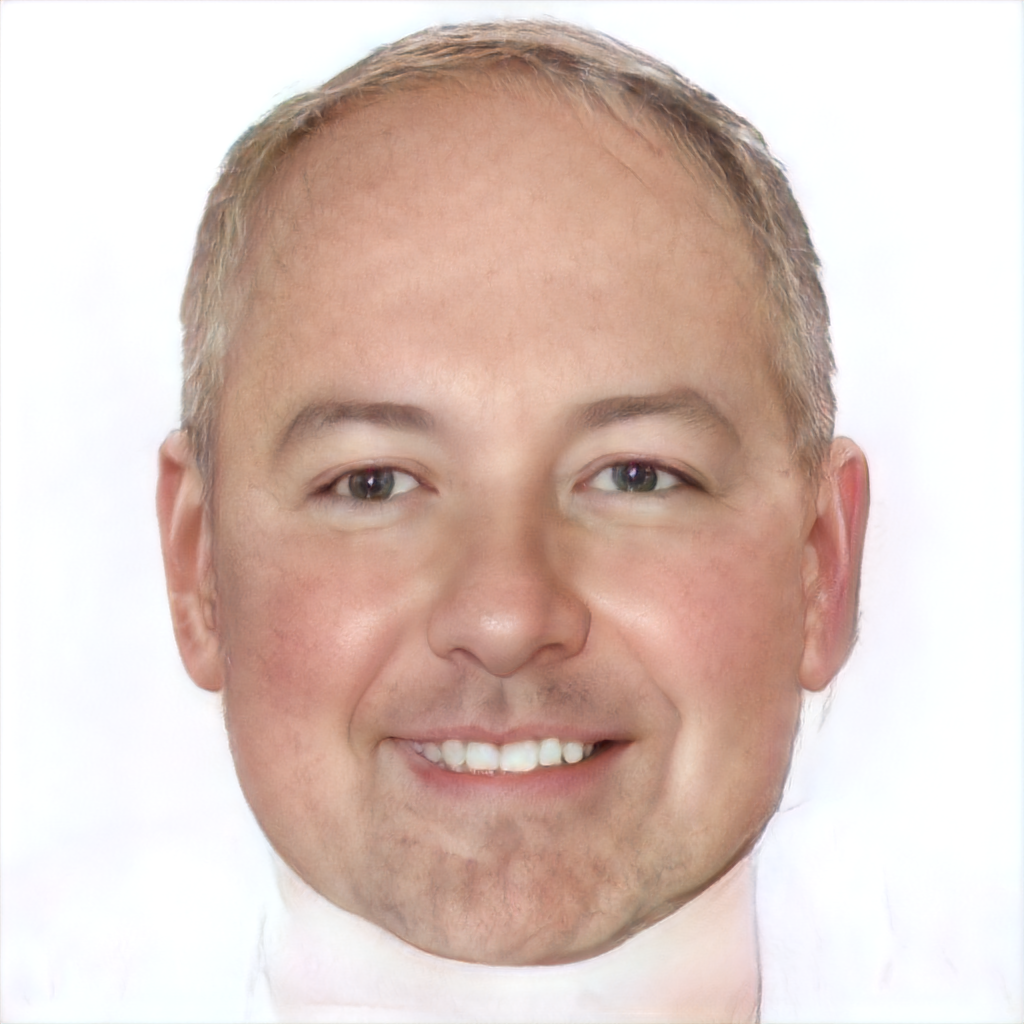} &
\includegraphics[width=0.105\textwidth]{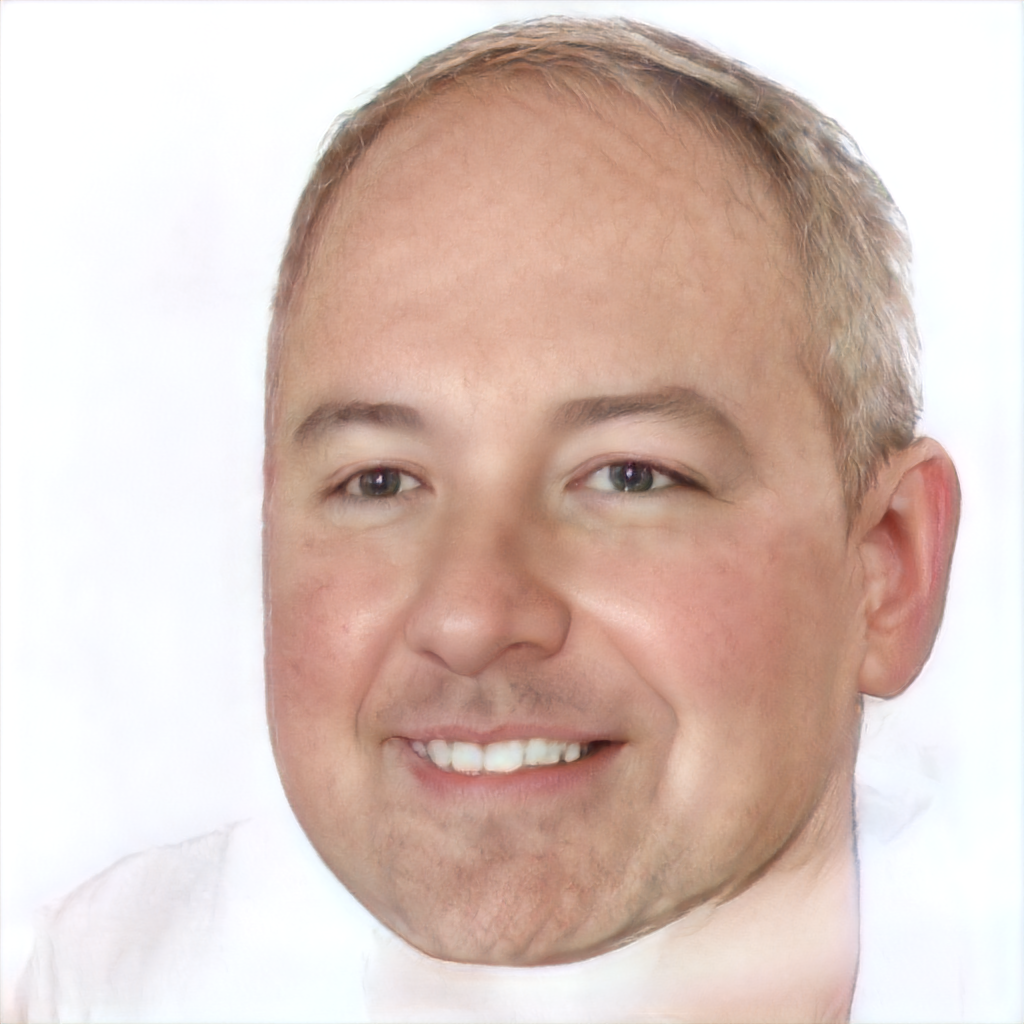} &
\includegraphics[width=0.105\textwidth]{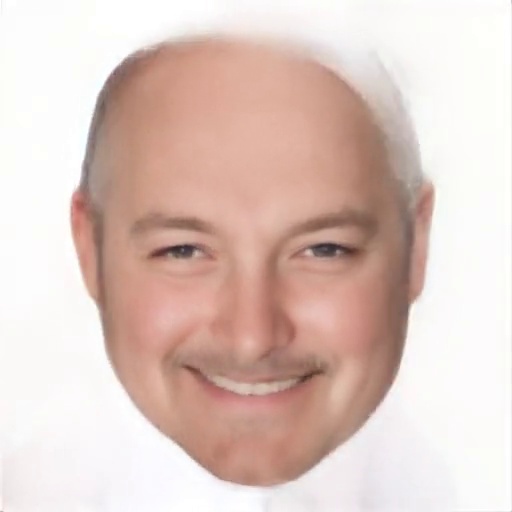} &
\includegraphics[width=0.105\textwidth]{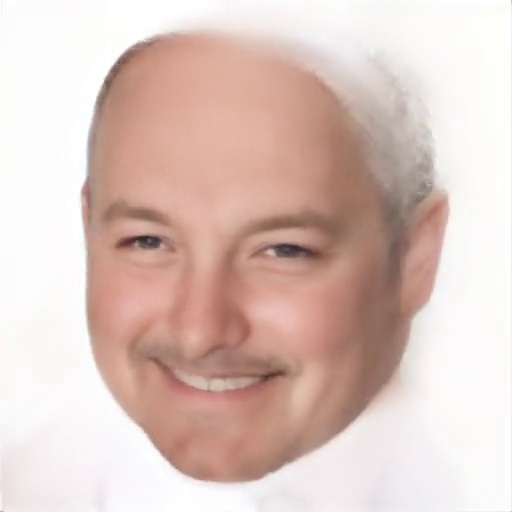} &
\includegraphics[width=0.105\textwidth]{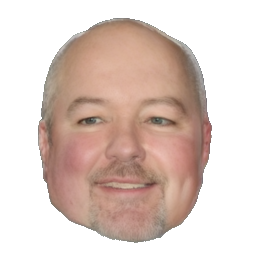} &
\includegraphics[width=0.105\textwidth]{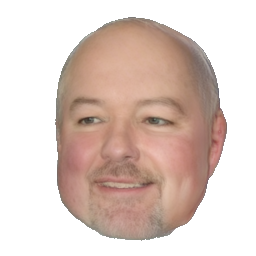} &
\includegraphics[width=0.105\textwidth]{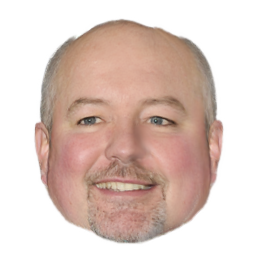} &
\includegraphics[width=0.105\textwidth]{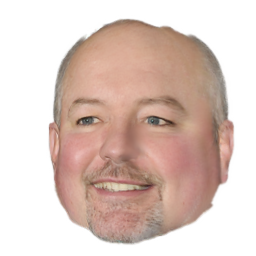} \\[-2pt]

\includegraphics[width=0.105\textwidth]{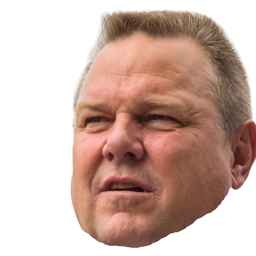} &
\includegraphics[width=0.105\textwidth]{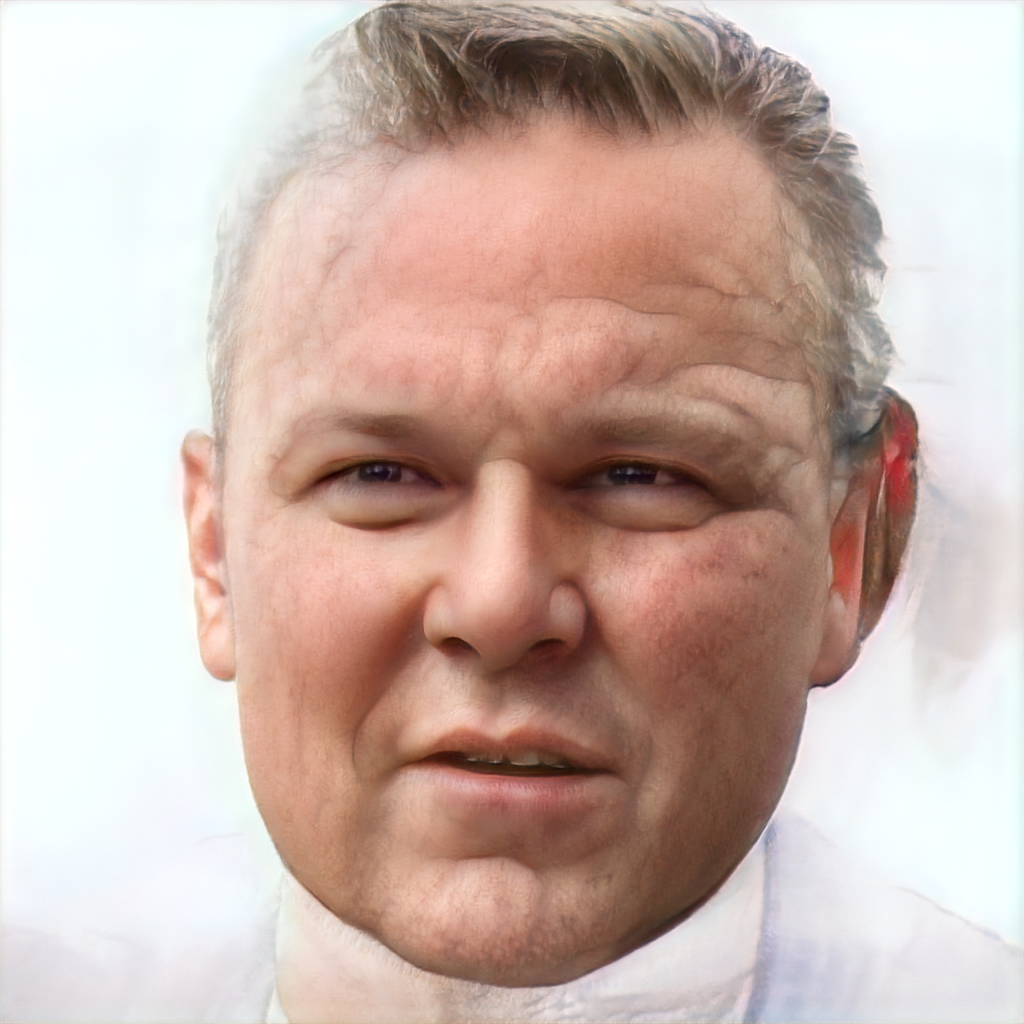} &
\includegraphics[width=0.105\textwidth]{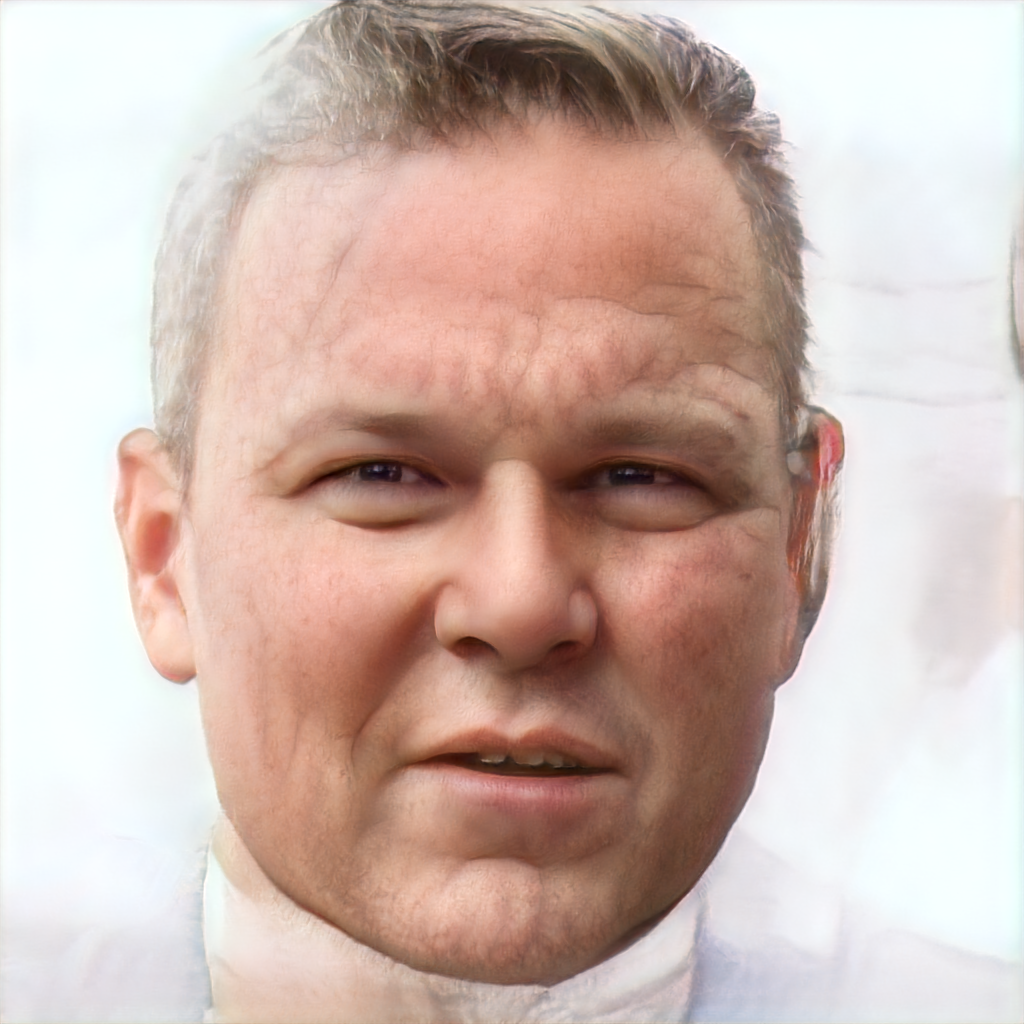} &
\includegraphics[width=0.105\textwidth]{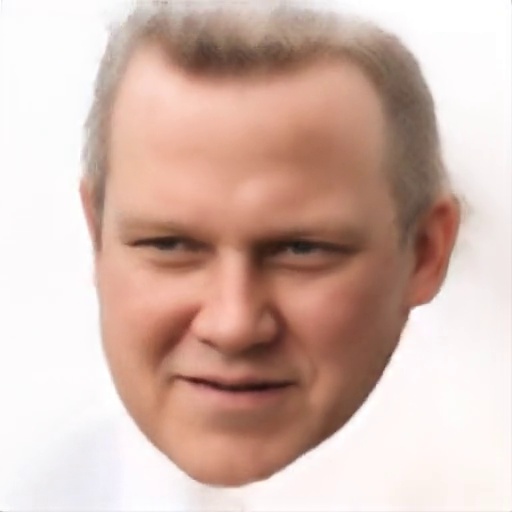} &
\includegraphics[width=0.105\textwidth]{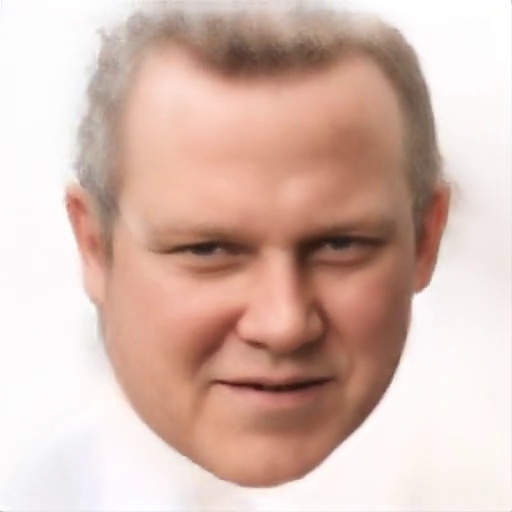} &
\includegraphics[width=0.105\textwidth]{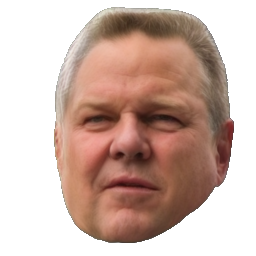} &
\includegraphics[width=0.105\textwidth]{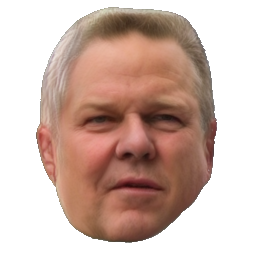} &
\includegraphics[width=0.105\textwidth]{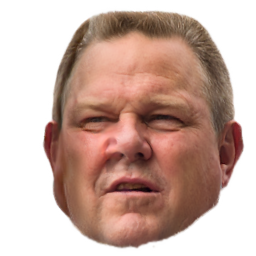} &
\includegraphics[width=0.105\textwidth]{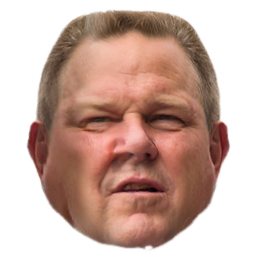} \\[-2pt]

\includegraphics[width=0.105\textwidth]{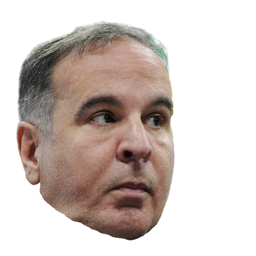} &
\includegraphics[width=0.105\textwidth]{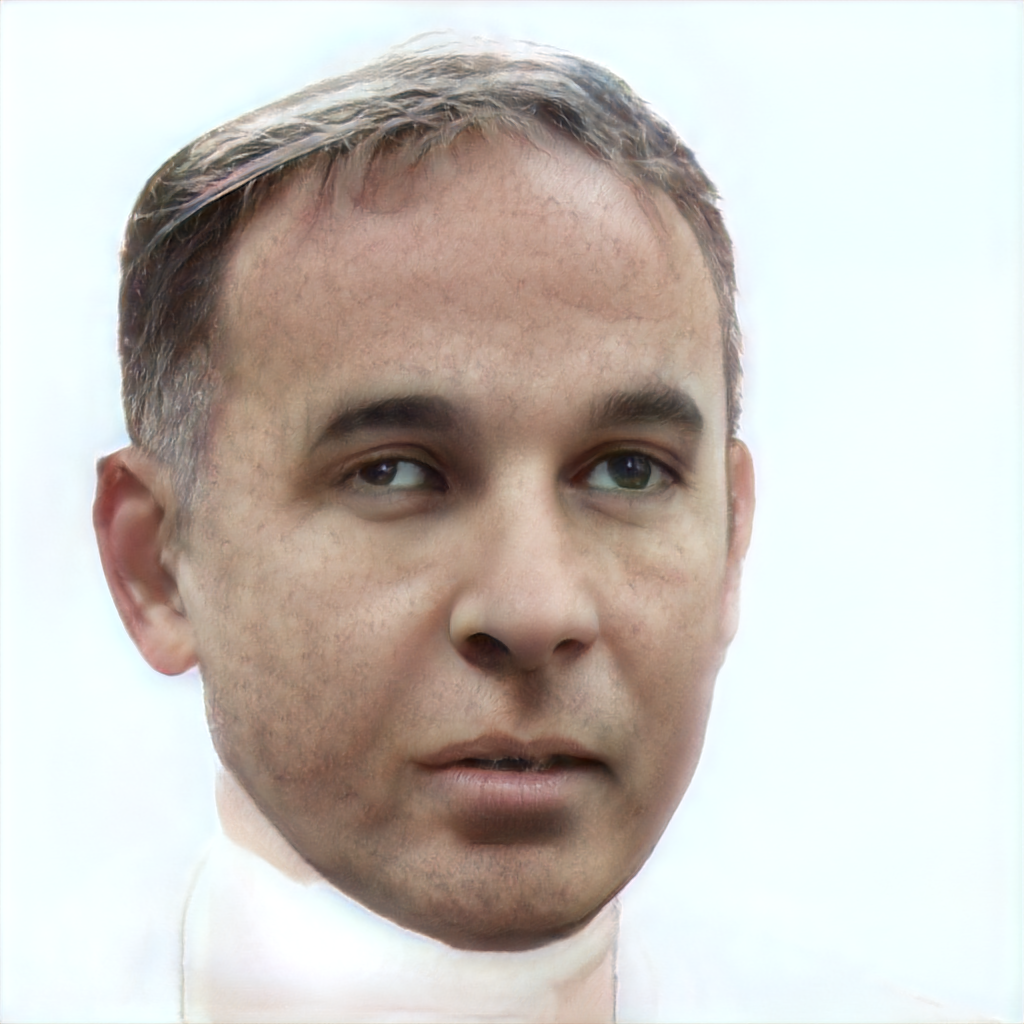} &
\includegraphics[width=0.105\textwidth]{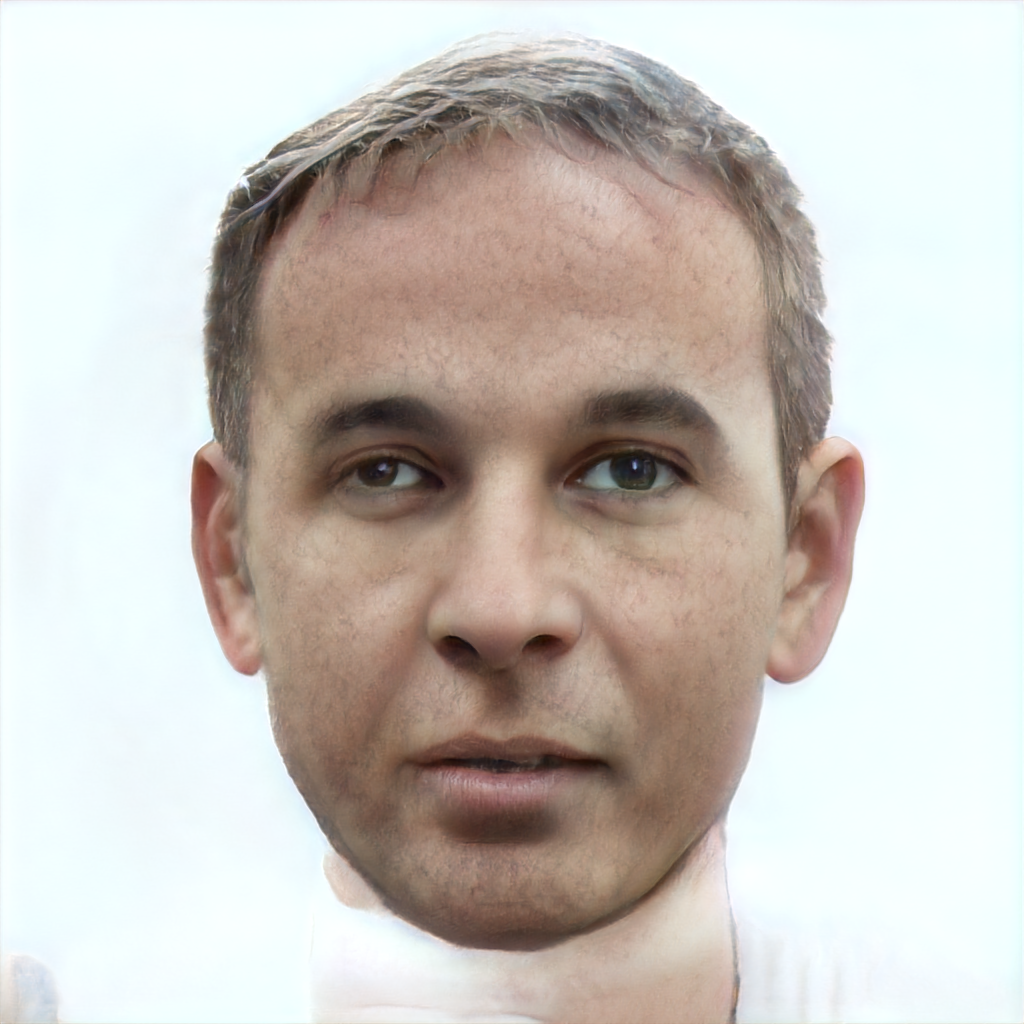} &
\includegraphics[width=0.105\textwidth]{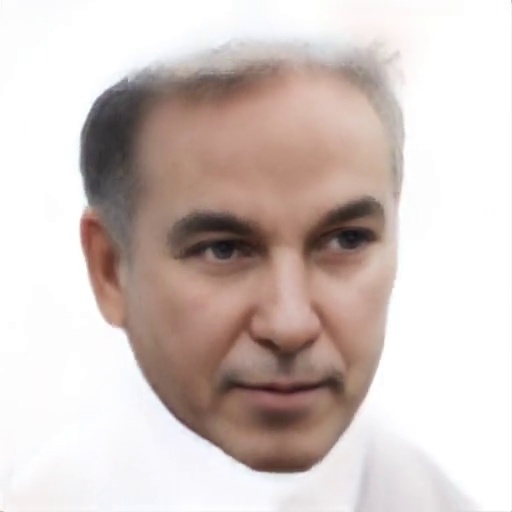} &
\includegraphics[width=0.105\textwidth]{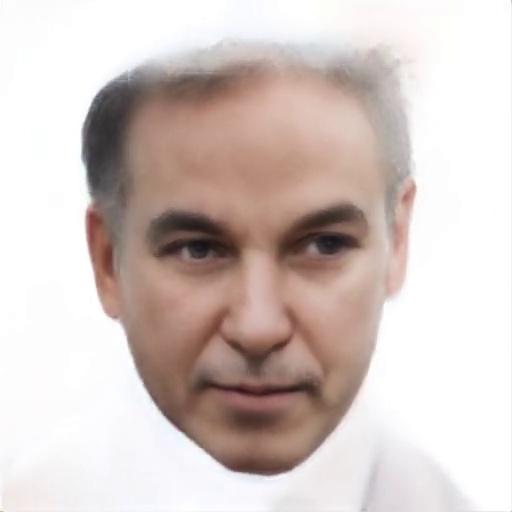} &
\includegraphics[width=0.105\textwidth]{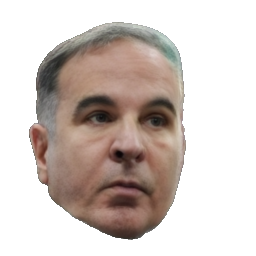} &
\includegraphics[width=0.105\textwidth]{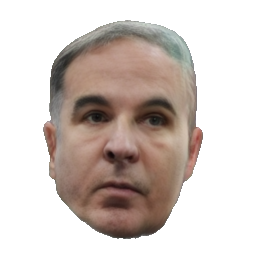} &
\includegraphics[width=0.105\textwidth]{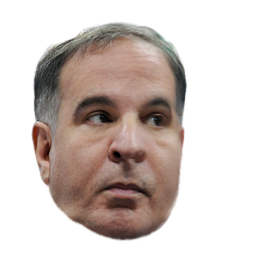} &
\includegraphics[width=0.105\textwidth]{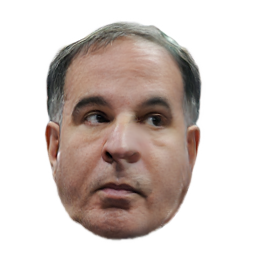} \\[-2pt]

\includegraphics[width=0.105\textwidth]{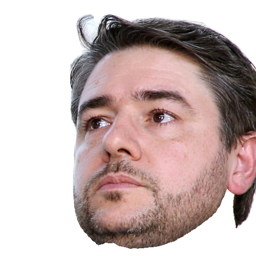} &
\includegraphics[width=0.105\textwidth]{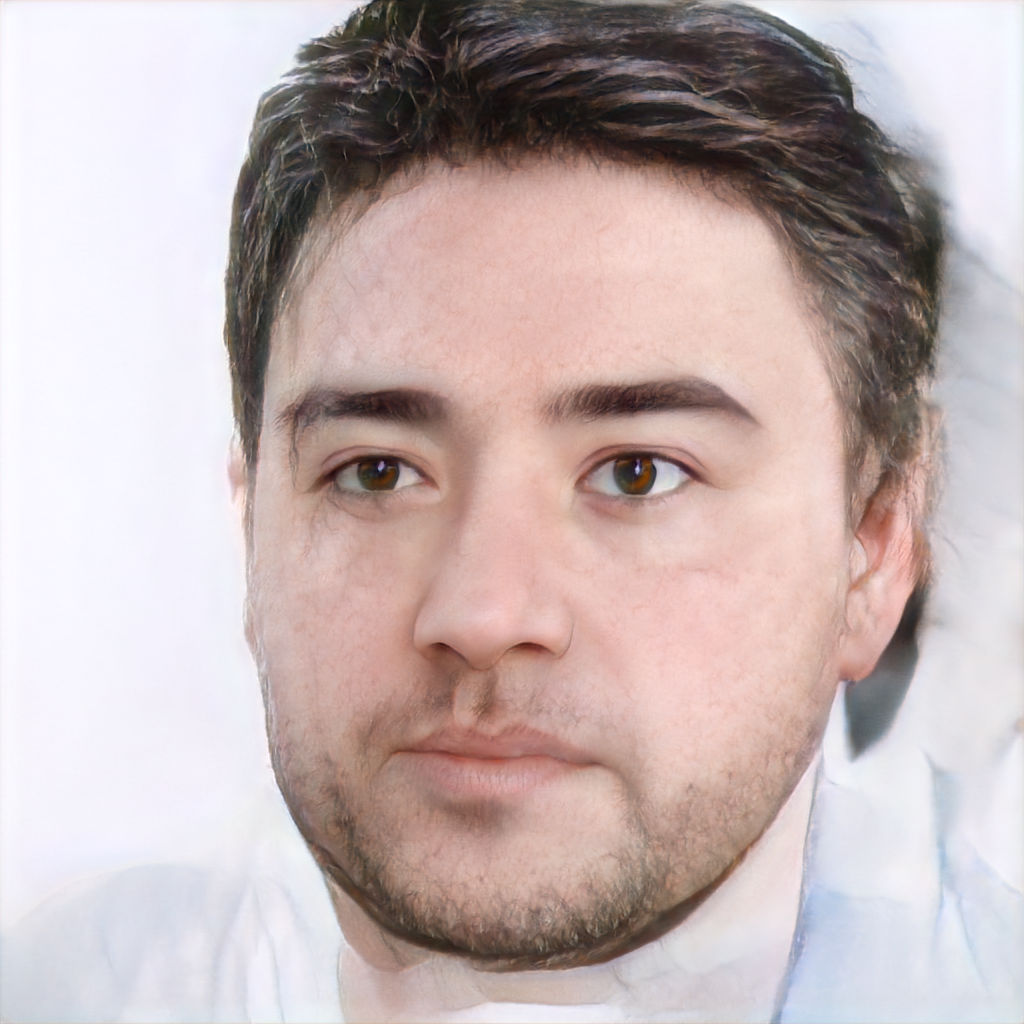} &
\includegraphics[width=0.105\textwidth]{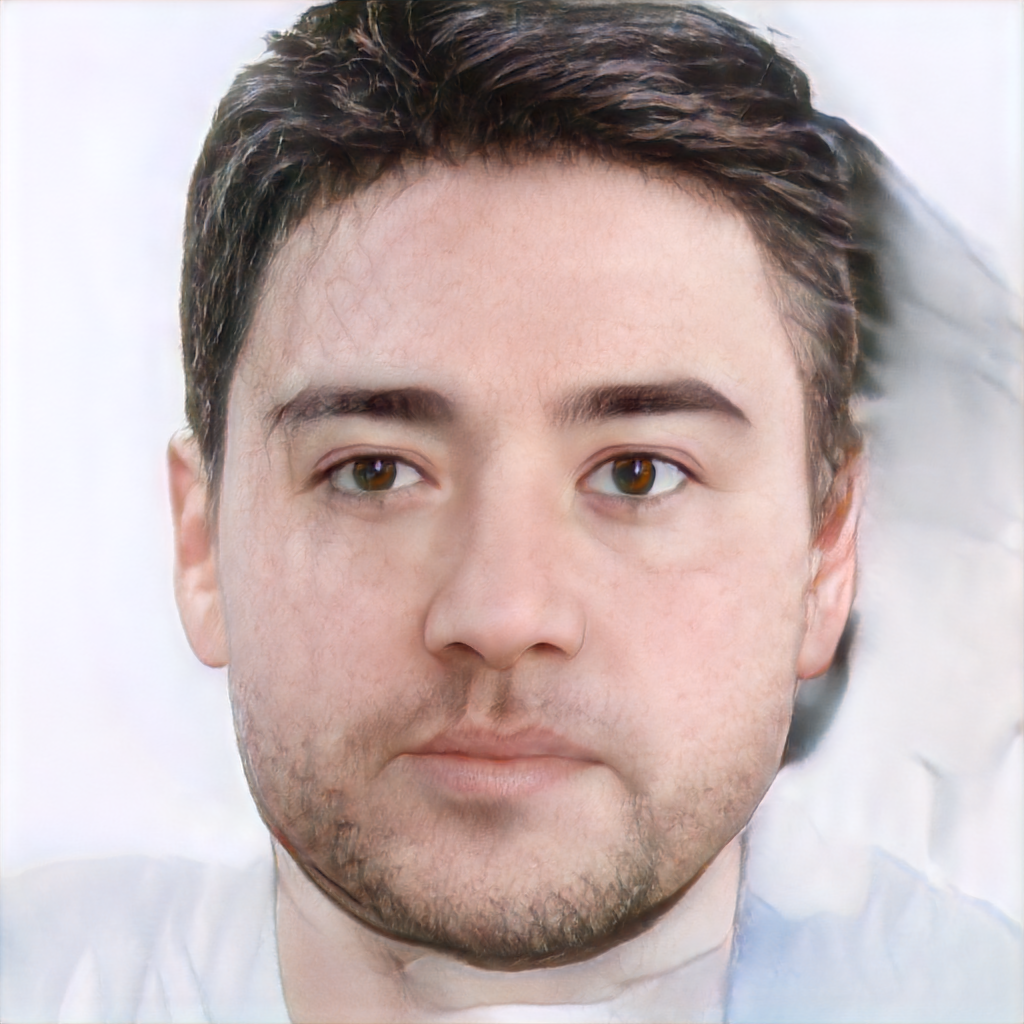} &
\includegraphics[width=0.105\textwidth]{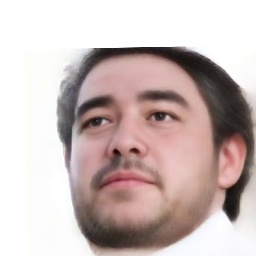} &
\includegraphics[width=0.105\textwidth]{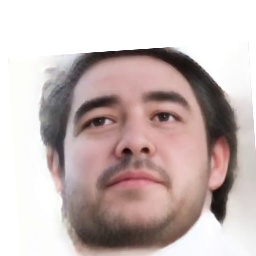} &
\includegraphics[width=0.105\textwidth]{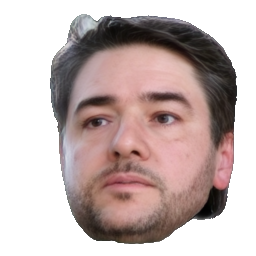} &
\includegraphics[width=0.105\textwidth]{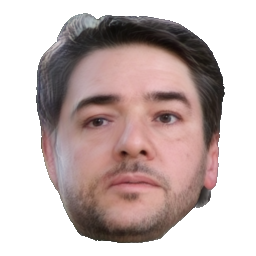} &
\includegraphics[width=0.105\textwidth]{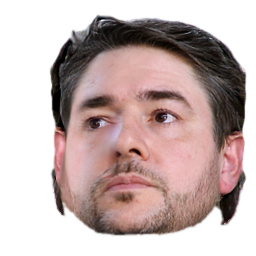} &
\includegraphics[width=0.105\textwidth]{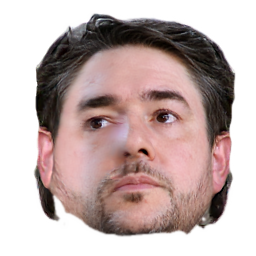} \\[-2pt]

\includegraphics[width=0.105\textwidth]{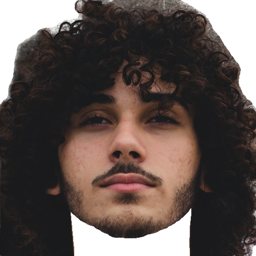} &
\includegraphics[width=0.105\textwidth]{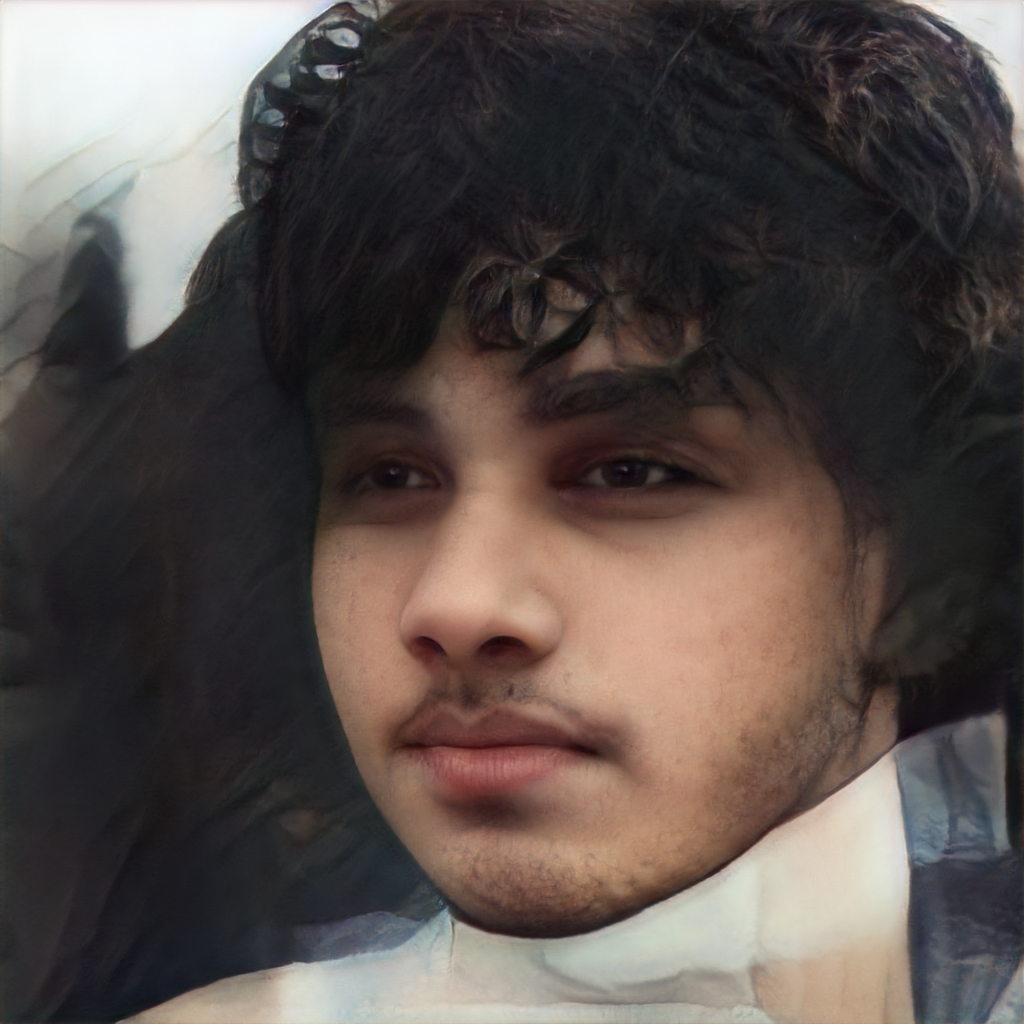} &
\includegraphics[width=0.105\textwidth]{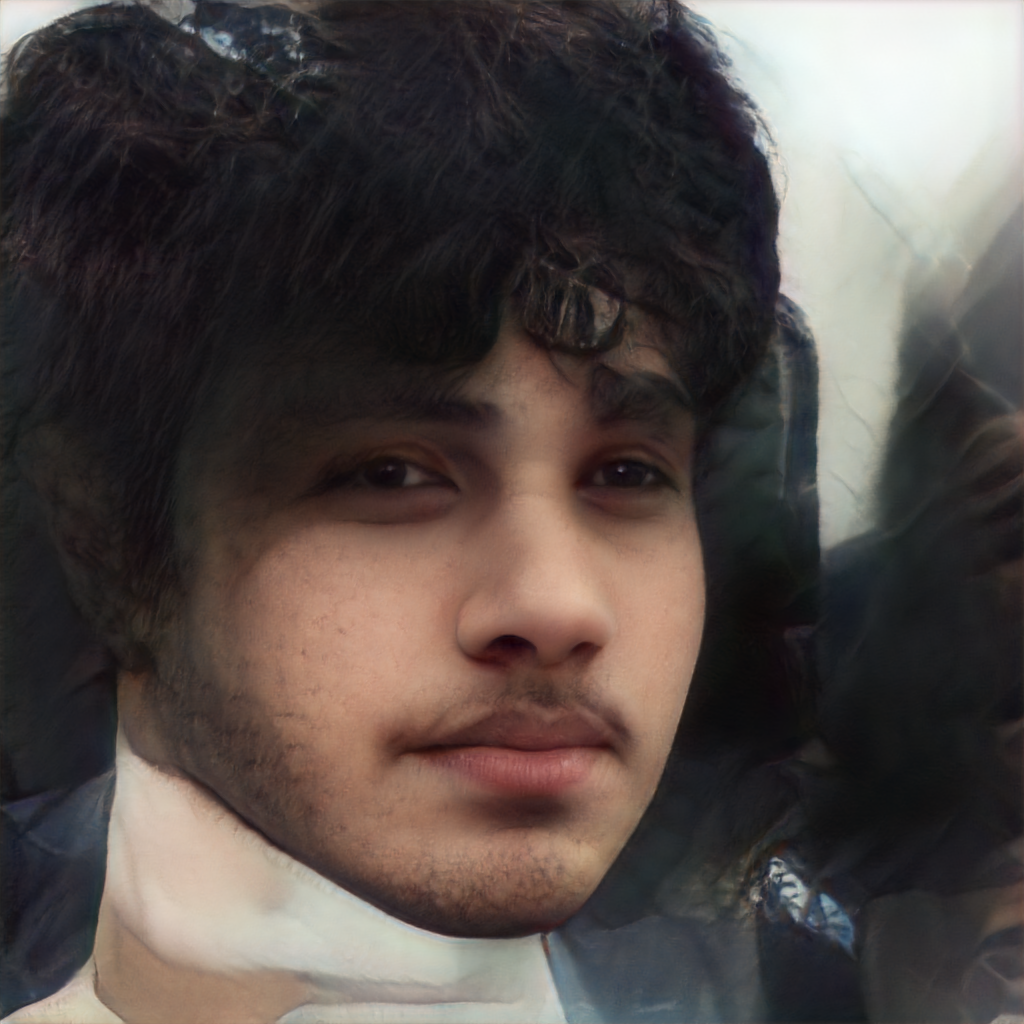} &
\includegraphics[width=0.105\textwidth]{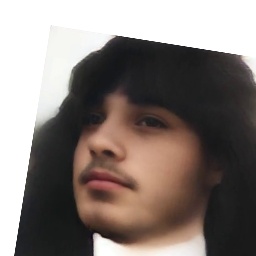} &
\includegraphics[width=0.105\textwidth]{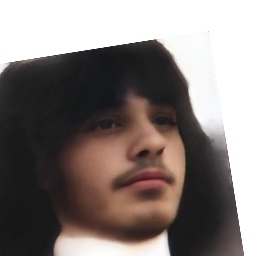} &
\includegraphics[width=0.105\textwidth]{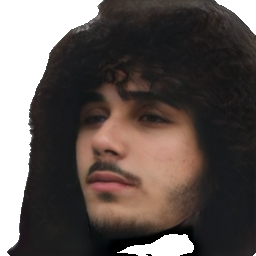} &
\includegraphics[width=0.105\textwidth]{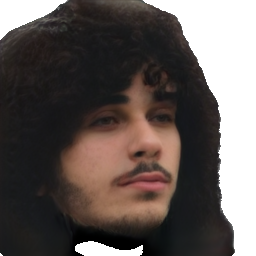} &
\includegraphics[width=0.105\textwidth]{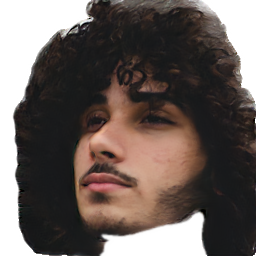} &
\includegraphics[width=0.105\textwidth]{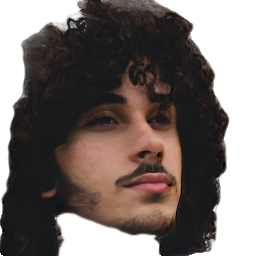} \\[-2pt]

\includegraphics[width=0.105\textwidth]{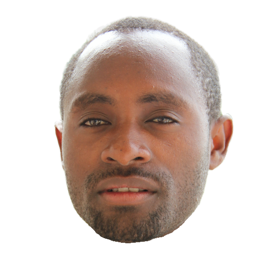} &
\includegraphics[width=0.105\textwidth]{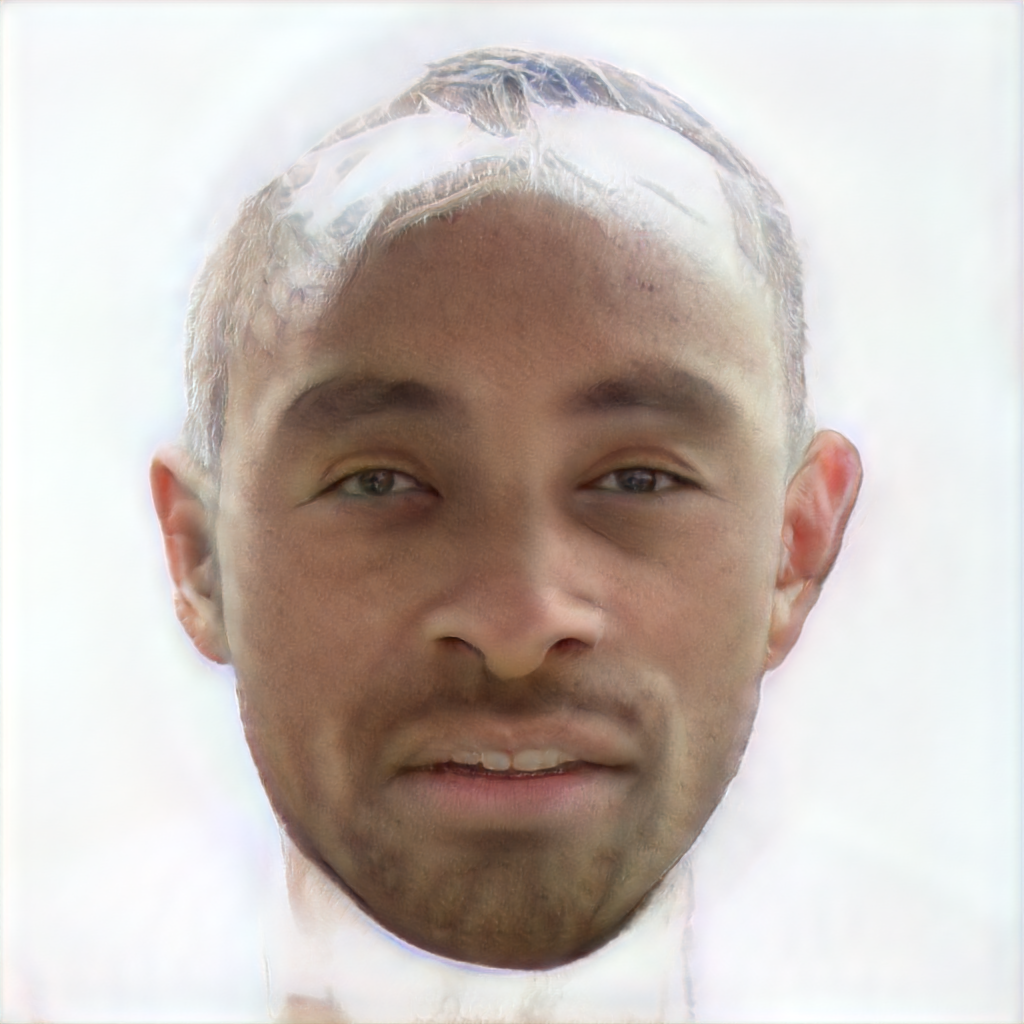} &
\includegraphics[width=0.105\textwidth]{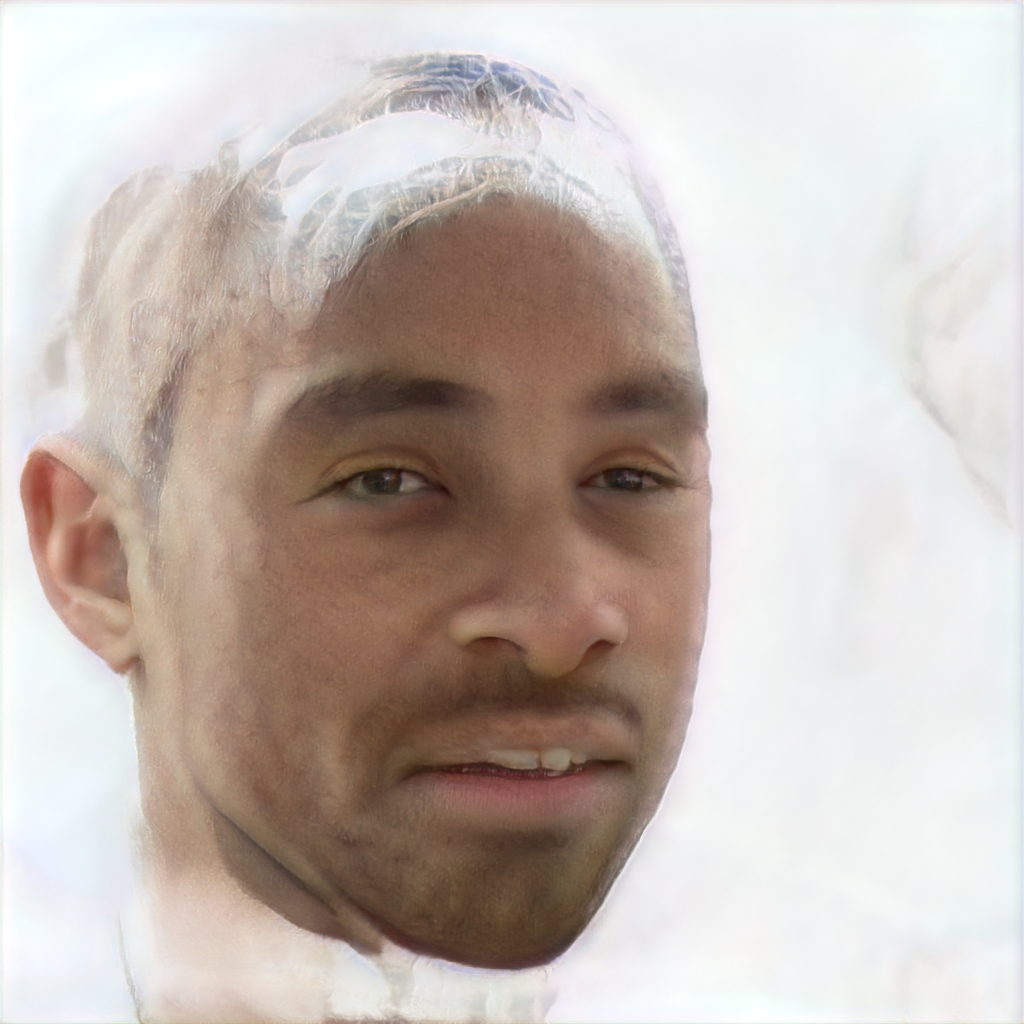} &
\includegraphics[width=0.105\textwidth]{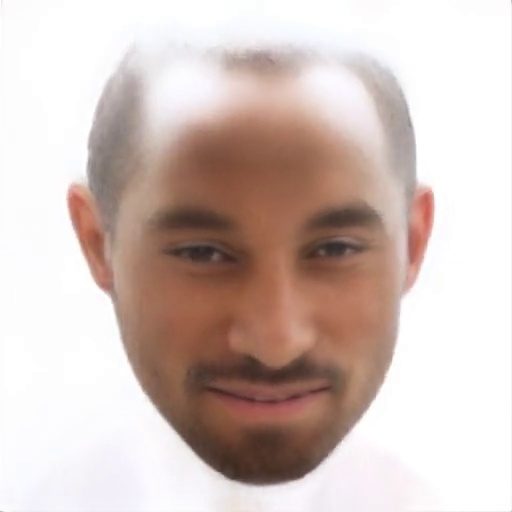} &
\includegraphics[width=0.105\textwidth]{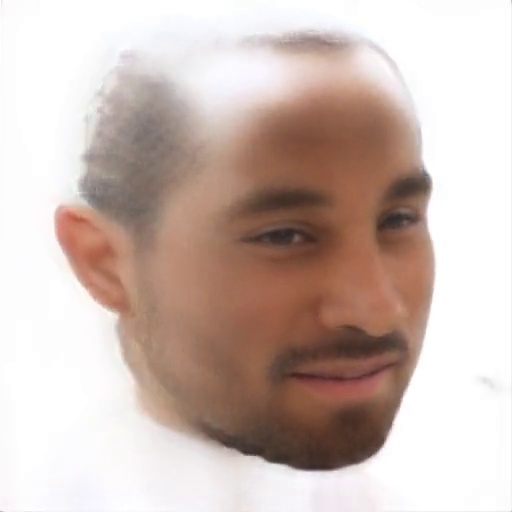} &
\includegraphics[width=0.105\textwidth]{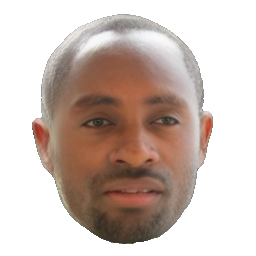} &
\includegraphics[width=0.105\textwidth]{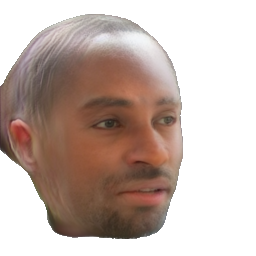} &
\includegraphics[width=0.105\textwidth]{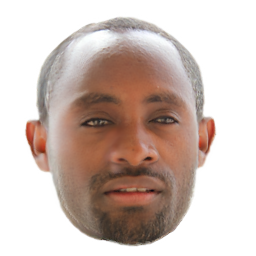} &
\includegraphics[width=0.105\textwidth]{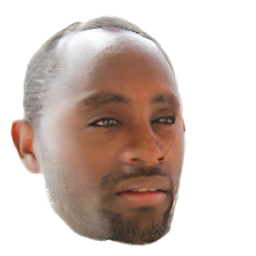} \\[-2pt]

\includegraphics[width=0.105\textwidth]{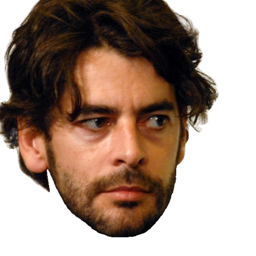} &
\includegraphics[width=0.105\textwidth]{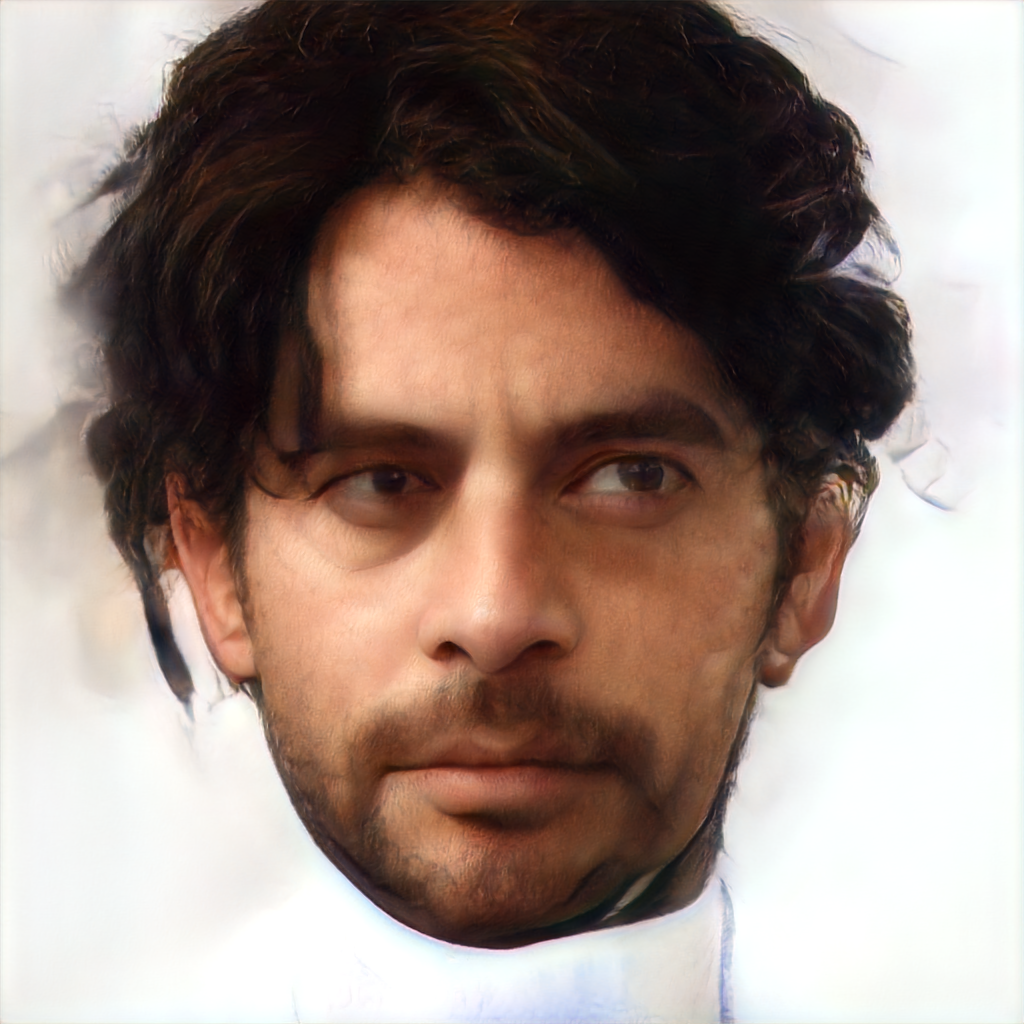} &
\includegraphics[width=0.105\textwidth]{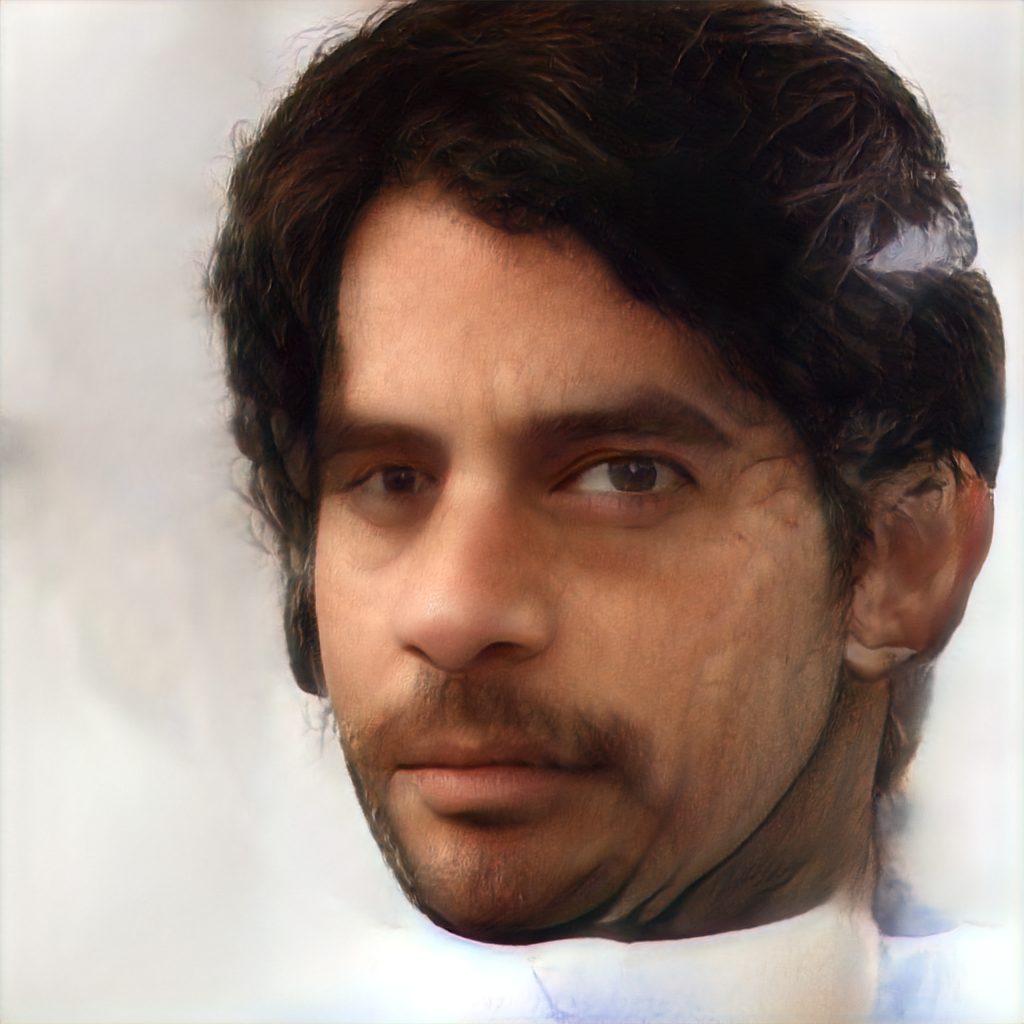} &
\includegraphics[width=0.105\textwidth]{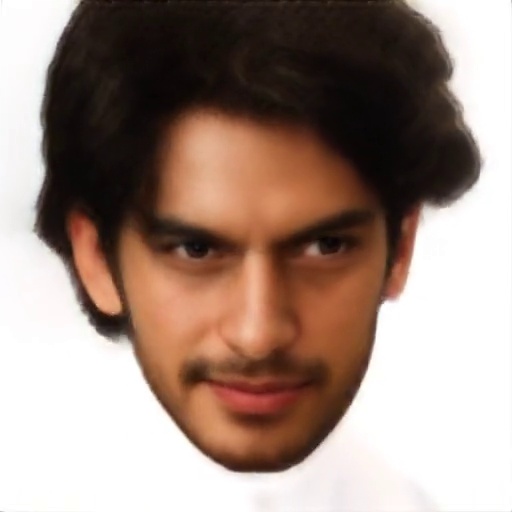} &
\includegraphics[width=0.105\textwidth]{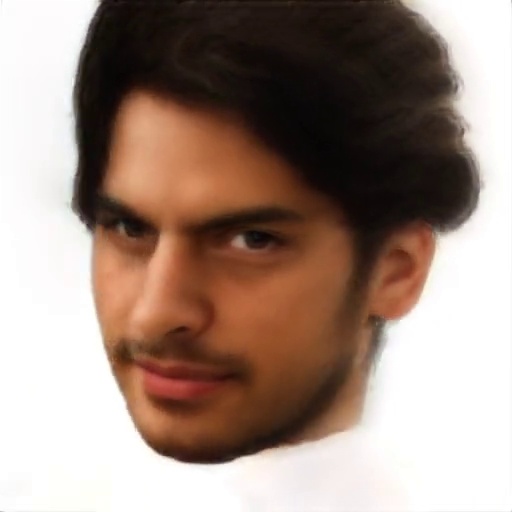} &
\includegraphics[width=0.105\textwidth]{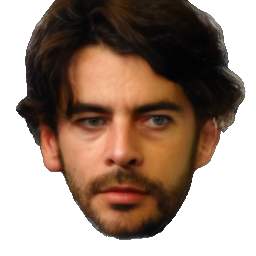} &
\includegraphics[width=0.105\textwidth]{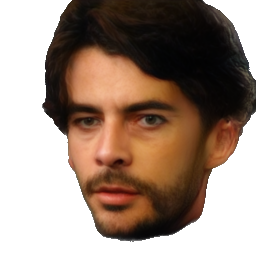} &
\includegraphics[width=0.105\textwidth]{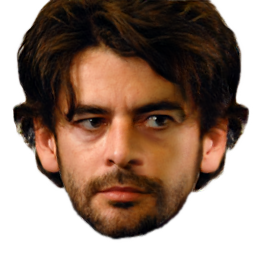} &
\includegraphics[width=0.105\textwidth]{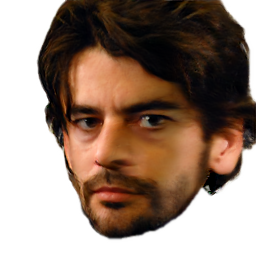} \\[-2pt]

\includegraphics[width=0.105\textwidth]{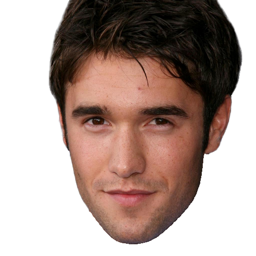} &
\includegraphics[width=0.105\textwidth]{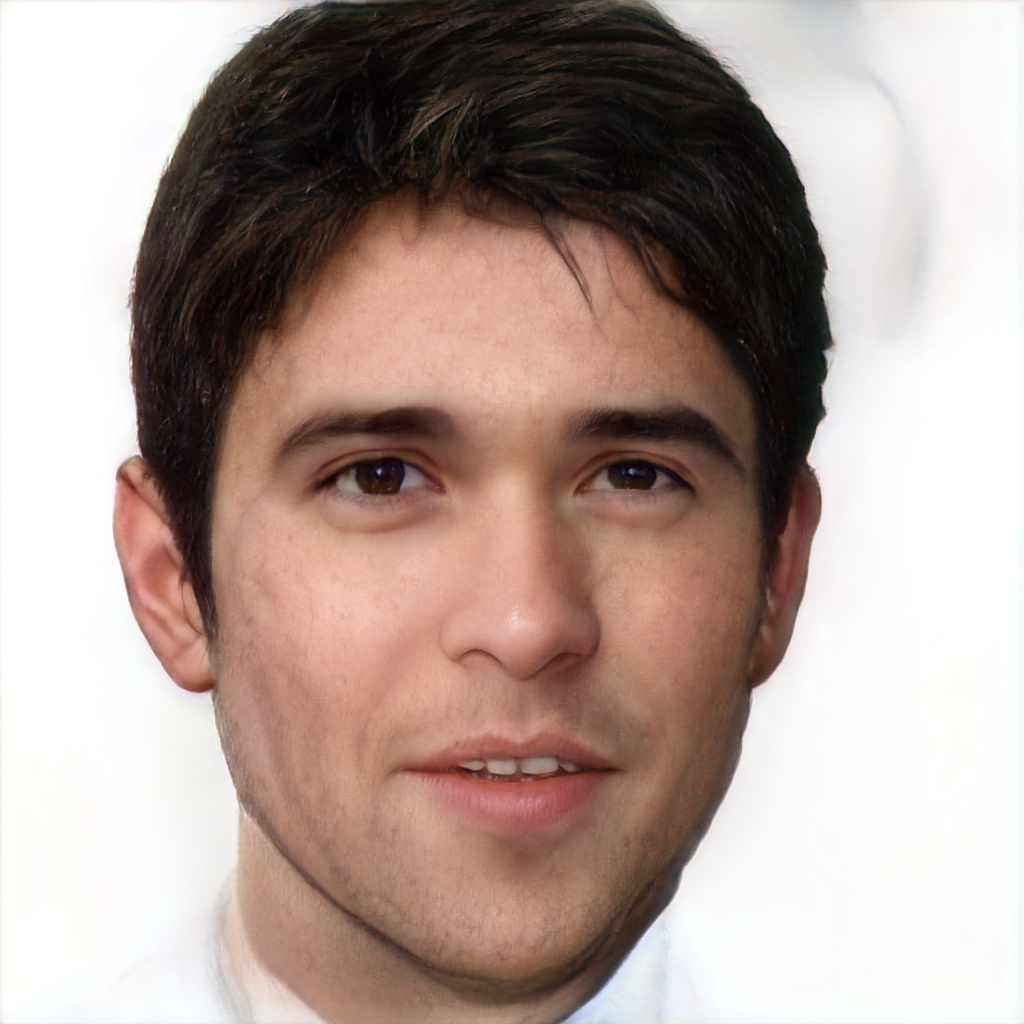} &
\includegraphics[width=0.105\textwidth]{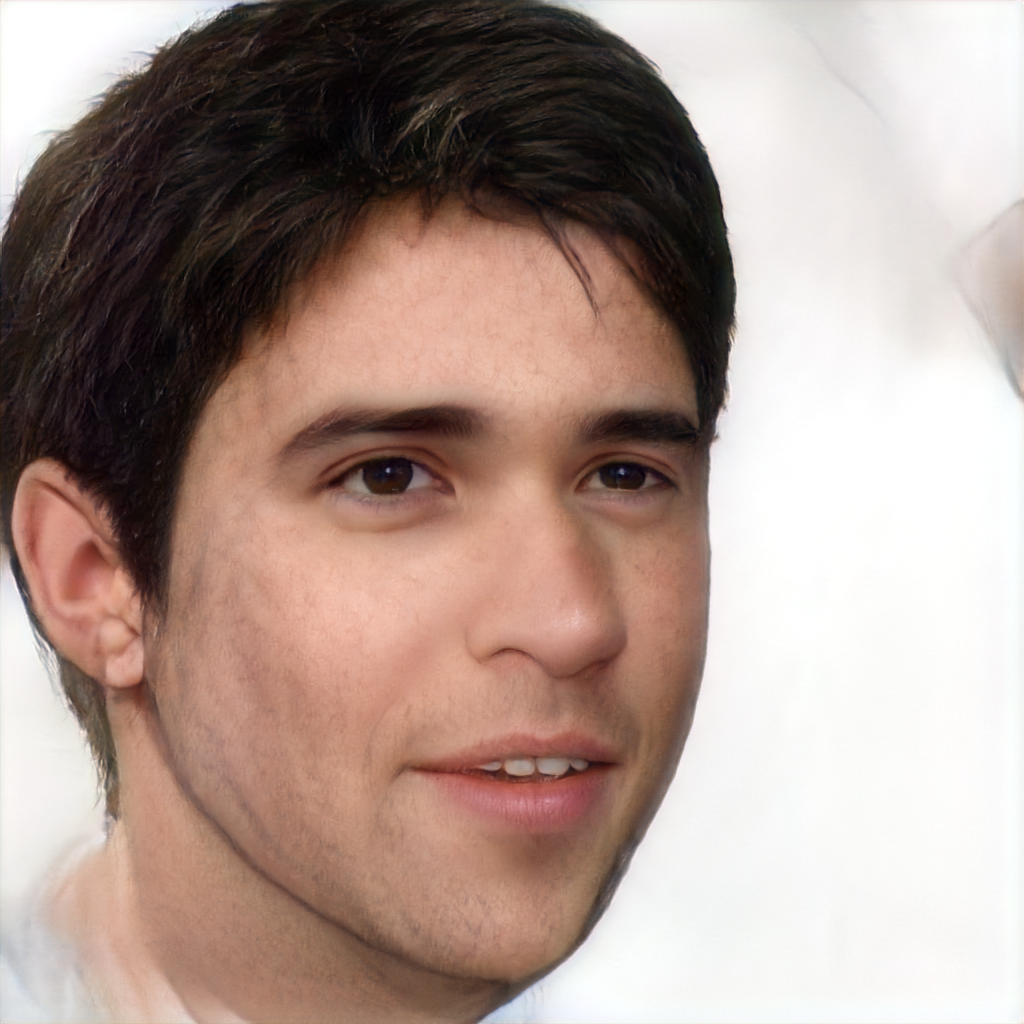} &
\includegraphics[width=0.105\textwidth]{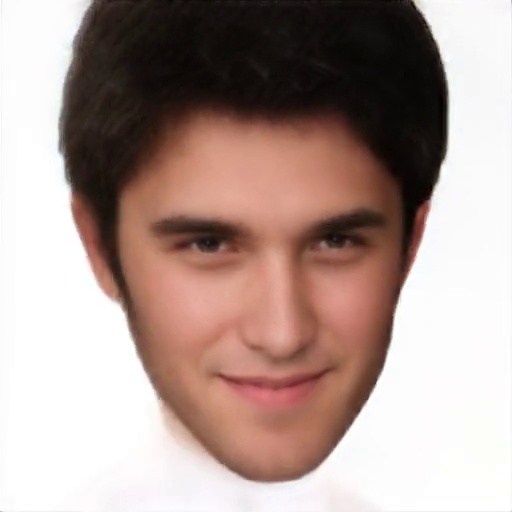} &
\includegraphics[width=0.105\textwidth]{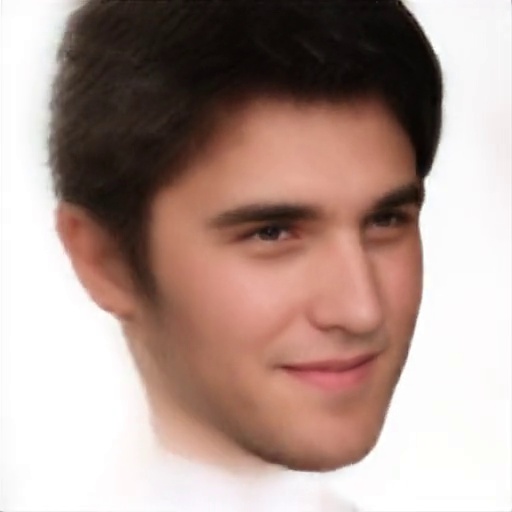} &
\includegraphics[width=0.105\textwidth]{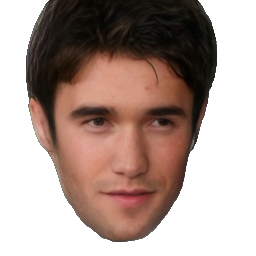} &
\includegraphics[width=0.105\textwidth]{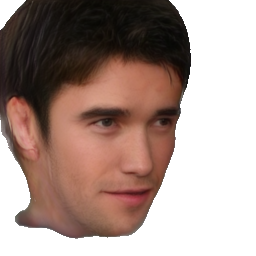} &
\includegraphics[width=0.105\textwidth]{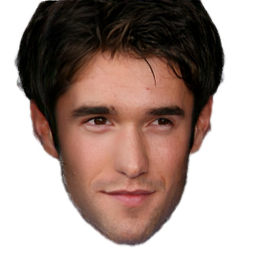} &
\includegraphics[width=0.105\textwidth]{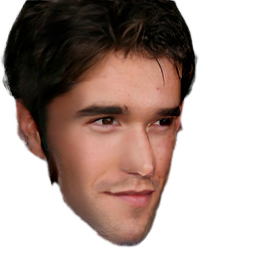} \\[-2pt]

\includegraphics[width=0.105\textwidth]{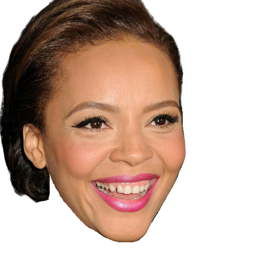} &
\includegraphics[width=0.105\textwidth]{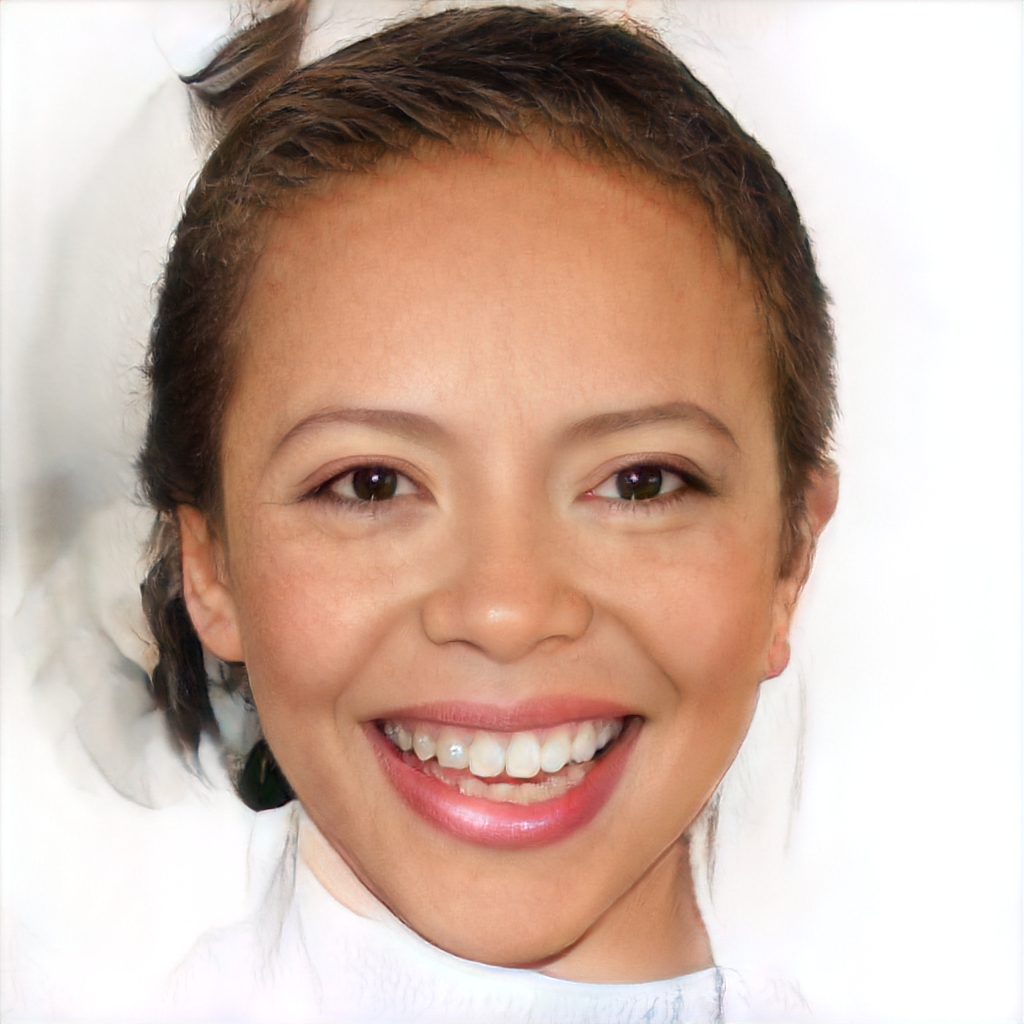} &
\includegraphics[width=0.105\textwidth]{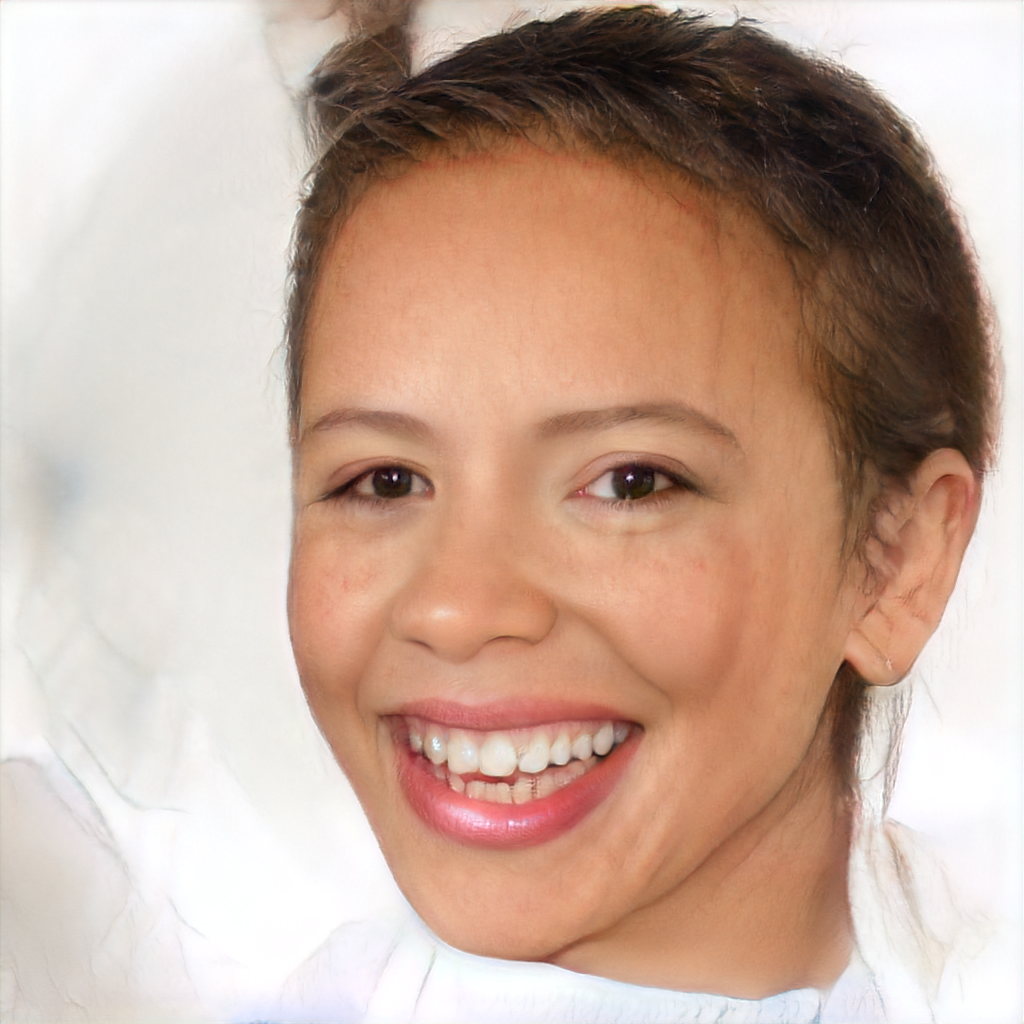} &
\includegraphics[width=0.105\textwidth]{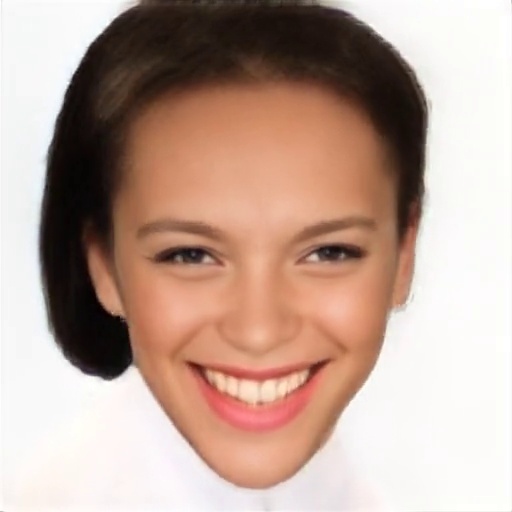} &
\includegraphics[width=0.105\textwidth]{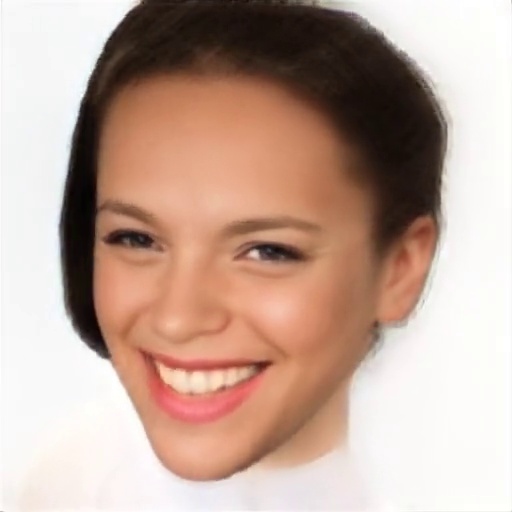} &
\includegraphics[width=0.105\textwidth]{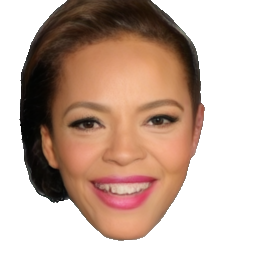} &
\includegraphics[width=0.105\textwidth]{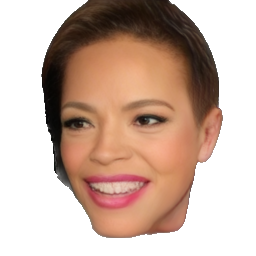} &
\includegraphics[width=0.105\textwidth]{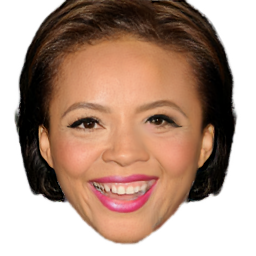} &
\includegraphics[width=0.105\textwidth]{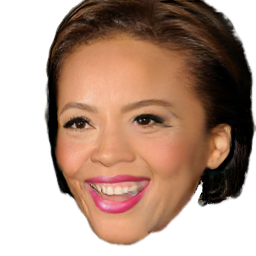} \\[-2pt]

\includegraphics[width=0.105\textwidth]{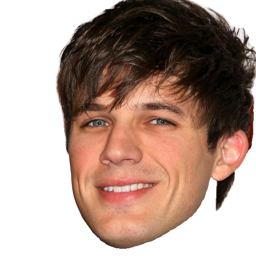} &
\includegraphics[width=0.105\textwidth]{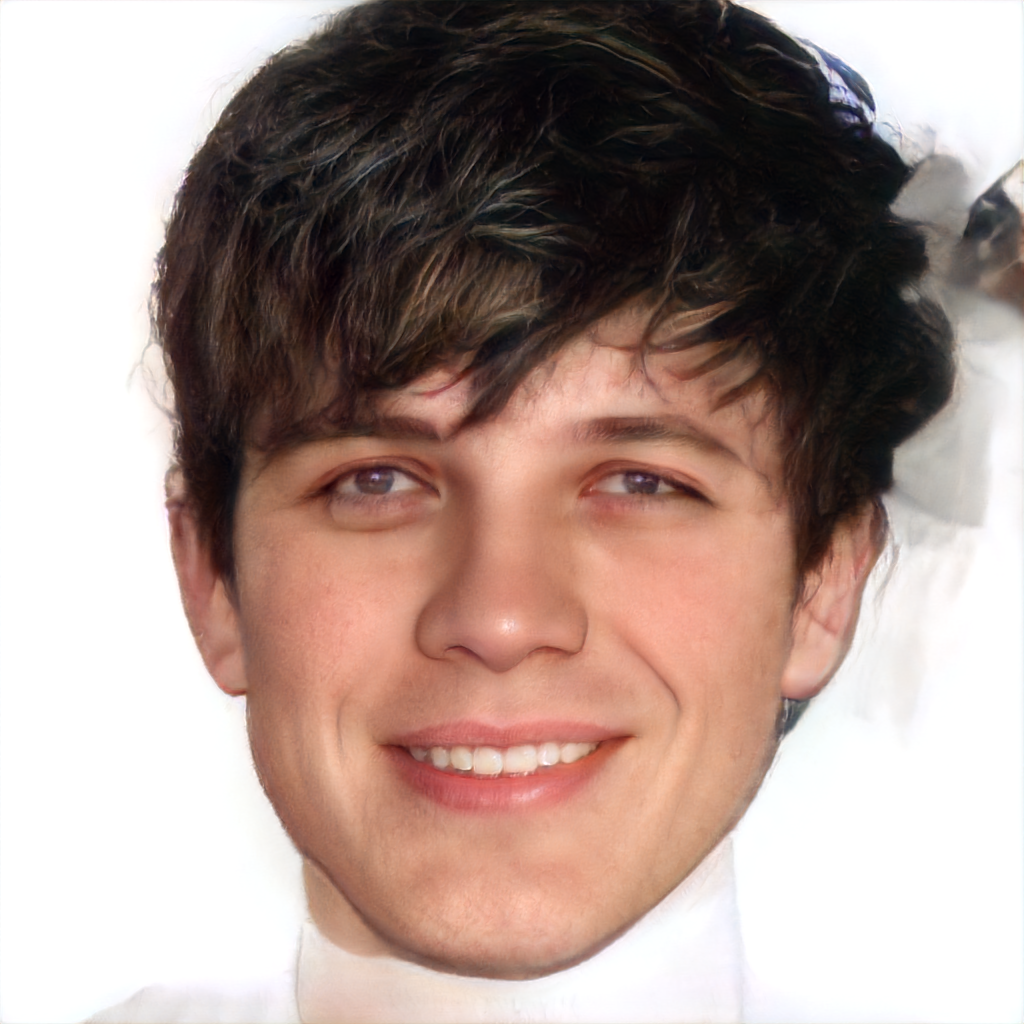} &
\includegraphics[width=0.105\textwidth]{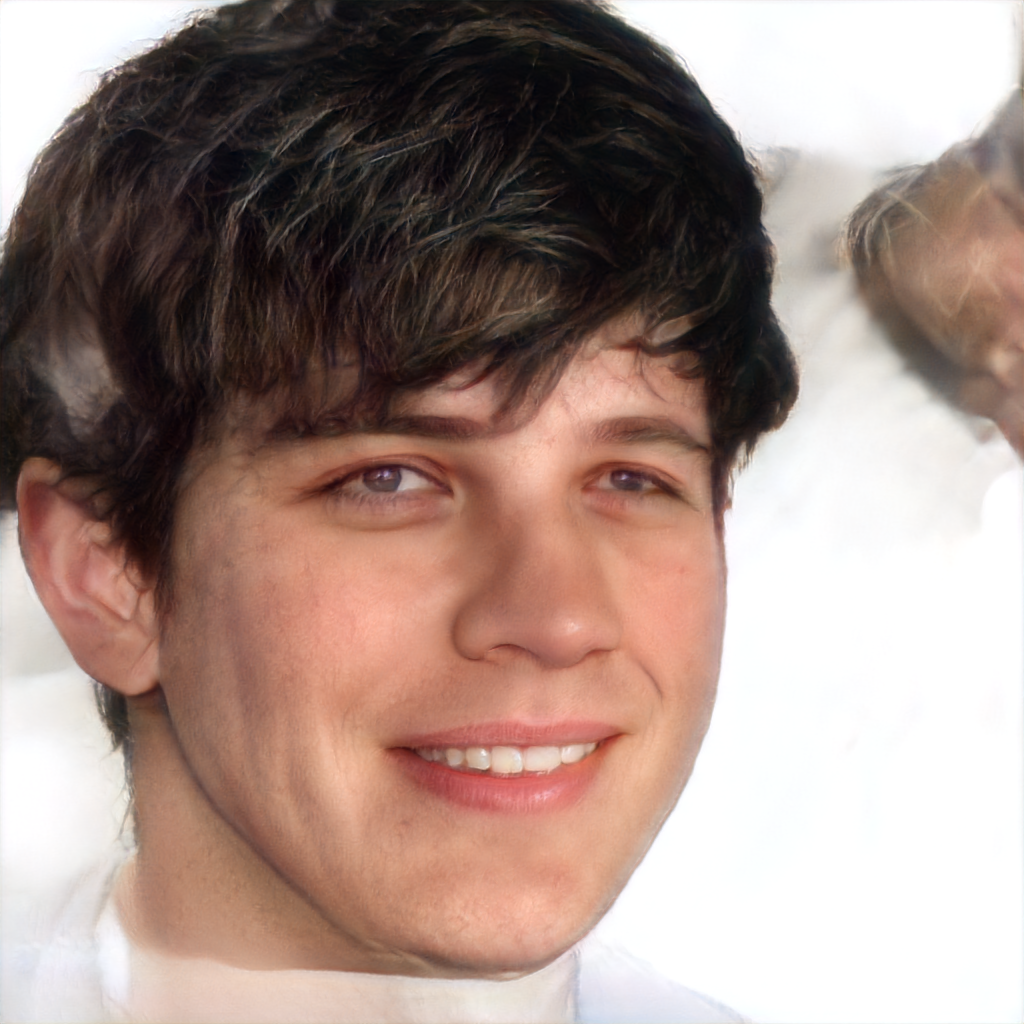} &
\includegraphics[width=0.105\textwidth]{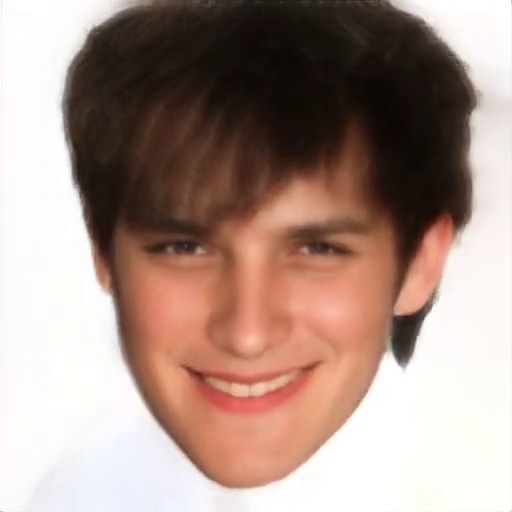} &
\includegraphics[width=0.105\textwidth]{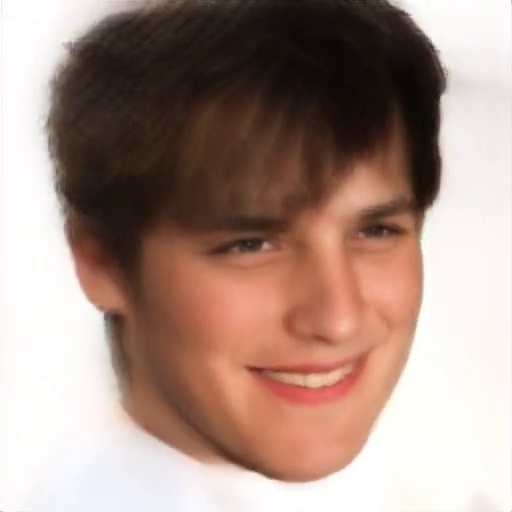} &
\includegraphics[width=0.105\textwidth]{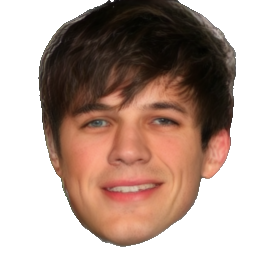} &
\includegraphics[width=0.105\textwidth]{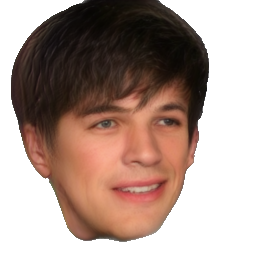} &
\includegraphics[width=0.105\textwidth]{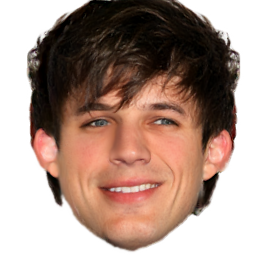} &
\includegraphics[width=0.105\textwidth]{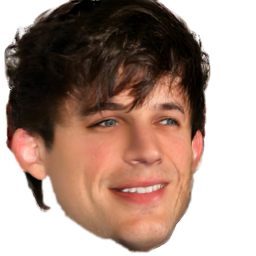} \\

Input & \multicolumn{2}{c}{E$3$DGE} & \multicolumn{2}{c}{G-NeRF} & \multicolumn{2}{c}{GAGAvatar} & \multicolumn{2}{c}{Ours} \\
\end{tabular}}
\caption{\small \textbf{Additional Comparisons.} We show improvements in preserving facial details, hair texture, and eye structure/color, and also better reconstructing occluded face regions (\textit{e.g.} recovering the subject’s occluded ear in the fourth row).}
\label{fig:nvs_supp}
\end{figure*}
\clearpage

\section{Additional Gaussian Budget Comparison}
To further analyze the efficiency of our adaptive allocation strategy, we compare our method using 12$k$ Gaussians with GAGAvatar~\cite{gagavatar} using 24$k$ Gaussians, giving GAGAvatar twice the Gaussian budget. As shown in Table~\ref{tab:higher_budget}, our method achieves competitive or better performance while using only half the number of Gaussians. These results show that our adaptive allocation strategy provides a compact and efficient Gaussian representation by focusing the budget on semantically important facial regions.

\begin{table}[h]
\centering
\resizebox{0.9\columnwidth}{!}{%
\setlength\tabcolsep{12pt}
\begin{tabular}{c c c c c}
\toprule
$M$ & Method & PSNR$\uparrow$ & SSIM$\uparrow$ & LPIPS$\downarrow$ \\
\midrule
24$k$ & GAGAvatar~\cite{gagavatar}  & $21.72$ & $\mathbf{0.837}$ & $0.100$ \\
12$k$ & Ours                        & $\mathbf{22.91}$ & $0.835$ & $\mathbf{0.096}$ \\
\bottomrule
\end{tabular}}
\caption{\small
\textbf{Additional NeRSemble NVS.} Additional quantitative comparison on the NeRSemble dataset using 12$k$ Gaussians for our method and 24$k$ Gaussians for GAGAvatar.
}
\label{tab:higher_budget}
\end{table}

\section{Additional Qualitative Results}
We show more qualitative results for the novel view synthesis (See Fig.~\ref{fig:nvs_supp}). Thanks to our GAM, EGGHead noticeably improves over the baselines in representing hair texture, eye structure and color, and other facial details such as creases, wrinkles, and dimples. Our face symmetry module further improves reconstruction quality in occluded regions (\textit{e.g.} recovering the subject’s occluded ear in the fourth row of Fig.~\ref{fig:nvs_supp}).

\section{Delighting Preprocessing}
\begin{figure}[ht]
\centering
\resizebox{0.9\columnwidth}{!}{
\setlength{\tabcolsep}{0.5pt}
\renewcommand{\arraystretch}{1.0}
\begin{tabular}{c c c}
\includegraphics[width=0.24\columnwidth]{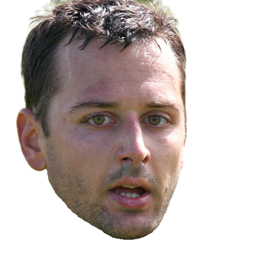} &
\includegraphics[width=0.24\columnwidth]{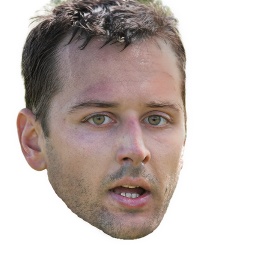} &
\includegraphics[width=0.24\columnwidth]{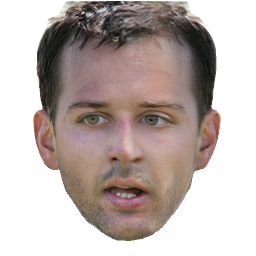} \\
{\scriptsize Input} & {\scriptsize Input (deshadowed)} & {\scriptsize Novel View} \\
\end{tabular}
}
\caption{\textbf{Delighting Preprocessing.} We apply an off-the-shelf delighting method to the input image during enrollment to reduce asymmetric illumination artifacts.}
\label{fig:exp_delit}
\end{figure}

As discussed in the main paper, uneven lighting can introduce artifacts when mirroring Gaussians across the facial symmetry plane. Fig.~\ref{fig:exp_delit} illustrates the effect of applying an off-the-shelf delighting method to the input image during enrollment.

\section{Licenses and Responsible Use of Human Data}
We use three existing datasets in our work: FFHQ~\cite{FFHQ}, CelebA-HQ~\cite{CelebA-HQ}, and NeRSemble~\cite{NeRSembleMR}. All datasets allow free use for non-commercial research purposes. Aside from the subject faces, none of them contains any additional personally identifiable information or offensive content.

\section{Potential Societal Impacts} 
Creating photorealistic human heads always has the potential for generating malicious edited content of real subjects. We certainly do not condone such misuse. One potential way to address this problem is to train deepfake methods~\cite{malp, reverse-engineering-of-generative-models-inferring-model-hyperparameters-from-generated-images, hierarchical-fine-grained-image-forgery-detection-and-localization, interpretabledeepfake, intrinsicdeepfake, icasspdeepfake, revisitingdeepfake, universaldeepfake} that can distinguish our generated heads from real images using some of our synthesized heads as part of training data.

{
    \small
    \bibliographystyle{ieeenat_fullname}
    \bibliography{main}

@String(PAMI = {IEEE Trans. Pattern Anal. Mach. Intell.})

@String(CVPR= {IEEE Conf. Comput. Vis. Pattern Recog.})

@String(ICCV= {Int. Conf. Comput. Vis.})

@String(ECCV= {Eur. Conf. Comput. Vis.})

@String(NIPS= {Adv. Neural Inform. Process. Syst.})

@String(BMVC= {Brit. Mach. Vis. Conf.})

@String(TOG= {ACM Trans. Graph.})

@String(TIP  = {IEEE Trans. Image Process.})

@String(ACMMM= {ACM Int. Conf. Multimedia})

@String(ICASSP=	{ICASSP})

@String(ICLR = {Int. Conf. Learn. Represent.})

@String(SIGGRAPH = {ACM SIGGRAPH})

@String(SIGGRAPHASIA = {ACM SIGGRAPH Asia})

@String(WACV = {IEEE Winter Conf. Appl. Comput. Vis.})

@inproceedings{gagavatar,
    title={Generalizable and Animatable Gaussian Head Avatar},
    author={Xuangeng Chu and Tatsuya Harada},
    booktitle=NIPS,
    year={2024}
}

@inproceedings{CelebA-HQ,
  title={Progressive growing of GANs for improved quality, stability, and variation.},
  author={Karras, Tero and Aila, Timo and Laine, Samuli and Lehtinen, Jaakko},
  booktitle=ICLR,
  year={2018}
}

@inproceedings{lan2022e3dge,
  title={{E3DGE}: Self-Supervised Geometry-Aware Encoder for Style-based {3D} {GAN} Inversion},
  author={Lan, Yushi and Meng, Xuyi and Yang, Shuai and Loy, Chen Change and Dai, Bo},
  booktitle=CVPR,
  year={2023}
}

@inproceedings{Khirodkar2024SapiensFF,
  title={Sapiens: Foundation for Human Vision Models},
  author={Rawal Khirodkar and Timur Bagautdinov and Julieta Martinez and Zhaoen Su and Austin James and Peter Selednik and Stuart Anderson and Shunsuke Saito},
  booktitle=ECCV,
  year={2024}
}

@inproceedings{Zheng2023GPSGaussianGP,
  title={GPS-Gaussian: Generalizable Pixel-Wise 3D Gaussian Splatting for Real-Time Human Novel View Synthesis},
  author={Shunyuan Zheng and Boyao Zhou and Ruizhi Shao and Boning Liu and Shengping Zhang and Liqiang Nie and Yebin Liu},
  booktitle=CVPR,
  year={2024}
}

@article{Hastings1970MonteCS,
  title={Monte Carlo Sampling Methods Using Markov Chains and Their Applications},
  author={W. K. Hastings},
  journal={Biometrika},
  year={1970}
}

@article{Budninskiy2016OptimalVT,
  title={Optimal voronoi tessellations with hessian-based anisotropy},
  author={M. Budninskiy and Beibei Liu and Fernando de Goes and Y. Tong and Pierre Alliez and Mathieu Desbrun},
  journal=TOG,
  year={2016}
}

@inproceedings{SSIM,
  title={Image quality assessment: from error visibility to structural similarity.},
  author={Wang, Zhou and Bovik, Alan C and Sheikh, Hamid R and Simoncelli, Eero P},
  booktitle=TIP,
  year={2004},
}

@inproceedings{NeRF,
  title={{NeRF}: Representing Scenes as Neural Radiance Fields for View Synthesis},
  author={Ben Mildenhall and Pratul P. Srinivasan and Matthew Tancik and Jonathan T. Barron and Ravi Ramamoorthi and Ren Ng},
  booktitle=ECCV,
  year={2020}
}

@InProceedings{zhuang2022mofanerf,
  title={{MoFaNeRF}: Morphable Facial Neural Radiance Field},
  author={Zhuang, Yiyu and Zhu, Hao and Sun, Xusen and Cao, Xun},
  booktitle=ECCV,
  year={2022}
}

@inproceedings{zhang2022fdnerf,
      title={{FDNeRF}: Few-shot Dynamic Neural Radiance Fields for Face Reconstruction and Expression Editing},
      author={Zhang, Jingbo and Li, Xiaoyu and Wan, Ziyu and Wang, Can and Liao, Jing},
      booktitle=SIGGRAPHASIA,
      year={2022}
    }

@inproceedings{ towards-high-fidelity-nonlinear-3d-face-morphable-model, 
  author = { Luan Tran and Feng Liu and Xiaoming Liu },
  title = { Towards High-fidelity Nonlinear 3{D} Face Morphable Model },
  booktitle = CVPR,
  year = { 2019 }
}

@article{ on-learning-3d-face-morphable-model-from-in-the-wild-images, 
  author = { Luan Tran and Xiaoming Liu },
  title = { On Learning 3{D} Face Morphable Model from In-the-wild Images },
  journal = PAMI,
    year = { 2019 }
}

@inproceedings{TowardsHighFidelityFaceReconstruction,
  title={Towards High-Fidelity 3{D} Face Reconstruction from In-the-Wild Images Using Graph Convolutional Networks},
  author={Lin, Jiangke and Yuan, Yi and Shao, Tianjia and Zhou, Kun},
  booktitle=CVPR,
  year={2020}
}

@inproceedings{LuanCVPR18,
  title={Nonlinear 3{D} face morphable model.},
  author={Tran, Luan and Liu, Xiaoming},
  booktitle=CVPR,
  year={2018}
}

@inproceedings{hong2021headnerf,
     author     = {Yang Hong and Bo Peng and Haiyao Xiao and Ligang Liu and Juyong Zhang},
     title      = {{HeadNeRF}: A Real-time {NeRF}-based Parametric Head Model},
     booktitle  = CVPR,
     year       = {2022}
  }

@inproceedings{DFNRMVS, 
  author = { Bai, Ziqian and Cui, Zhaopeng and Rahim, Jamal Ahmed and Liu, Xiaoming and Tan, Ping },
  title = {Deep Facial Non-Rigid Multi-View Stereo },
  booktitle = CVPR,
  year = { 2020 }
}

@inproceedings{Pytorch, 
  author = {Adam Paszke and Sam Gross and Soumith Chintala and Gregory
Chanan and Edward Yang and Zachary DeVito and Zeming Lin and Alban Desmaison and Luca Antiga and Adam Lerer},
  title = {Automatic differentiation in {P}y{T}orch },
  booktitle = {NeurIPSW},
  year = { 2017 }
}

@inproceedings{deng2022gram,
    title={{GRAM}: Generative Radiance Manifolds for {3D}-Aware Image Generation},
    author={Deng, Yu and Yang, Jiaolong and Xiang, Jianfeng and Tong, Xin},
    booktitle=CVPR,
    year={2022}
}

@inproceedings{chanmonteiro2020pi-GAN,
                author = {Chan, Eric
                          and Monteiro, Marco
                          and Kellnhofer, Petr
                          and Wu, Jiajun
                          and Wetzstein, Gordon},
                title = {{pi-GAN}: Periodic Implicit Generative Adversarial
                  Networks for {3D}-Aware Image Synthesis},
                booktitle = CVPR,
                year={2021}
            }

@inproceedings{ 3d-face-modeling-from-diverse-raw-scan-data,
  author = { Feng Liu and Luan Tran and Xiaoming Liu },
  title = { {3D} Face Modeling from Diverse Raw Scan Data },
  booktitle = ICCV,
  year = { 2019 },
}

@inproceedings{ riggable-3d-face-reconstruction-via-in-network-optimization,
  author = { Ziqian Bai and Zhaopeng Cui and Xiaoming Liu and Ping Tan },
  title = { Riggable {3D} Face Reconstruction via In-Network Optimization },
  booktitle = CVPR,
  year = { 2021 },
}

@inproceedings{DECA,
  title = {Learning an Animatable Detailed {3D} Face Model from In-the-Wild Images},
  author = {Feng, Yao and Feng, Haiwen and Black, Michael J. and Bolkart, Timo},
  booktitle = TOG,
  year = {2021}
}

@InProceedings{NeRFace,
    author    = {Gafni, Guy and Thies, Justus and Zollh{\"o}fer, Michael and Nie{\ss}ner, Matthias},
    title     = {Dynamic Neural Radiance Fields for Monocular {4D} Facial Avatar Reconstruction},
    booktitle = CVPR,
    year      = {2021}
}

@inproceedings{
  yu25gaia,
  title={{GAIA}: Generative Animatable Interactive Avatars with Expression-conditioned Gaussians},
  author={Zhengming Yu and Tianye Li and Jingxiang Sun and Omer Shapira and Seonwook Park and Michael Stengel and Matthew Chan and Xin Li and Wenping Wang and Koki Nagano and Shalini De Mello},
  booktitle=SIGGRAPH,
  year={2025},
}

@inproceedings{he2025head,
    title={{3DGH}: {3D} Head Generation with Composable Hair and Face},
    author={He, Chengan and Li, Junxuan and Kirschstein, Tobias and Sevastopolsky, Artem and Saito, Shunsuke and Tan, Qingyang and Romero, Javier and Cao, Chen and Rushmeier, Holly and Nam, Giljoo},
    booktitle=TOG,
    year={2025}
}

@inproceedings{Kant2025Pippo,
    title={Pippo: High-Resolution Multi-View Humans from a Single Image},
    author={Yash Kant and Ethan Weber and Jin Kyu Kim and Rawal Khirodkar and Su Zhaoen
        and Julieta Martinez and Igor Gilitschenski and Shunsuke Saito and Timur Bagautdinov},
    booktitle=CVPR,
    year={2025},
}

@inproceedings{WangARXIV2025,
         title   = {Relightable Full-body Gaussian Codec Avatars},
         author  = {Shaofei Wang and Tomas Simon and Igor Santesteban and Timur Bagautdinov and Junxuan Li and Vasu Agrawal and Fabian Prada and Shoou-I Yu and Pace Nalbone and Matt Gramlich and Roman Lubachersky and Chenglei Wu and Javier Romero and Jason Saragih and Michael Zollhoefer and Andreas Geiger and Siyu Tang and Shunsuke Saito},
         booktitle = SIGGRAPH,
         year    = {2025}
}

@inproceedings{tang2024gaf,
      title={{GAF}: Gaussian Avatar Reconstruction from Monocular Videos via Multi-view Diffusion},
      author={Tang, Jiapeng and Davoli, Davide and Kirschstein, Tobias and Schoneveld, Liam and Niessner, Matthias},
      booktitle=CVPR,
      year={2025}
    }

@inproceedings{FLAME,
  title = {Learning a model of facial shape and expression from {4D} scans},
  author = {Li, Tianye and Bolkart, Timo and Black, Michael J. and Li, Hao and Romero, Javier},
  booktitle = TOG,
  year = {2017}
}

@article{shang2020self,
  title={Self-Supervised Monocular {3D} Face Reconstruction by Occlusion-Aware Multi-view Geometry Consistency},
  author={Shang, Jiaxiang and Shen, Tianwei and Li, Shiwei and Zhou, Lei and Zhen, Mingmin and Fang, Tian and Quan, Long},
  journal=ECCV,
  year={2020}
}

@InProceedings{CompositionalRF,
    	  author    = {Wang, Ziyan and Bagautdinov, Timur and Lombardi, Stephen and Simon, Tomas and 
		Saragih, Jason and Hodgins, Jessica and Zollhofer, Michael},
    	  title     = {Learning Compositional Radiance Fields of Dynamic Human Heads},
    	  booktitle = CVPR,
    	  year      = {2021},
	}

@inproceedings{LOLNeRF,
    title	= {{LOLNeRF}: Learn from One Look},
    author	= {Daniel Rebain and Mark J. Matthews and Kwang Yi and Dmitry Lagun and Andrea Tagliasacchi},
    year	= {2022},
    booktitle	= CVPR
}

@inproceedings{EG3D,
    author = {Eric R. Chan and Connor Z. Lin and Matthew A. Chan and Koki Nagano and Boxiao Pan and Shalini De Mello and Orazio Gallo and Leonidas Guibas and Jonathan Tremblay and Sameh Khamis and Tero Karras and Gordon Wetzstein},
    title = {Efficient Geometry-aware {3D} Generative Adversarial Networks},
    booktitle = CVPR,
    year = {2022}
  }

@inproceedings{GeoD,
    title   = {Improving {3D}-aware Image Synthesis with A Geometry-aware Discriminator},
    author  = {Shi, Zifan and Xu, Yinghao and Shen, Yujun and Zhao, Deli and Chen, Qifeng and Yeung, Dit-Yan},
    booktitle = NIPS,
    year    = {2022}
  }

@inproceedings{MonocularNeuralReflectance, 
  author = { B R, Mallikarjun and Tewari, Ayush and Oh, Tae-Hyun and Weyrich, Tim and Bickel, Bernd and Seidel, Hans-Peter and Pfister, Hanspeter and Matusik, Wojciech and Elgharib, Mohamed and Theobalt, Christian },
  title = {Monocular Reconstruction of Neural Face Reflectance Fields},
  booktitle = CVPR,
  year = { 2021 }
}

@inproceedings{UncertaintyFaceReconstruction,
  title={Uncertainty-Aware Mesh Decoder for High Fidelity 3{D} Face Reconstruction},
  author={Lee, Gun-Hee and Lee, Seong-Whan},
  booktitle=CVPR,
  year={2020}
}

@inproceedings{GRAF,
  title = {{GRAF}: Generative Radiance Fields for {3D}-Aware Image Synthesis},
  author = {Schwarz, Katja and Liao, Yiyi and Niemeyer, Michael and Geiger, Andreas},
  booktitle = NIPS,
  year = {2020}
}

@article{Gao-portraitnerf,
  Author    = {Gao, Chen and Shih, Yichang and Lai, Wei-Sheng and Liang, Chia-Kai and Huang, Jia-Bin},
  Title     = {Portrait Neural Radiance Fields from a Single Image},
  journal   = {arXiv preprint arXiv:2012.05903},
  year      = {2020}
}

@inproceedings{ blind-removal-of-facial-foreign-shadows,
  author = { Yaojie Liu and Andrew Hou and Xinyu Huang and Liu Ren and Xiaoming Liu },
  title = { Blind Removal of Facial Foreign Shadows },
  booktitle = BMVC,
  year = { 2022 },
}

@inproceedings{he21unsupervised,
  title     = {Unsupervised Portrait Shadow Removal via Generative Priors},
  author    = {He, Yingqing and Xing, Yazhou and Zhang, Tianjia and Chen, Qifeng},
  booktitle = ACMMM,
  year      = {2021}
}

@article{infamousnerf,
  Author    = {Andrew Hou and Feng Liu and Zhiyuan Ren and Michel Sarkis and Ning Bi and Yiying Tong and Xiaoming Liu},
  Title     = {{INFAMOUS-NeRF}: {ImproviNg FAce MOdeling Using Semantically-Aligned Hypernetworks with Neural Radiance Fields}},
  journal   = {arXiv preprint arXiv:2312.16197},
  year      = {2023}
}

@inproceedings{stylenerf,
    title={{StyleNeRF}: A Style-based {3D} Aware Generator for High-resolution Image Synthesis},
    author={Jiatao Gu and Lingjie Liu and Peng Wang and Christian Theobalt},
    booktitle=ICLR,
    year={2022}
}

@InProceedings{wang2022faceverse,
title={{FaceVerse}: a Fine-grained and Detail-controllable {3D} Face Morphable Model from a Hybrid Dataset},
author={Wang, Lizhen and Chen, Zhiyua and Yu, Tao and Ma, Chenguang and Li, Liang and Liu, Yebin},
booktitle=CVPR,
year={2022},
}

@article{park2021hypernerf,
  author = {Park, Keunhong and Sinha, Utkarsh and Hedman, Peter and Barron, Jonathan T. and Bouaziz, Sofien and Goldman, Dan B and Martin-Brualla, Ricardo and Seitz, Steven M.},
  title = {{HyperNeRF}: A Higher-Dimensional Representation for Topologically Varying Neural Radiance Fields},
  journal = TOG,
  year = {2021},
}

@inproceedings{park2021nerfies,
  author    = {Park, Keunhong and Sinha, Utkarsh and Barron, Jonathan T. and Bouaziz, Sofien and Goldman, Dan B and Seitz, Steven M. and Martin-Brualla, Ricardo},
  title     = {Nerfies: Deformable Neural Radiance Fields},
  booktitle   = ICCV,
  year      = {2021},
}

@article{joint-face-alignment,
  author = { Feng Liu and Qijun Zhao and Xiaoming Liu and Dan Zeng },
  title = { Joint Face Alignment and 3D Face Reconstruction with Application to Face Recognition },
  journal = PAMI,
  year = { 2018 },
}

@inproceedings{disentangling-features,
  author = { Feng Liu and Dan Zeng and Qijun Zhao and Xiaoming Liu },
  title = { Disentangling Features in {3D} Face Shapes for Joint Face Reconstruction and Recognition },
  booktitle = CVPR,
  year = { 2018 },
}

@inproceedings{NHA,
  title={Neural Head Avatars from Monocular RGB Videos},
  author={Grassal, Philip-William and Prinzler, Malte and Leistner, Titus and Rother, Carsten and Nie{\ss}ner, Matthias and Thies, Justus},
  booktitle=CVPR,
  year={2022}
}

@inproceedings{3DMM,
  title={A morphable model for the synthesis of 3{D} faces},
  author={Blanz, Volker and Vetter, Thomas},
  booktitle=SIGGRAPH,
  year={1999}
}

@inproceedings{FFHQ,
  title={A style-based generator architecture for generative adversarial networks.},
  author={Karras, Tero and Laine, Samuli and Aila, Timo},
  booktitle=CVPR,
  year={2019}
}

@inproceedings{RigNeRF,
  title={{RigNeRF}: Fully Controllable Neural {3D} Portraits},
  author={ShahRukh Athar and Zexiang Xu and Kalyan Sunkavalli and Eli Shechtman and Zhixin Shu},
  year={2022},
  booktitle=CVPR
}

@inproceedings{3DFaceFill,
      title={{3DFaceFill}: An Analysis-By-Synthesis Approach to Face Completion},
      author={Rahul Dey and Vishnu Boddeti},
      year={2022},
      booktitle=WACV,
}

@inproceedings{diffusionrig,
      author    = {Zheng Ding and Cecilia Zhang and Zhihao Xia and Lars Jebe and Zhuowen Tu and Xiuming Zhang},
      title     = {{DiffusionRig}: Learning Personalized Priors for Facial Appearance Editing},
      booktitle = CVPR,
      year      = {2023},
}

@inproceedings{shi2023pof3d,
    title   = {Learning {3D}-aware Image Synthesis with Unknown Pose Distribution},
    author  = {Zifan Shi and Yujun Shen and Yinghao Xu and Sida Peng and Yiyi Liao and Sheng Guo and Qifeng Chen and Dit-Yan Yeung},
    booktitle = CVPR,
    year    = {2023}
}

@inproceedings{TaoAvatar,
    title   = {{TaoAvatar}: Real-Time Lifelike Full-Body Talking Avatars for Augmented Reality via {3D} Gaussian Splatting},
    author  = {Jianchuan Chen and Jingchuan Hu and GaiGe Wang and Zhonghua Jiang and TianSong Zhou and Zhiwen Chen and Chengfei Lv},
    booktitle = CVPR,
    year    = {2025}
}

@inproceedings{FFHQUV,
    title   = {{FFHQ-UV}: Normalized Facial {UV}-Texture Dataset for {3D} Face Reconstruction},
    author  = {Haoran Bai and Di Kang and Haoxian Zhang and Jinshan Pan and Linchao Bao},
    booktitle = CVPR,
    year    = {2023}
}

@inproceedings{ face-relighting-with-geometrically-consistent-shadows,
  author = { Andrew Hou and Michel Sarkis and Ning Bi and Yiying Tong and Xiaoming Liu },
  title = { Face Relighting with Geometrically Consistent Shadows },
  booktitle=CVPR,
  year = { 2022 }
}

@inproceedings{ 3DMM-RF,
  author = { Stathis Galanakis and Baris Gecer and Alexandros Lattas and Stefanos Zafeiriou },
  title = { {3DMM-RF}: Convolutional Radiance Fields for {3D} Face Modeling},
  booktitle=WACV,
  year = { 2023 }
}

@inproceedings{kim2022adaface,
  title={{AdaFace}: Quality Adaptive Margin for Face Recognition},
  author={Kim, Minchul and Jain, Anil K and Liu, Xiaoming},
  booktitle=CVPR,
  year={2022}
}

@inproceedings{NeRFInvertor,
  title={{NeRFInvertor}: High Fidelity {NeRF-GAN} Inversion for Single-Shot Real Image Animation.},
  author={Yu Yin and Kamran Ghasedi and HsiangTao Wu and Jiaolong Yang and Xin Tong and Yun Fu},
  booktitle=CVPR,
  year={2023}
}

@inproceedings{QuantManip,
  title={Quantitative Manipulation of Custom Attributes on {3D}-Aware Image Synthesis},
  author={Hoseok Do and EunKyung Yoo and Taehyeong Kim and Chul Lee and Jin young Choi},
  booktitle=CVPR,
  year={2023}
}

@Article{gaussiansplatting,
      author       = {Kerbl, Bernhard and Kopanas, Georgios and Leimk{\"u}hler, Thomas and Drettakis, George},
      title        = {{3D} Gaussian Splatting for Real-Time Radiance Field Rendering},
      journal      = TOG,
      year         = {2023}
}

@inproceedings{relightablegaussianavatars,
          author = {Shunsuke Saito and Gabriel Schwartz and Tomas Simon and Junxuan Li and Giljoo Nam},
          title = {Relightable Gaussian Codec Avatars}, 
          booktitle = CVPR,
          year = {2024},
        }

@inproceedings{gaussianheadavatar,
                title={Gaussian Head Avatar: Ultra High-fidelity Head Avatar via Dynamic Gaussians},
                author={Xu, Yuelang and Chen, Benwang and Li, Zhe and Zhang, Hongwen and Wang, Lizhen and Zheng, Zerong and Liu, Yebin},
                booktitle=CVPR,
                year={2024}
              }

@inproceedings{qiangaussianavatars,
  title={{GaussianAvatars}: Photorealistic Head Avatars with Rigged {3D} Gaussians},
  author={Qian, Shenhan and Kirschstein, Tobias and Schoneveld, Liam and Davoli, Davide and Giebenhain, Simon and Nie{\ss}ner, Matthias},
  booktitle=CVPR,
  year={2024}
}

@InProceedings{PanoHead,
    author    = {An, Sizhe and Xu, Hongyi and Shi, Yichun and Song, Guoxian and Ogras, Umit Y. and Luo, Linjie},
    title     = {{PanoHead}: Geometry-Aware {3D} Full-Head Synthesis in 360deg},
    booktitle = CVPR,
    year      = {2023}
}

@inproceedings{Loshchilov2017DecoupledWD,
  title={Decoupled Weight Decay Regularization},
  author={Ilya Loshchilov and Frank Hutter},
  booktitle=ICLR,
  year={2017},
}

@inproceedings{ma2024gaussianblendshapes,
  author     = {Shengjie Ma and Yanlin Weng and Tianjia Shao and Kun Zhou},
  title      = {{3D} Gaussian Blendshapes for Head Avatar Animation},
  booktitle  = SIGGRAPH,
  year       = {2024},
}

@inproceedings{chen2024monogaussianavatar,
  title={Monogaussianavatar: Monocular gaussian point-based head avatar},
  author={Chen, Yufan and Wang, Lizhen and Li, Qijing and Xiao, Hongjiang and Zhang, Shengping and Yao, Hongxun and Liu, Yebin},
  booktitle=SIGGRAPH,
  year={2024}
}

@inproceedings{shao2024splattingavatar,
  title = {{SplattingAvatar: Realistic Real-Time Human Avatars with Mesh-Embedded Gaussian Splatting}},
  author = {Shao, Zhijing and Wang, Zhaolong and Li, Zhuang and Wang, Duotun and Lin, Xiangru and Zhang, Yu and Fan, Mingming and Wang, Zeyu},
  booktitle = CVPR,
  year = {2024}
}

@inproceedings{malp,
    author    = {Asnani, Vishal and Yin, Xi and Hassner, Tal and Liu, Xiaoming},
    title     = {{MaLP}: Manipulation Localization Using a Proactive Scheme},
    booktitle = CVPR,
    year      = {2023}
}

@article{ reverse-engineering-of-generative-models-inferring-model-hyperparameters-from-generated-images,
  author = { Vishal Asnani and Xi Yin and Tal Hassner and Xiaoming Liu },
  title = { Reverse Engineering of Generative Models: Inferring Model Hyperparameters from Generated Images },
  journal = PAMI,
  year = { 2023 },
}

@inproceedings{ hierarchical-fine-grained-image-forgery-detection-and-localization,
  author = { Xiao Guo and Xiaohong Liu and Zhiyuan Ren and Steven Grosz and Iacopo Masi and Xiaoming Liu },
  title = { Hierarchical Fine-Grained Image Forgery Detection and Localization },
  booktitle = CVPR,
  year = { 2023 }
}

@inproceedings{interpretabledeepfake,
  author = {Agil Aghasanli and Dmitry Kangin and Plamen Angelov},
  title = { Interpretable-through-prototypes deepfake detection for diffusion models },
  booktitle = { ICCVW },
  year = { 2023 }
}

@inproceedings{intrinsicdeepfake,
  author = {Peter Lorenz and Ricard L. Durall and Janis Keuper},
  title = { Detecting Images Generated by Deep Diffusion Models using their Local
Intrinsic Dimensionality },
  booktitle = { ICCVW },
  year = { 2023 }
}

@inproceedings{icasspdeepfake,
  author = {Riccardo Corvi and Davide Cozzolino and Giada Zingarini and Giovanni Poggi and Koki Nagano and Luisa Verdoliva},
  title = { ON THE DETECTION OF SYNTHETIC IMAGES GENERATED BY DIFFUSION MODELS },
  booktitle = ICASSP,
  year = { 2023 }
}

@inproceedings{revisitingdeepfake,
  author = {Sarthak Kamat and Shruti Agarwal and Trevor Darrell and Anna Rohrbach},
  title = { Revisiting Generalizability in Deepfake Detection: Improving Metrics and
Stabilizing Transfer },
  booktitle = { ICCVW },
  year = { 2023 }
}

@inproceedings{universaldeepfake,
  author = {Utkarsh Ojha and Yuheng Li and Yong Jae Lee},
  title = {Towards universal fake image detectors that generalize across generative models},
  booktitle = CVPR,
  year = { 2023 }
}

@inproceedings{LPIPS,
  title={The Unreasonable Effectiveness of Deep Features as a Perceptual Metric},
  author={Zhang, Richard and Isola, Phillip and Efros, Alexei A and Shechtman, Eli and Wang, Oliver},
  booktitle=CVPR,
  year={2018}
}

@inproceedings{Hu2023GaussianAvatarTR,
  title={GaussianAvatar: Towards Realistic Human Avatar Modeling from a Single Video via Animatable 3D Gaussians},
  author={Liangxiao Hu and Hongwen Zhang and Yuxiang Zhang and Boyao Zhou and Boning Liu and Shengping Zhang and Liqiang Nie},
  booktitle=CVPR,
  year={2024}
}

@InProceedings{Lee_2024_CVPR,
  author    = {Lee, Joo Chan and Rho, Daniel and Sun, Xiangyu and Ko, Jong Hwan and Park, Eunbyung},
  title     = {Compact {3D} Gaussian Representation for Radiance Field},
  booktitle = CVPR,
  year      = {2024},
}

@InProceedings{Niedermayr_2024_CVPR,
    author    = {Niedermayr, Simon and Stumpfegger, Josef and Westermann, R\"udiger},
    title     = {Compressed {3D} Gaussian Splatting for Accelerated Novel View Synthesis},
    booktitle = CVPR,
    year      = {2024},
}

@article{navaneet2023compact3d,
          title={{Compact3D}: Smaller and Faster Gaussian Splatting with Vector Quantization},
          author={Navaneet, KL and Meibodi, Kossar Pourahmadi and Koohpayegani, Soroush Abbasi and Pirsiavash, Hamed},
          journal={arXiv preprint arXiv:2311.18159},
          year={2023}
        }

@inproceedings{fan2023lightgaussian,
  author    = {Zhiwen Fan and Kevin Wang and Kairun Wen and Zehao Zhu and Dejia Xu and Zhangyang Wang},
  title     = {{LightGaussian: Unbounded 3D Gaussian Compression with 15x Reduction and 200+ FPS}},
  year      ={2024}, 
  booktitle=NIPS, 
}

@inproceedings{eagles,
  author    = {Sharath Girish and Kamal Gupta and Abhinav Shrivastava},
  title     = {{EAGLES: Efficient Accelerated 3D Gaussians with Lightweight EncodingS}},
  year      ={2024}, 
  booktitle=ECCV, 
}

@Article{reducingmemfootprintGS,
                author       = {Papantonakis, Panagiotis and Kopanas, Georgios and Kerbl, Bernhard and Lanvin, Alexandre and Drettakis, George},
                title        = {Reducing the Memory Footprint of {3D} Gaussian Splatting},
                journal      = {Proceedings of the ACM on Computer Graphics and Interactive Techniques},
                year         = {2024},
          }

@inproceedings{GNeRF,
  author = {Zixiong, Huang and Qi, Chen and Libo, Sun and Yifan, Yang and Naizhou, Wang and Mingkui, Tan and Qi, Wu},
  title = {{G-NeRF}: Geometry-enhanced Novel View Synthesis from Single-View Images},
  booktitle = CVPR,
  year = {2024}
}

@inproceedings{therelightables,
  title={The {Relightables}: Volumetric performance capture of humans with realistic relighting},
  author={Kaiwen Guo and Peter Lincoln and Philip Davidson and Jay Busch and Xueming Yu and Matt Whalen and Geoff Harvey and Sergio Orts-Escolano and Rohit Pandey and Jason Dourgarian and Matthew DuVall and Danhang Tang and Anastasia Tkach and Adarsh Kowdle and Emily Cooper and Mingsong Dou and Sean Fanello and Graham Fyffe and Christoph Rhemann and Jonathan Taylor and Paul Debevec and Shahram Izadi},
  booktitle=TOG,
  year={2019}
}

@inproceedings{FID,
  author = {Martin Heusel and Hubert Ramsauer and Thomas Unterthiner and
Bernhard Nessler and Gunter Klambauer and Sepp Hochreiter},
  title = {{GANs} trained by a two time-scale update rule converge to
a nash equilibrium},
  booktitle = NIPS,
  year = {2017}
}

@article{Canny1986ACA,
  title={A Computational Approach to Edge Detection},
  author={John F. Canny},
  journal=PAMI,
  year={1986}
}

@Article{dlib09,
  author = {Davis E. King},
  title = {Dlib-ml: A Machine Learning Toolkit},
  journal = {Journal of Machine Learning Research},
  year = {2009},
  volume = {10},
  pages = {1755-1758},
}

@article{NeRSembleMR,
  title={NeRSemble: Multi-view Radiance Field Reconstruction of Human Heads},
  author={Tobias Kirschstein and Shenhan Qian and Simon Giebenhain and Tim Walter and Matthias Nie{\ss}ner},
  journal=TOG,
  year={2023},
}

@inproceedings{cap4d,
  title={CAP4D: Creating Animatable 4D Portrait Avatars with Morphable Multi-View Diffusion Models},
  author={Felix Taubner and Ruihang Zhang and Mathieu Tuli and David B. Lindell},
  booktitle=CVPR,
  year={2025},
}

@inproceedings{facelift,
  title={FaceLift: Learning Generalizable Single Image 3D Face Reconstruction from Synthetic Heads},
  author={Weijie Lyu and Yi Zhou and Ming-Hsuan Yang and Zhixin Shu},
  booktitle=ICCV,
  year={2025},
}

@inproceedings{kwon2024ghg,
  title={Generalizable Human Gaussians for Sparse View Synthesis},
  author={Youngjoong Kwon and Baole Fang and Yixing Lu and Haoye Dong and Cheng Zhang and Francisco Vicente Carrasco and Albert Mosella-Montoro and Jianjin Xu and Shingo Takagi and Daeil Kim and Aayush Prakash and Fernando De la Torre},
  booktitle=ECCV,
  year={2024}
}

@inproceedings{yu2018bisenet,
  title={Bisenet: Bilateral segmentation network for real-time semantic segmentation},
  author={Yu, Changqian and Wang, Jingbo and Peng, Chao and Gao, Changxin and Yu, Gang and Sang, Nong},
  booktitle=ECCV,
  year={2018}
}

@inproceedings{kirschstein2024gghead,
    author = {Kirschstein, Tobias and Giebenhain, Simon and Tang, Jiapeng and Georgopoulos, Markos and Nie\ss{}ner, Matthias},
    title = {{GGHead: Fast and Generalizable 3D Gaussian Heads}},
    year = {2024},
    booktitle = SIGGRAPHASIA,
}
}

\end{document}